\documentclass[10pt,twocolumn,letterpaper]{article}

\usepackage{listliketab}
\PassOptionsToPackage{table}{xcolor}
\PassOptionsToPackage{capitalize}{cleveref}

\usepackage{marvosym} %
\usepackage{cvpr}      %

\usepackage[dvipsnames]{xcolor}

\newcommand{\bH}{\mathbf{H}}

\newcommand{\cI}{\mathcal{I}}

\newcommand{\cL}{\mathcal{L}}

\renewcommand{\eqref}[1]{Eq.~\ref{#1}}

\def\mc{\mathcal}
\def\mb{\mathbf}
\def\it{\textit}

\makeatletter
\DeclareRobustCommand\onedot{\futurelet\@let@token\@onedot}
\def\@onedot{\ifx\@let@token.\else.\null\fi\xspace}

\makeatother

\definecolor{applegreen}{rgb}{0.55, 0.71, 0.0}

\newcommand{\boldparagraph}[1]{\vspace{0.2cm}\noindent{\bf #1:} }
\newcolumntype{P}[1]{>{\centering\arraybackslash}m{#1}}

\newcommand{\del}[1]{}

\newif\ifcomment
\commenttrue
\ifcomment
	\newcommand{\ag}[1]{ \noindent {\color{red} {\bf Andreas:} {#1}} }
	\newcommand{\yl}[1]{ \noindent {\color{cyan} {\bf Yiyi:} {#1}} }

\else
	\newcommand{\ag}[1]{}
	\newcommand{\jx}[1]{}
	\newcommand{\kc}[1]{}
	\newcommand{\yl}[1]{}
\fi

\definecolor{cvprblue}{rgb}{0.21,0.49,0.74}
\usepackage[pagebackref,breaklinks,colorlinks,citecolor=cvprblue]{hyperref}
\usepackage{cleveref}
\usepackage{multirow}
\usepackage{pifont}
\usepackage[table]{xcolor}

\def\paperID{183} %
\def\paperName{FreeFix\xspace}
\def\confName{3DV\xspace}
\def\confYear{2026\xspace}

\crefname{figure}{Fig.}{Figure}

\title{FreeFix: Boosting 3D Gaussian Splatting via Fine-Tuning-Free Diffusion Models}

\author{
Hongyu Zhou$^{1}$,
Zisen Shao$^{2}$,
Sheng Miao$^{1}$,
Pan Wang$^{3}$,
Dongfeng Bai$^{3}$ \\
Bingbing Liu$^{3}$,
Yiyi Liao$^{1}\textsuperscript{\Letter}$
\\
{\normalsize $^{1}$ Zhejiang University \quad $^{2}$ University of Maryland, College Park \quad $^{3}$ Huawei}
\\
}

\begin{document}

\twocolumn[{%
\renewcommand\twocolumn[1][]{#1}%
\maketitle
\vspace{-1.0cm}
\begin{center}
    \centering
    \captionsetup{type=figure}
    \includegraphics[width=\textwidth]{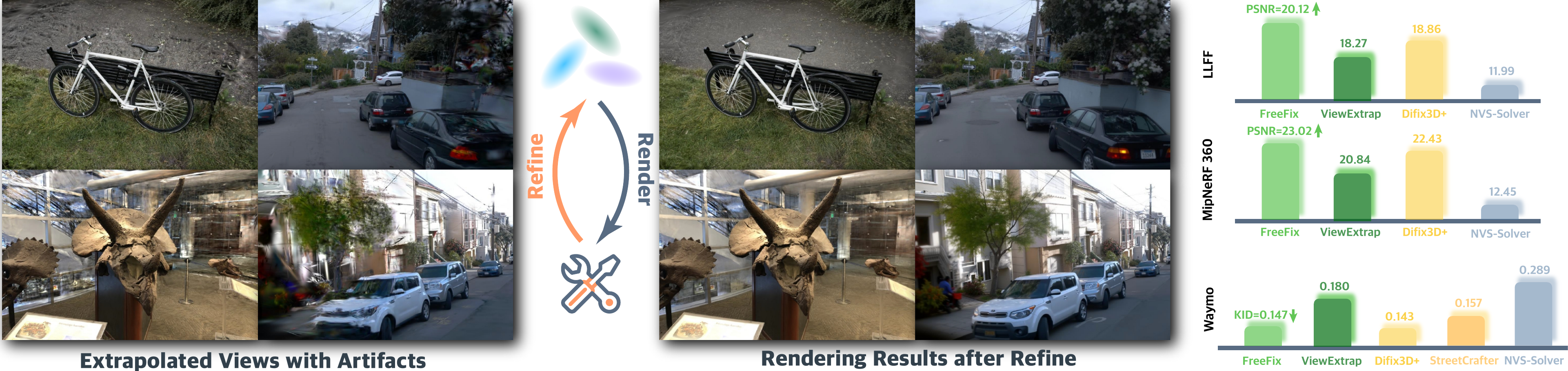}
    \vspace{-0.6cm}
    \captionof{figure}{\textbf{Overview of \paperName.} We present \paperName, a method designed to improve the rendering results of extrapolated views in 3D Gaussian Splatting, without requiring fine-tuning of diffusion models. Experiments on multiple datasets show that \paperName provides performance that is comparable to, or even superior to, most advanced methods that require fine-tuning.}
    \label{fig:teaser}
\end{center}%
}]

\begin{abstract}
Neural Radiance Fields and 3D Gaussian Splatting have advanced novel view synthesis, yet still rely on dense inputs and often degrade at extrapolated views. Recent approaches leverage generative models, such as diffusion models, to provide additional supervision, but face a trade-off between generalization and fidelity: fine-tuning diffusion models for artifact removal improves fidelity but risks overfitting, while fine-tuning-free methods preserve generalization but often yield lower fidelity. We introduce \paperName, a fine-tuning-free approach that pushes the boundary of this trade-off by enhancing extrapolated rendering with pretrained image diffusion models. We present an interleaved 2D–3D refinement strategy, showing that image diffusion models can be leveraged for consistent refinement without relying on costly video diffusion models. Furthermore, we take a closer look at the guidance signal for 2D refinement and propose a per-pixel confidence mask to identify uncertain regions for targeted improvement. Experiments across multiple datasets show that \paperName improves multi-frame consistency and achieves performance comparable to or surpassing fine-tuning-based methods, while retaining strong generalization ability. Our project page is at \href{https://xdimlab.github.io/freefix}{https://xdimlab.github.io/freefix}.

\end{abstract}

\vspace{-0.5cm}
    
\section{Introduction}
 
Novel view synthesis (NVS) is a fundamental problem in 3D computer vision, playing an important role in advancing mixed reality and embodied artificial intelligence. Neural Radiance Fields (NeRF) \cite{mildenhall2021nerf} and 3D Gaussian Splatting (3DGS) \cite{kerbl20233d} have achieved high-fidelity rendering, with 3DGS in particular becoming the mainstream choice for its real-time rendering capability. However, both methods require densely captured training images, which are often difficult to obtain, and they tend to produce artifacts at extrapolated viewpoints, namely those outside the interpolation range of the training views. These limitations hinder their use in downstream applications such as autonomous driving simulation and free-viewpoint user experiences.

Recent work has explored addressing artifacts in extrapolated view rendering with 3DGS. Existing approaches fall into two categories: adding regularization terms during training or augmenting supervision views using generative models. The regularization terms are often derived from 3D priors \cite{yu2022monosdf, zhou2024hugsim, khan2024autosplat, zeng2025oblique, wang2024planerf}, or additional sensors \cite{pan2025pings}, but they are typically hand-crafted and limited to specific scene types. Moreover, their lack of hallucination capability further restricts their applicability.
In leveraging diffusion models (DMs), some approaches fine-tune them with paired data, e.g., by using sparse LiDAR inputs or extrapolated renderings with artifacts to generate refined images. Many of these methods train on domain-specific datasets, such as those for autonomous driving \cite{yan2025streetcrafter, wang2024freevs, ni2025recondreamer, wang2024stag}, which inevitably compromises the generalization ability of DMs. More recently, Difix3D+ \cite{wu2025difix3d+} fine-tunes SD Turbo \cite{sauer2024fast} on a wider range of 3D datasets, improving generalization. However, the substantial effort required to curate 3D data and the high fine-tuning cost make this approach time-consuming and expensive to extend to other DMs.
An alternative line of work seeks to improve extrapolated rendering without fine-tuning, typically by providing extrapolated renderings as guidance during the denoising step. This preserves the generalization capacity of DMs trained on large-scale data, but such methods still lag behind fine-tuned approaches that are specifically adapted to the task.

Given the generalization–fidelity trade-off, we ask: can extrapolated view rendering be improved with DMs without sacrificing generalization? To address this challenge, we focus on fine-tuning-free methods and enhance their effectiveness for NVS extrapolation. This is achieved with our proposed \textit{2D–3D interleaved refinement strategy} combined with \textit{per-pixel confidence guidance for fine-tuning-free image refinement}. Specifically, given a trained 3DGS, we sample an extrapolated viewpoint, render the 2D image, refine it with a 2D image diffusion model (IDMs), and integrate the refined image back into the 3D scene by updating the 3DGS before proceeding to the next viewpoint.
This interleaved 2D-3D refinement ensures that previously enhanced views inform subsequent 2D refinements and improve multi-view consistency. Importantly, we introduce a confidence-guided 2D refinement, where a per-pixel confidence map rendered from the 3DGS highlights regions requiring further improvement by the 2D DM. This contrasts with previous training-free methods that rely solely on rendering opacity, leaving the DM to identify artifact regions on its own. While our confidence guidance could in principle be applied to video diffusion models (VDMs), advanced video backbones are typically more computationally expensive and use temporal down-sampling, which prevents the direct use of per-pixel guidance. We show that our 2D–3D interleaved optimization strategy achieves consistent refined images without relying on VDMs.

Our contribution can be summarized as follows: 1) We propose a simple yet effective approach for enhancing extrapolated 3DGS rendering without the need for fine-tuning DMs, featuring a 2D–3D interleaved refinement strategy and per-pixel confidence guidance. 2) Our method is compatible with various DMs and preserves generalization across diverse scene contents. 3) Experimental results demonstrate that our approach significantly outperforms existing fine-tuning-free methods and achieves comparable or even superior performance to training-based methods.

\section{Related Work}

Numerous works have made efforts on improving quality of NVS. In this section, we will discuss related works in NVS and 3D reconstruction. Furthermore, we will explore efforts that improve NVS quality by incorporating priors from geometry, physics or generative models.

\boldparagraph{Novel View Synthesis}
NVS aims to generate photorealistic images of a scene from novel viewpoints. Early methods primarily relied on traditional image-based rendering techniques, such as Light Field Rendering \cite{levoy2023light}, Image-Based Rendering \cite{shum2007image}, and Multi-Plane Image \cite{zhou2018stereo, tucker_single-view_2020}. These approaches typically interpolate between existing views and are often limited by dense input imagery and struggle with complex occlusions. The advent of deep learning revolutionized NVS, led by two major paradigms: NeRF \cite{mildenhall2021nerf} and 3DGS \cite{kerbl20233d}. NeRF implicitly represents a scene and achieves high-quality results, but its training and rendering speeds are slow. In contrast, 3DGS offers rapid training and real-time rendering. However, a significant limitation of 3DGS is the occurrence of visual artifacts in extrapolated views, which are viewpoints far from the training data. These artifacts compromise the realism and geometric fidelity of the synthesized images. Mitigating these artifacts is the focus of this paper.

\boldparagraph{NVS with Geometry Priors}
To enhance the robustness of NVS models and reduce reconstruction ambiguity, many works have introduced geometry priors. These priors provide key information about the scene's 3D structure, which can be explicitly provided by external sensors like LiDAR or depth cameras \cite{pan2025pings, yan2025streetcrafter, wang2024freevs, raj2025spurfies, yan2024gs, matsuki2024gaussian, keetha2024splatam}. Other methods utilize strong structural priors often found in real-world scenes, such as the assumption that the ground is a flat plane \cite{zhou2024hugsim, khan2024autosplat, feng2024rogs}, the sky can be modeled as a dome \cite{chen2024g3r, ye2024gaustudio}, or that walls and tables in indoor scenes are predominantly orthogonal \cite{yu2022monosdf}. These structural assumptions help regularize the reconstruction process. While these geometry priors can mitigate some reconstruction challenges, they often fall short of completely solving the artifact problem in extrapolated views, especially when the initial geometric prior is itself inaccurate.

\begin{figure*}
    \centering
    \includegraphics[width=\textwidth]{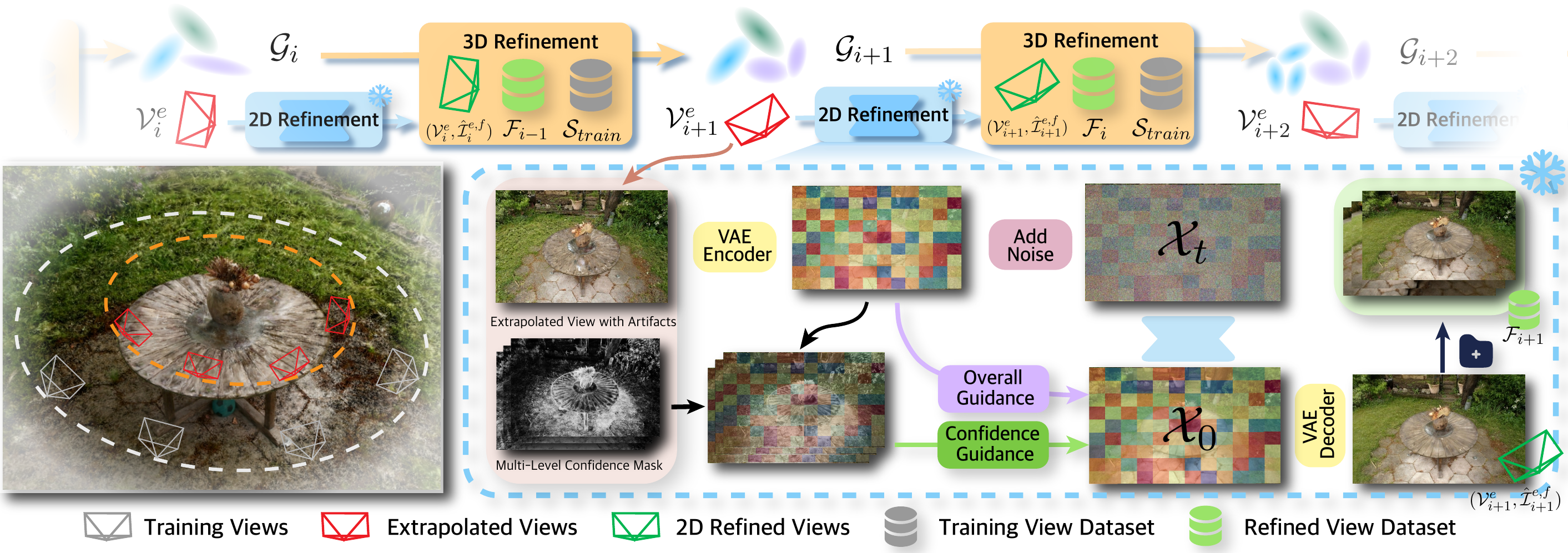}
    \vspace{-0.5cm}
    \caption{\textbf{Method.} \paperName improves the rendering quality of extrapolated views in 3DGS without fine-tuning DMs, as illustrated in the bottom left of the pipeline. We propose an interleaved strategy that combines 2D and 3D refinement to utilize image diffusion models for generating multi-frame consistent results, as shown at the top of the pipeline. In the 2D refinement stage, we also introduce confidence guidance and overall guidance to enhance the quality and consistency of the denoising results.}
\vspace{-0.3cm}
\label{fig: pipeline}
\end{figure*}

\boldparagraph{NVS with Generative Priors}
Generative priors leverage pre-trained generative models to assist NVS tasks, particularly when dealing with data scarcity or missing information. Early works explored using Generative Adversarial Networks (GANs) to improve rendering quality \cite{xu2019view, ramirez2021unsupervised, schwarz2020graf}, where the GAN's discriminator ensured the local realism of synthesized images. More recently, DMs \cite{wang_planerf_2023, podell2023sdxl, flux2024, wan2025wan, yang2024cogvideox, kong2024hunyuanvideo, labs2025flux, wang2025seedvr} have gained prominence for their powerful generative capabilities. Their application in NVS falls into two main categories. The first involves fine-tuning a pre-trained DM, which has learned powerful priors from datasets \cite{wu2025difix3d+, yan2025streetcrafter, wang2024stag, wu2024reconfusion, zhou2025stable, yu2025sgd, yu2024viewcrafter}. This process adapts the model's knowledge to scene-specific appearances but can be computationally expensive and time-consuming. The second category, which aligns with our proposed method, leverages a pre-trained DM as a zero-shot prior without fine-tuning. The key challenge here is determining what part of the rendered image should be used as guidance for the DM, and how to maintain multi-view consistency. Using the opacity channel of the rendered image as guidance is a common but often crude solution \cite{you2024nvs, liu2024novel, yu2025wonderworld}, as areas with high opacity can still be artifacts. Additionally, ensuring consistency across different novel views using IDMs is a critical problem. While VDMs \cite{wang_planerf_2023, wan2025wan, yang2024cogvideox, kong2024hunyuanvideo} can inherently handle this, they are often computationally heavy and not suitable for all applications.

\section{Method}

The \paperName pipeline is illustrated in \cref{fig: pipeline}. In this section, we will first define our task and the relevant notations in \cref{sec: preliminaries}. Next, we will introduce the interleaved refinement strategy for 2D and 3D refinement in \cref{sec: interleave}. Finally, we will discuss the guidance utilized in diffusion denoising in \cref{sec: guide_method}.

\subsection{Preliminaries} \label{sec: preliminaries}

\boldparagraph{Task Definition} 
In the paper, we focus on the task of refining existing 3DGS. Specifically, given a 3DGS model $\mc{G}_\it{init}$ reconstructed from sparse view or partial observations $\mc{S}_\it{train} = \{ (\mc{V}^t_0, \mc{I}^t_0), (\mc{V}^t_1, \mc{I}^t_1), ..., (\mc{V}^t_n, \mc{I}^t_n) \}$, artifacts tend to appear on the rendering results $\pi(\mc{V}_i^e; \mc{G}_\it{init})$, which are rendered from a continuous trajectory consisting of $m$ extrapolated views $\mc{T}_\it{ext} = \{\mc{V}^e_0, \mc{V}^e_1, ..., \mc{V}^e_m \}$. Our objective is to fix these artifacts in the extrapolated views and refine the initial 3DGS into $\mc{G}_\it{refined}$. The extrapolated view rendering results from the refined 3DGS, $\pi(\mc{V}^e_i; \mc{G}_\it{refined})$, are expected to show improvements over the initial 3DGS results.

\boldparagraph{3D Gaussian Splatting} 
3D Gaussian Splatting defines 3D Gaussians as volumetric particles, which are parameterized by their positions $\mb{\mu}$, rotations $\mb{q}$, scales $\mb{s}$, opacities $\mb{\eta}$, and color $\mb{c}$. The covariance $\mb{\Sigma}$ of 3D Gaussians is defined as $\mb{\Sigma} = \mb{R}\mb{S}\mb{S}^T\mb{R}^T$, where $\mb{R} \in \mb{SO}(3)$ and $\mb{S} \in \mathbb{R}^{3\times3}$ represent the matrix formats of $\mb{q}$ and $\mb{s}$. Novel views can be rendered from 3DGS as follows:
\begin{align}
    & \alpha_i = \mb{\eta}_i \exp[-\frac{1}{2}(\mb{p}-\mb{\mu}_i)^T \mb{\Sigma}_i^{-1} (\mb{p}-\mb{\mu}_i)] \nonumber \\
    & \pi(\mc{V}; \mc{G}) = \sum_{i=1}^{N} \alpha_i \mb{c}_i \prod_j^{i-1}(1-\alpha_i) 
\end{align}
Note that $\mb{c}_i$ can be replaced as other attributions to render additional modalities. For example, $\pi(\mc{V}; (\mc{G}, \mb{d}_i)) = \sum_{i=1}^{N} \alpha_i \mb{d}_i \prod_j^{i-1}(1-\alpha_i)$ denotes the rendering of a depth map, where $\mb{d}_i$ represents the depth of each Gaussian relative to viewpoint $\mc{V}$.

\begin{figure}[t!]
     \centering
     \small 
     \setlength{\tabcolsep}{0pt}
     \def\mywidth{3.9cm}
     \begin{tabular}{P{\mywidth}P{\mywidth}}
     
     \includegraphics[width=\mywidth]{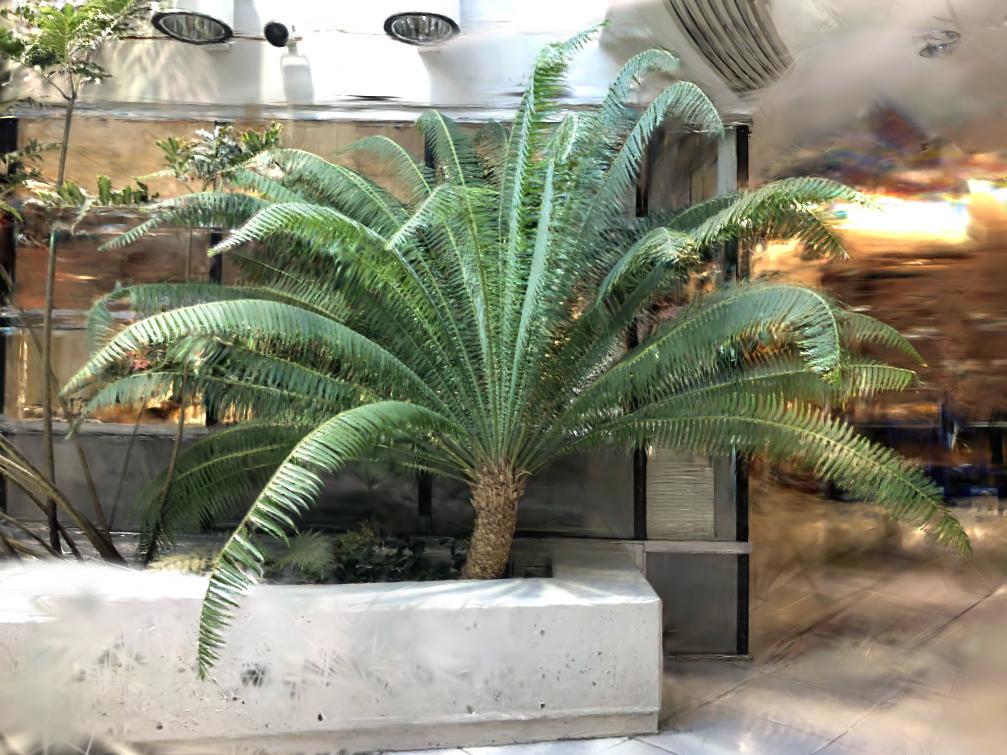}  &
     \includegraphics[width=\mywidth]{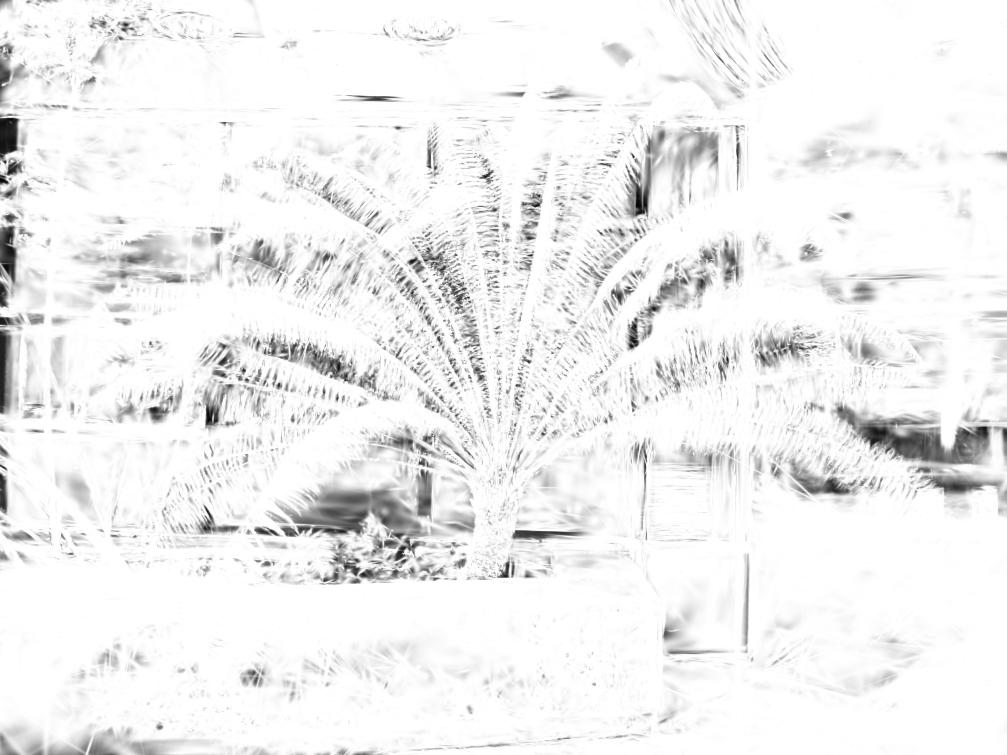}  \\

     {Rendered RGB w Artifacts} & {Rendered Opacity Map \bf{(a)}} \\

     \includegraphics[width=\mywidth]{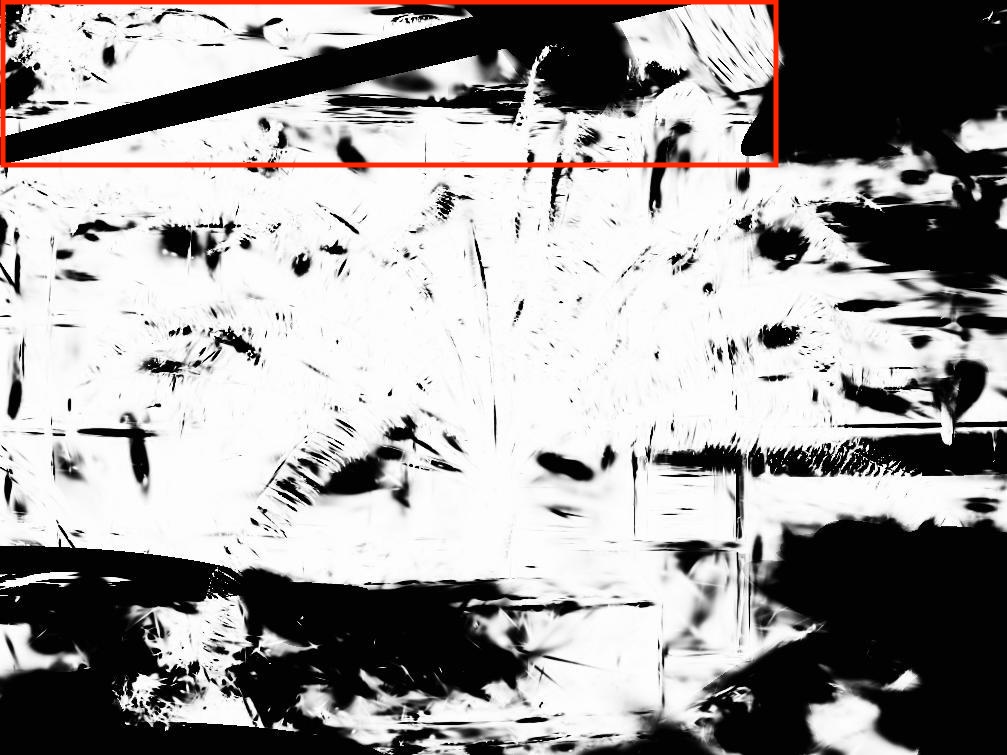} &
     \includegraphics[width=\mywidth]{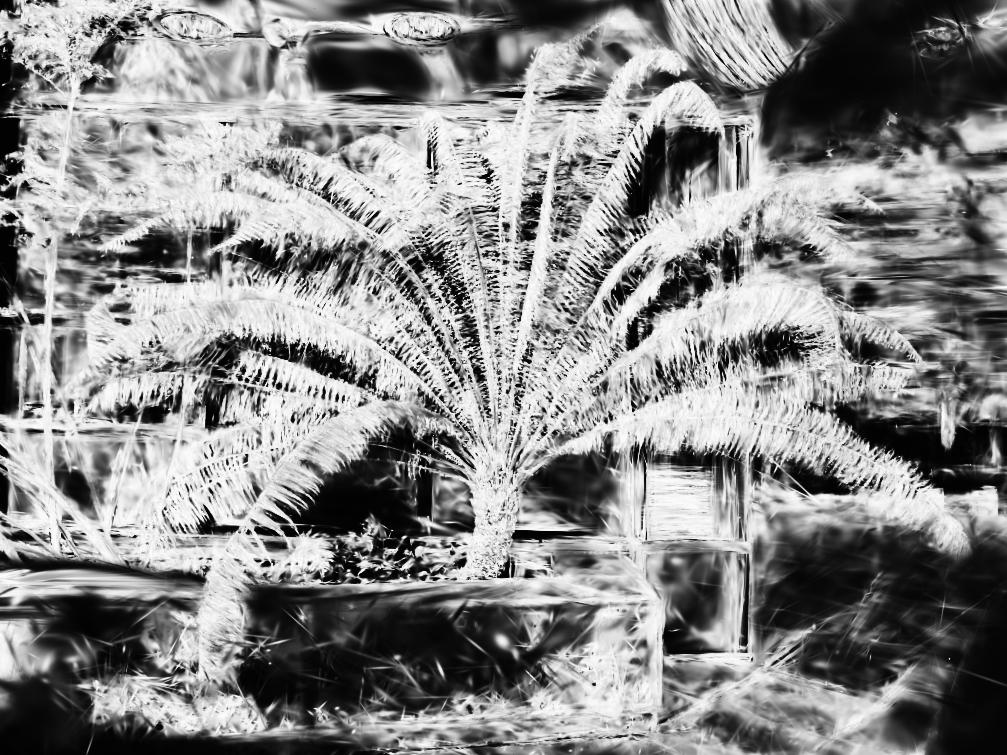} \\

     {1 - Uncertainty Mask \bf{(b)}} & {Certainty Mask \bf{(c)}}
     
     \end{tabular}
     \vspace{-0.2cm}
     \caption{\textbf{Masks Comparison.}
     We aim to generate masks for guidance during denoising to fix artifacts in rendered RGBs. \textbf{(a)} Rendered opacity maps do not account for the presence of artifacts. \textbf{(b)} Uncertainty Masks are aware of artifacts; however, due to their numerical instability, the volume rendering processing can be \textit{overwhelmed} by low-opacity Gaussians with large uncertainties. \textbf{(c)} The certainty mask we propose is numerically stable and robust against various types of artifacts.
     }
     \vspace{-0.4cm}
\label{fig: masks_comp}
\end{figure}

\boldparagraph{Diffusion Models}
DMs generate a prediction $\hat{x}_0 \sim p_\it{data}$ that aligns with real-world distribution through iterative denoising. Specifically, the input of DMs is pure noise $\epsilon \sim \mc{N}(0, I)$ or real world data with added noise $x_t = (1 - \sigma)x_0 + \sigma \epsilon$. DMs utilize a learnable denoising model $\mathbb{F}_\theta$ to minimize the denoising score matching objective:
\begin{align}
    & \hat{x}^t_0 = x_t - \sigma_t \mathbb{F}_\theta(x_t, t) \nonumber \\
    & \mathbb{E}_{x_0, \epsilon, t}[  || x_0 - \hat{x}^t_0||_2^2 ]
\label{eq: pred_x0}
\end{align}
The next step denoising input $x_{t-1}$ is derived as follows:
\begin{equation}
    x_{t-1} = x_t + (\sigma_{t-1} - \sigma_t) \mathbb{F}_\theta(x_t, t)
\label{eq: t_t-1}
\end{equation}
The denoising step iterates until the prediction $\hat{x}_0$ is obtained.

\subsection{Method Overview} \label{sec: pipeline_overview}

DMs are powerful tools for improving 3D reconstruction results due to their ability to hallucinate contents. VDMs are widely used for improving 3DGS \cite{kerbl20233d} because of the inherent capability to apply attention across frames, ensuring multi-frame consistency. However, the temporal attention mechanism also introduces a computational burden, 
which also limits the output length of VDMs, as the computation complexity is quadratic in relation to the sequence length. Furthermore, recent advanced VDMs \cite{yang2024cogvideox, kong2024hunyuanvideo, wan2025wan} utilize 3D VAE as their encoder and decoder, which performs temporal down-sampling, making it challenging to apply per-pixel confidence guidance.

Due to the above reasons, we select IDMs as the backbone in \paperName. However, most existing IDMs are not designed for the novel view synthesis task and do not take reference views as input. IP-Adapter \cite{ye2023ip} accepts image prompts as input, but it is intended for style prompts rather than novel view synthesis. Directly applying IDMs can lead to inconsistency across frames and finally result in blurriness in refined 3DGS. To tackle the problem, we propose an interleaved refining strategy, multi-level confidence guidance, and overall guidance. 

\subsection{Interleaved Refinement Strategy} \label{sec: interleave}

\boldparagraph{2D Refinement}
As mentioned in \cref{sec: preliminaries}, the trajectory of extrapolated views $\mc{T}_\it{ext} = \{\mc{V}^e_0, \mc{V}^e_1, ..., \mc{V}^e_m \}$ in our task definition is intended to be continuous. This continuous trajectory setting ensures that adjacent views $\mc{V}^e_i$ and $\mc{V}^e_{i+1}$ undergo only small transformations. A naive approach to keep consistency would be warping pixels from $\mc{V}^e_i$ to $\mc{V}^e_{i+1}$ and using DMs for inpainting. However, both rendered depth and predicted depth are not reliable for warping. Instead, we propose an interleaved refining strategy to enhance multi-view consistency.

Specifically, the refining process is interleaved and incremental along the trajectory $\mc{T}$. Given the current view $\mc{V}^e_i$, the current 3DGS $\mc{G}_{i-1}$ and rendered image $\hat{\mc{I}}^e_i = \pi(\mc{V}^e_i; \mc{G}_{i-1})$, we utilize denoising with guidance, as discussed in \cref{sec: guide_method}, to obtain the fixed image $\hat{\mc{I}}^{e,f}_i$. We also maintain a fixed image set $\mc{F}_{i-1} = \{ (\mc{V}_0^e, \hat{\mc{I}}^{e,f}_0), (\mc{V}_1^e, \hat{\mc{I}}^{e,f}_1), ..., (\mc{V}_{i-1}^e, \hat{\mc{I}}^{e,f}_{i-1}) \}$. We refine the current 3DGS $\mc{G}_{i-1}$ to $\mc{G}_i$ by using the training set $\mc{S}_{train}$, the previous refined view set $\mc{F}_{i-1}$ and the current refined image $\hat{\mc{I}}^{e,f}_i$.

\boldparagraph{3D Refinement}
The supervision during 3D Refinement for $\mc{G}_i$ comes from current refined view $(\mc{V}_{i}^e, \hat{\mc{I}}^{e,f}_{i})$, $\mc{F}_{i-1}$ and $S_{train}$. The detailed sampling strategy for training is illustrated in the supplements.

The generated results do not guarantee 3D consistency with training views, so we employ a smaller training loss for the generated views to prevent inaccurately generated areas from distorting 3D scenes. Additionally, the generated results exhibit slightly color bias compared to training views, which are often difficult for humans to distinguish. However, when applying the interleaved refining strategy, these slight color biases will accumulate, which may lead to a blurry and over-gray effect. We implement a simple yet efficient technique similar to \cite{zhou2024hugs} to tackle the problem. For each generated view, we define two optimizable affine matrices $\mathcal{A}_f \in \mathbb{R}^{3 \times 3}$ and $\mathcal{A}_b \in \mathbb{R}^{3 \times 1}$. The rendering results used for computing the training loss are applied to these affine matrices to avoid learning color bias:
\begin{align}
    & \hat{\mc{I}}^{e'} = \mathcal{A}_f \times \hat{\mc{I}^e} + \mathcal{A}_b \nonumber \\
    & \cL = (1 - \lambda_s) || \hat{\mc{I}}^{e'} - \hat{\cI}^{e,f} ||_1 + \lambda_s \textit{SSIM}(\hat{\mc{I}}^{'}, \hat{\cI}^{e,f})
\end{align}

\begin{figure}[t!]
     \centering
     \small 
     \setlength{\tabcolsep}{0pt}
     \def\mywidth{4.0cm}
     \begin{tabular}{P{\mywidth}P{\mywidth}}
     
     \includegraphics[width=\mywidth]{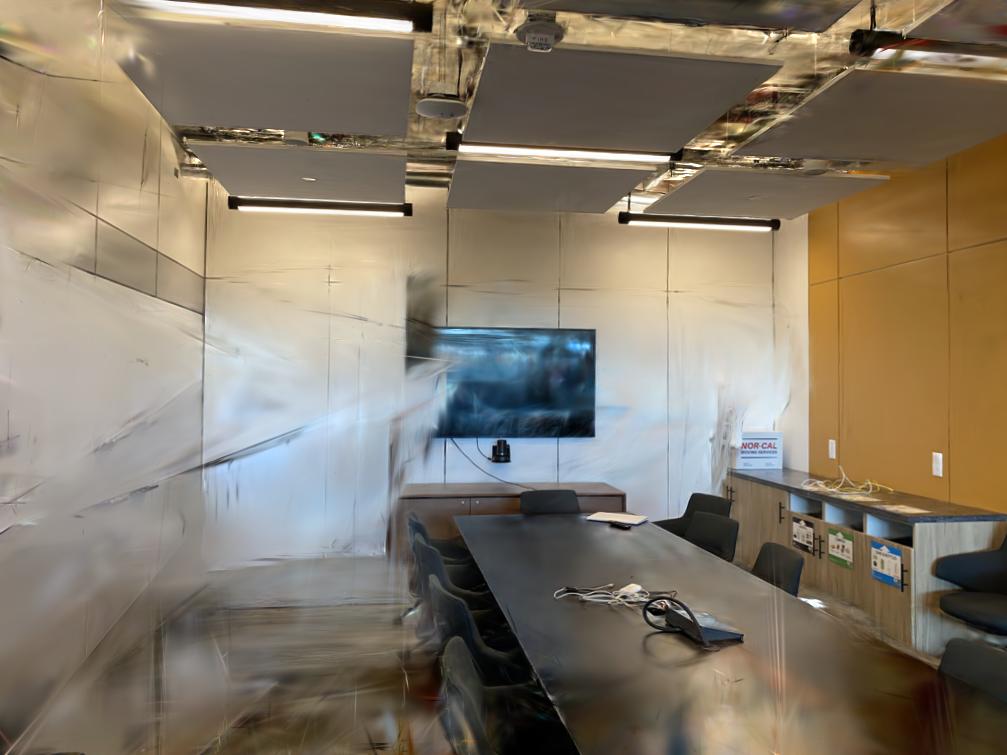}  &
     \includegraphics[width=\mywidth]{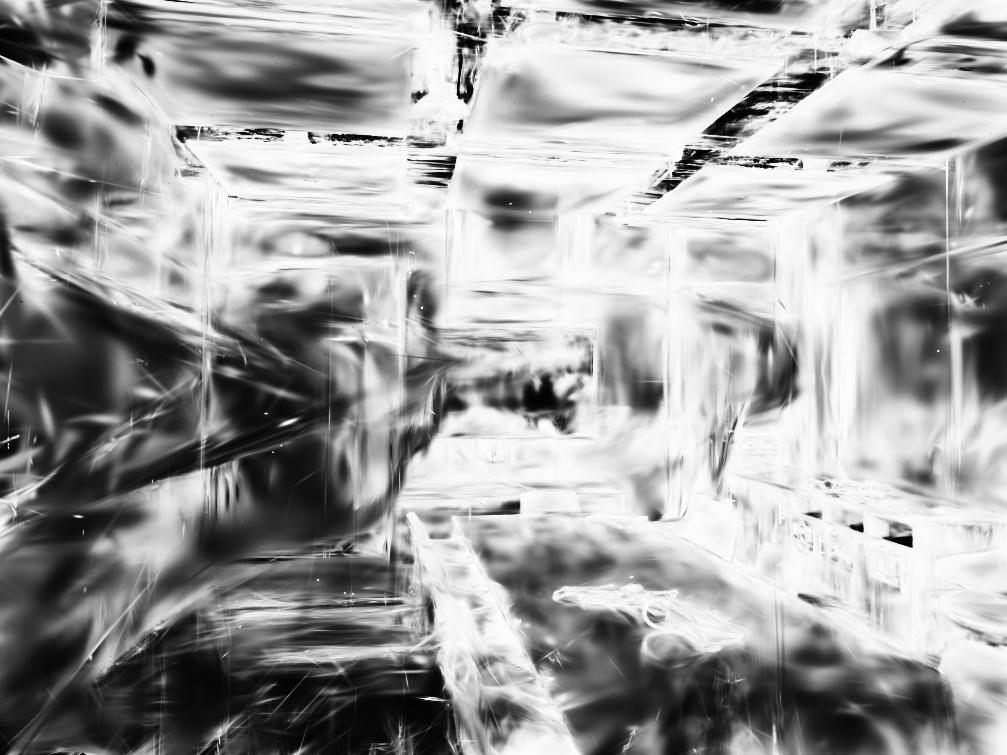}  \\

     {Rendered RGB w Artifacts} & {$\gamma_c=0.001$} \\
     
     \includegraphics[width=\mywidth]{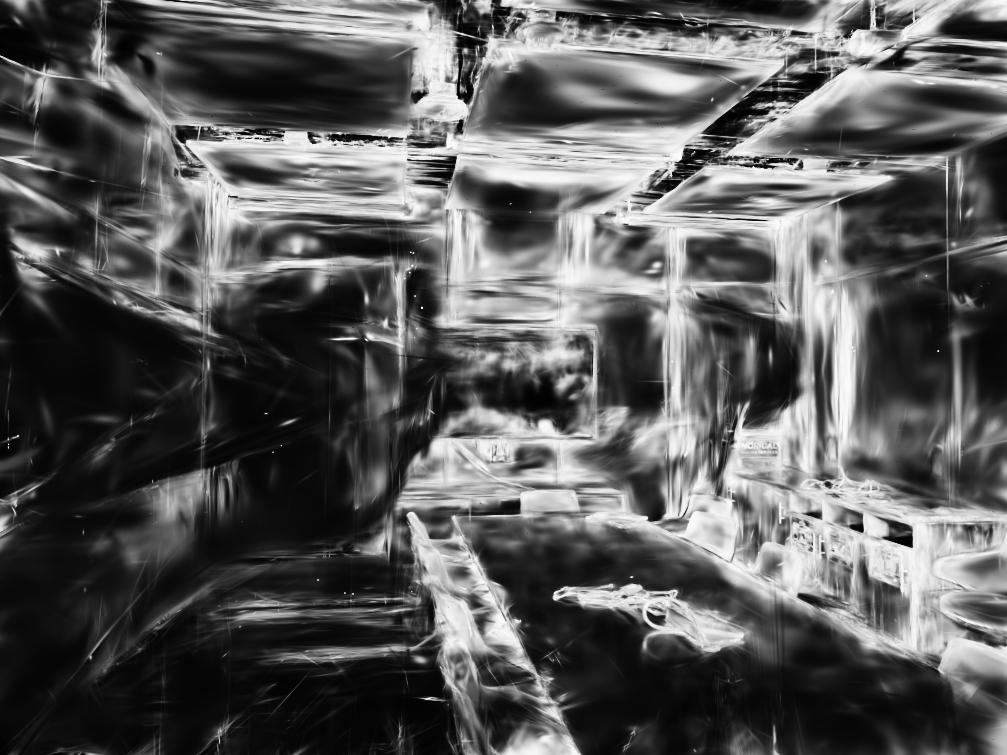} &
     \includegraphics[width=\mywidth]{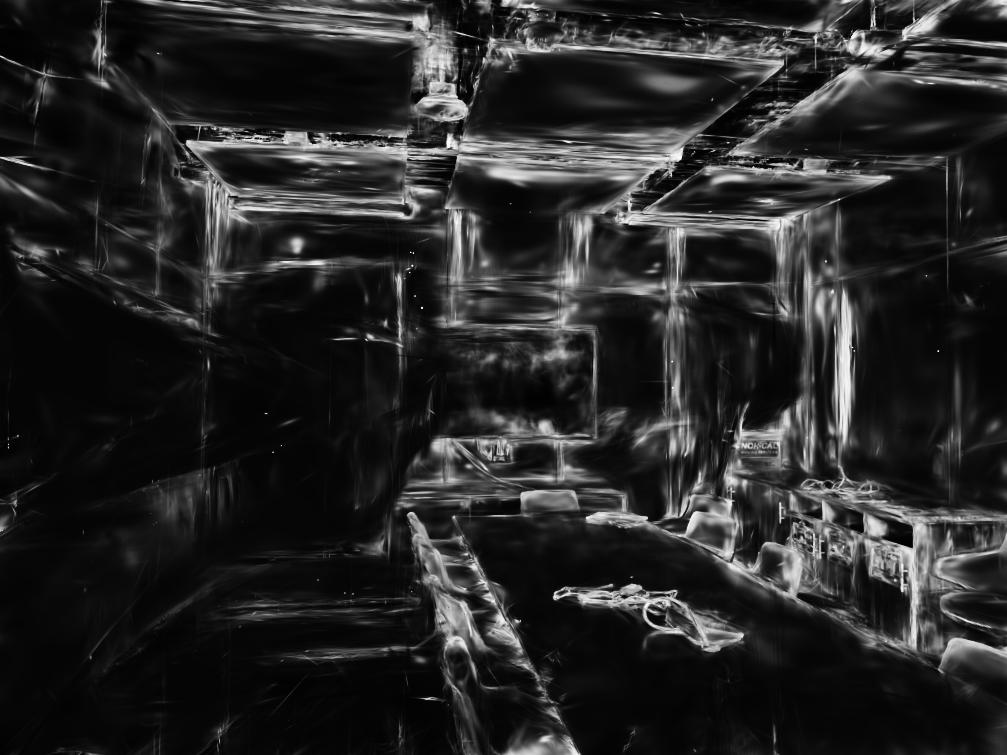} \\

     {$\gamma_c=0.01$} & {$\gamma_c=0.1$}

     \end{tabular}
     \vspace{-0.2cm}
     \caption{\textbf{Multi-Level Certainty Masks.} \paperName employs multiple $\gamma_c$ to obtain multi-level certainty masks as guidance. Each level of mask guides a different stage of denoising. A small $\gamma_c$ with high overall certainty is used for the early stages of denoising, while a large $\gamma_c$ which offers greater accuracy, is applied during the later stages of denoising.
     }
     \vspace{-0.6cm}
\label{fig: multi_level_mask}
\end{figure}

\subsection{Denoising with Guidance} \label{sec: guide_method}

Given the rendered results of an extrapolated view, even though the image contains artifacts, most areas can still be regarded as photo-realistic rendering results. These regions with relatively high fidelity can provide essential information for generating an image free of artifacts, while maintaining almost the same content.

Experiments in Difix3D+ \cite{wu2025difix3d+} have demonstrated that adding noise to images with artifacts and directly applying denoising using DMs can effectively remove these artifacts; however, the strength of the added noise is quite sensitive. For regions with significant artifacts, a larger scale of noise is needed to repaint those areas, while a smaller scale of noise is sufficient for areas with minimal artifacts. Although it may seem intuitive to apply different levels of noise to different regions, this approach does not align the data distribution of DMs. Instead, employing guidance during the diffusion denoising step is more practical and has been widely adopted in \cite{liu2024novel, you2024nvs}.

\boldparagraph{Confidence Map}
Utilizing appropriate guidance is an effective method for generating high-fidelity images while preserving accurate rendering results. However, current approaches that use warp masks or rendering opacities as guidance weights do not account for the presence of artifacts. For example, as illustrated in \cref{fig: masks_comp} (a), even when severe artifacts are present, the rendering opacities remain high, indicating that these artifacts continue to act as strong guidance during the denoising process.
To tackle this issue, we propose utilizing confidence masks as guidance weights, as shown in \cref{fig: masks_comp} (c). The confidence scores are derived from Fisher information, which is also referenced in \cite{jiang2024fisherrf, hanson2025pup}. Specifically, Fisher information measures the amount of information that the observation $(x,y)$ carries about the unknown parameters $w$ that model $p_f(y|x;w)$. In the context of novel view synthesis, Fisher information can be defined as:
\begin{equation}
    p_f(\pi(\mc{V}; \mc{G}) | \mc{V}; \mc{G})
\label{eq: fisher_def}
\end{equation}
where $\mc{V}$ and $\mc{G}$ represent viewpoint and 3DGS respectively, while $\pi(\mc{V}; \mc{G})$ denotes the volume rendering results at the specific view $\mc{V}$.

The negative log likelihood of Fisher information in \cref{eq: fisher_def}, which serves as the uncertainty $\mc{\bar{C}}_{\mc{V}; \mc{G}}$ of $\mc{G}$ at view $\mc{V}$, can be approximately derived as a Hessian matrix, the detailed derivation can be found in the supplementary materials:
\begin{align}
    \mc{\bar{C}}_{\mc{V}; \mc{G}} & = -\log p_f(\pi | \mc{V}; \mc{G}) \nonumber \\ 
    & = \bH^{''}[\pi|\mc{V}; \mc{G}] \nonumber \\ 
    & = \nabla_{\mc{G}}\pi(\mc{V};\mc{G})^T\nabla_{\mc{G}}\pi(\mc{V};\mc{G})
\label{eq: uncertainty}
\end{align}

\cite{jiang2024fisherrf, hanson2025pup} renders the attribute $\mc{\bar{C}}_{\mc{V}; \mc{G}}$ in volume rendering to obtain the uncertainty map. However, uncertainty is not a numerically stable representation, as its value can range from $[0, +\infty)$. As illustrated in \cref{fig: masks_comp} (b), the numeric instability of uncertainty may render an inaccurate uncertainty map. This often occurs when there are Gaussians with low opacity and high uncertainty, which can \textit{overwhelm} the volume rendering. Instead, we use the complementary value as guidance, certainty $\mc{C}_{\mc{V}; \mc{G}}$, also referred to as confidence in this paper, which has a stable numeric range of $[0, 1]$.
The certainty $\mc{C}_{\mc{V}; \mc{G}}^{\gamma_c}$ is defined as:
\begin{equation}
    \mc{C}_{\mc{V}; \mc{G}}^{\gamma_c} = \exp[-\gamma_c\mc{\bar{C}}_{\mc{V}; \mc{G}}]
\label{eq: certainty}
\end{equation}
where $\gamma_c$ is a hyperparameter. When $\gamma_c = 1$, we actually use the original Fisher information as the confidence. When render $\mc{C}_{\mc{V}; \mc{G}}$ with hyperparameter as an attribute in 3DGS, and multiply with rendered opacity $\mc{M}^{\alpha}$, we obtain the confidence map $\mc{M}_{\mc{V}; \mc{G}}^{\gamma_c}$:
\begin{align}
    \mc{M}^{\alpha} & = \pi(\mc{V};(\mc{G}, \alpha)) \nonumber \\
    \mc{M}_{\mc{V}; \mc{G}}^{\gamma_c} & = \pi(\mc{V};(\mc{G}, \mc{C}_{\mc{V}; \mc{G}}^{\gamma_c})) \odot \mc{M}^{\alpha}
\label{eq: certainty_map}
\end{align}

\begin{figure*}[t!]
     \centering
     \small 
     \setlength{\tabcolsep}{0pt}
     \def\mywidth{3.55cm}
     \begin{tabular}{P{0.5cm}P{\mywidth}P{\mywidth}P{\mywidth}P{\mywidth}P{\mywidth}}
     
     \rotatebox{90}{~~~~LLFF / Fortress} &
     \includegraphics[width=\mywidth]{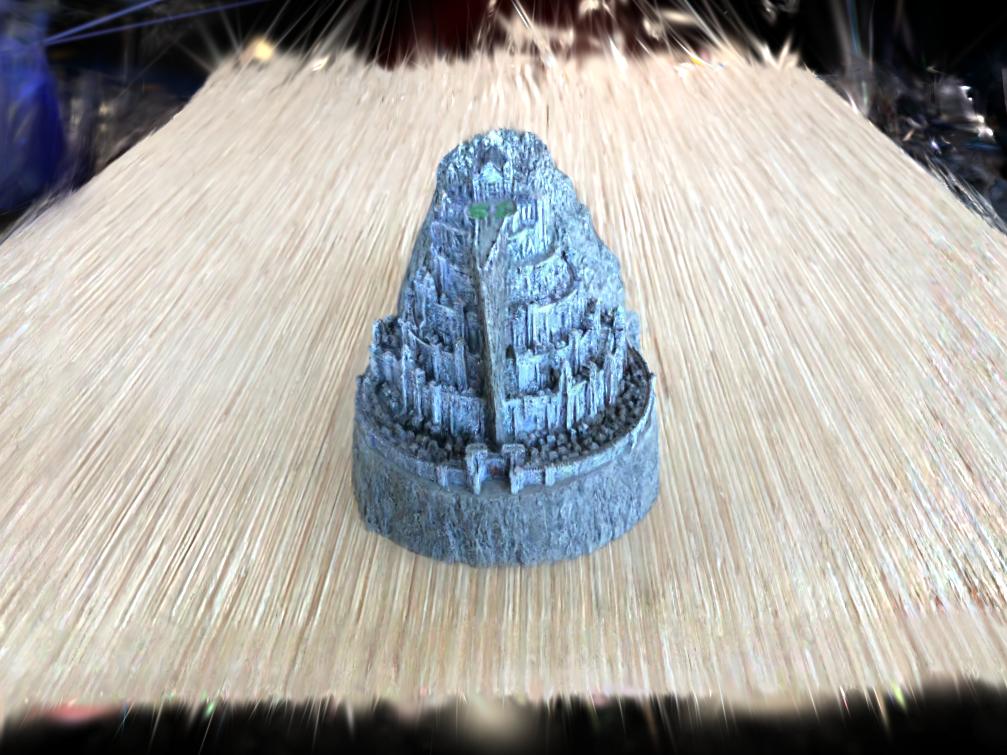}  &
     \includegraphics[width=\mywidth]{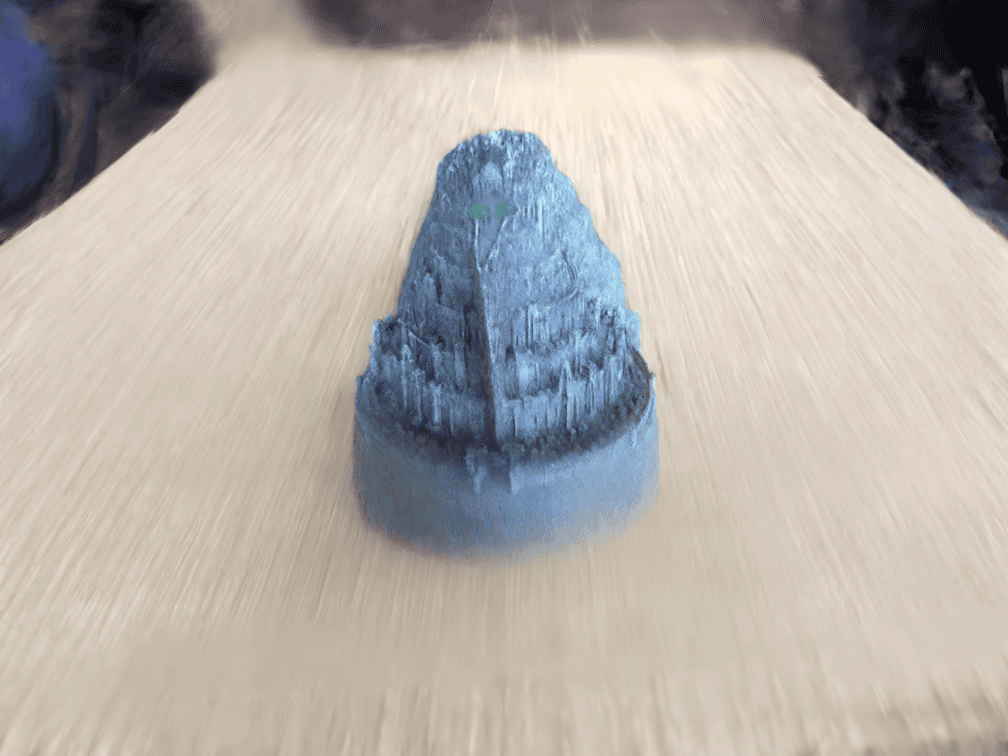} &
     \includegraphics[width=\mywidth]{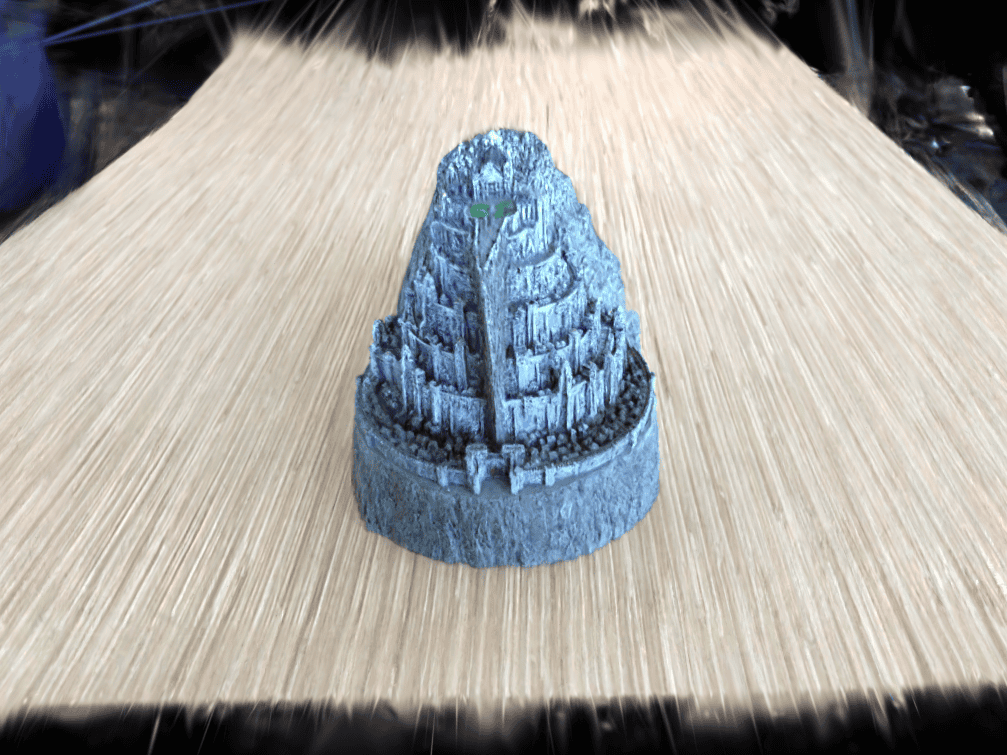} &
     \includegraphics[width=\mywidth]{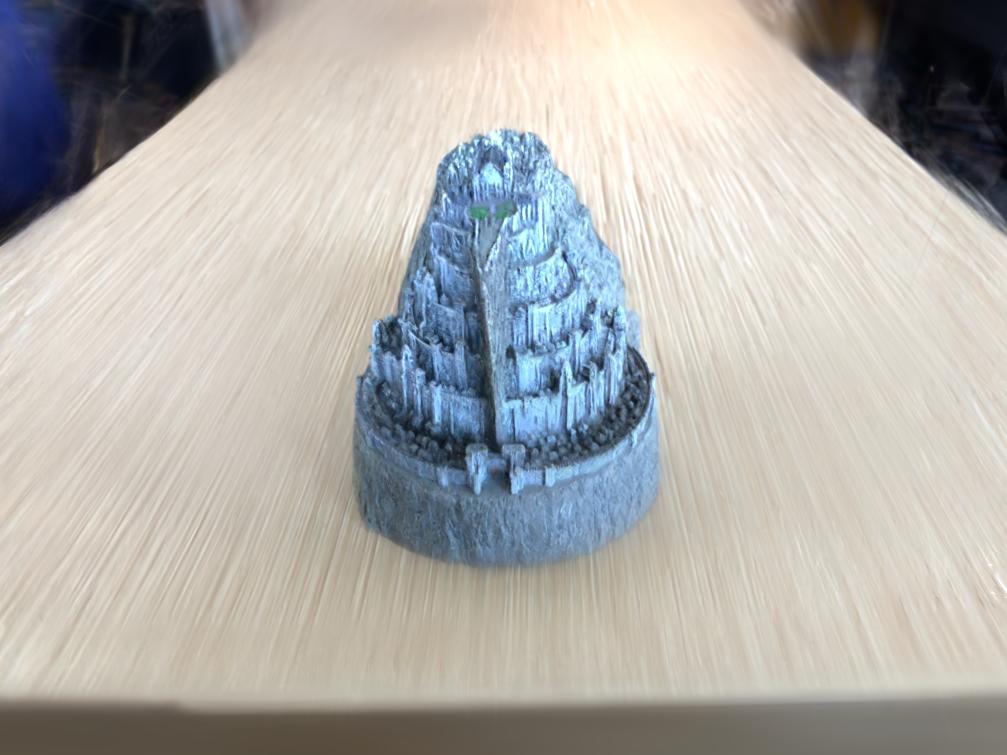} &
     \includegraphics[width=\mywidth]{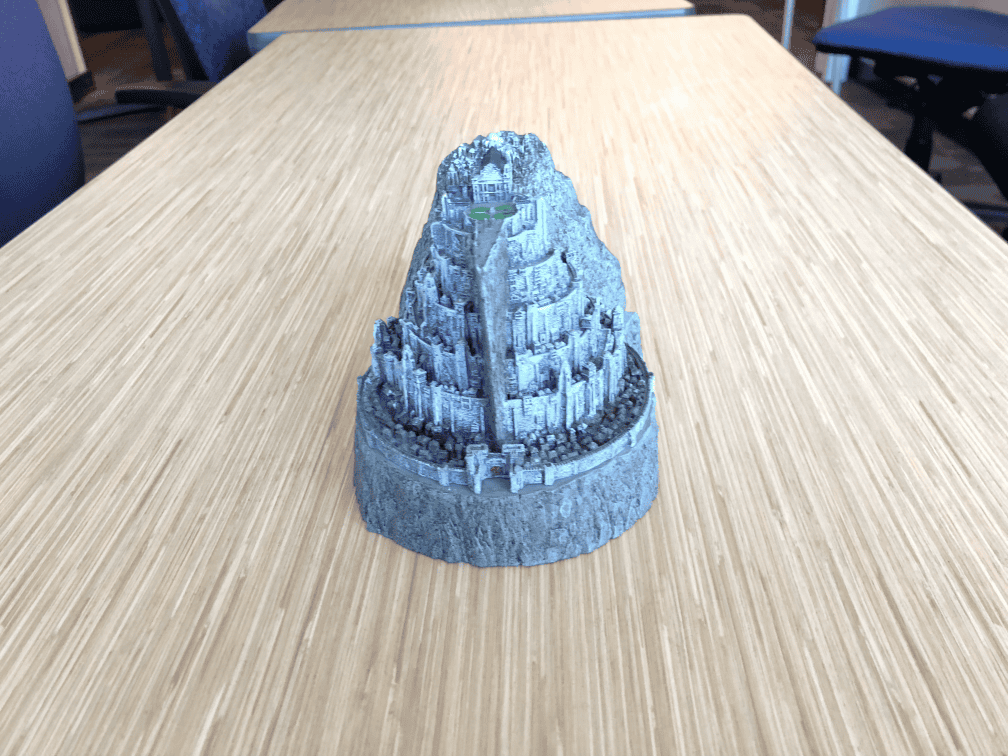} \\

     \rotatebox{90}{~~~~~LLFF / Leaves} &
     \includegraphics[width=\mywidth]{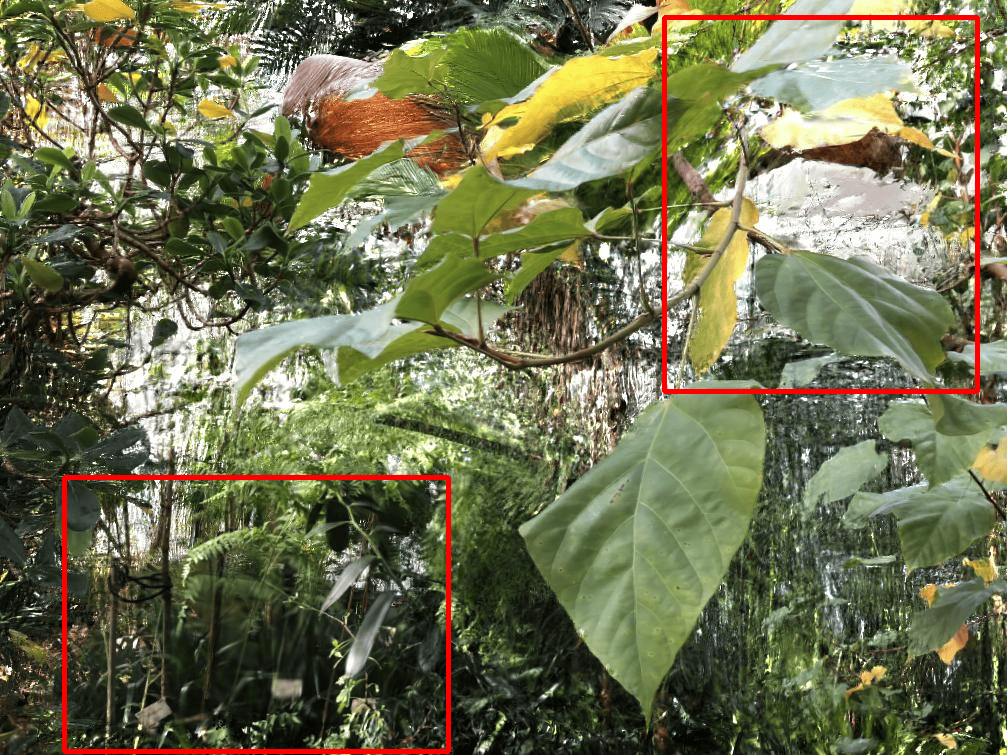}  &
     \includegraphics[width=\mywidth]{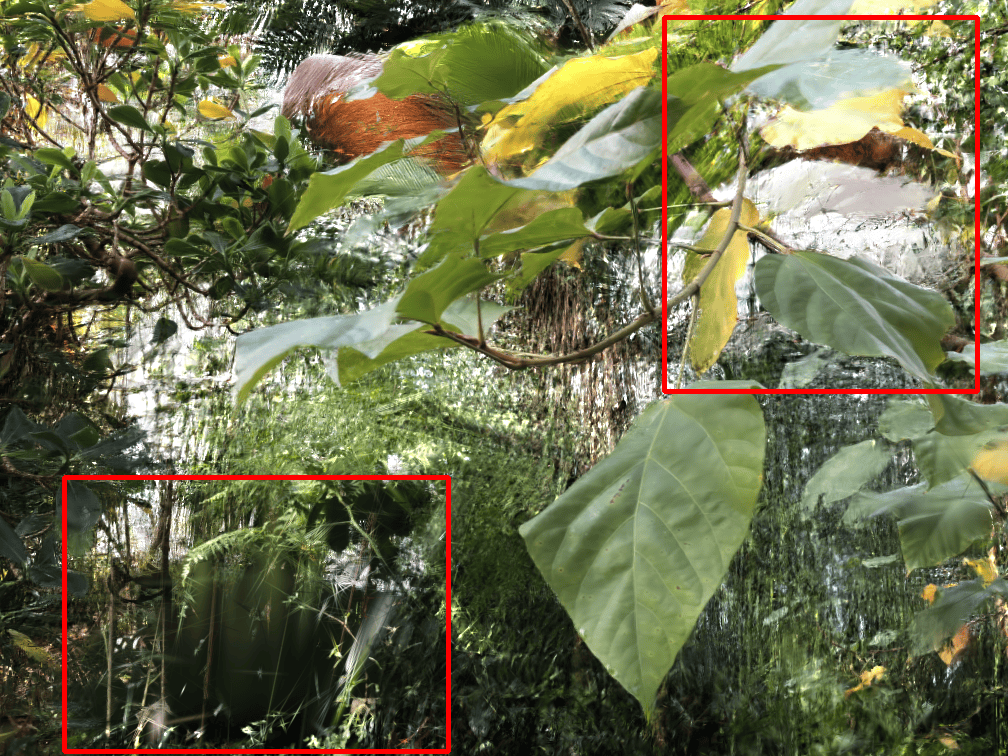} &
     \includegraphics[width=\mywidth]{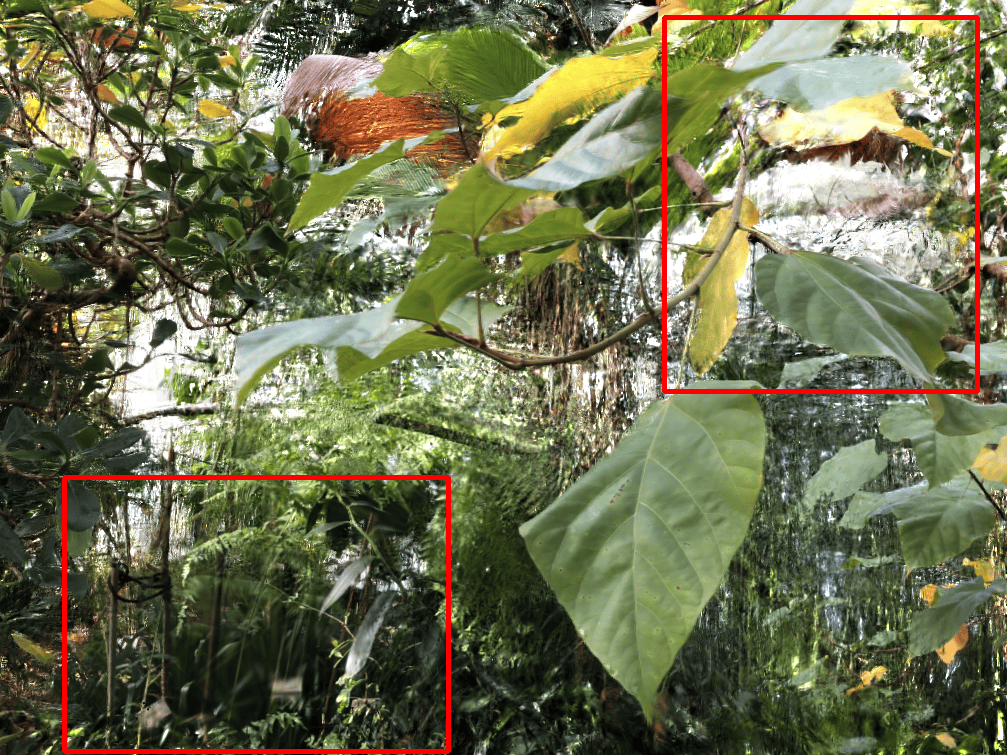} &
     \includegraphics[width=\mywidth]{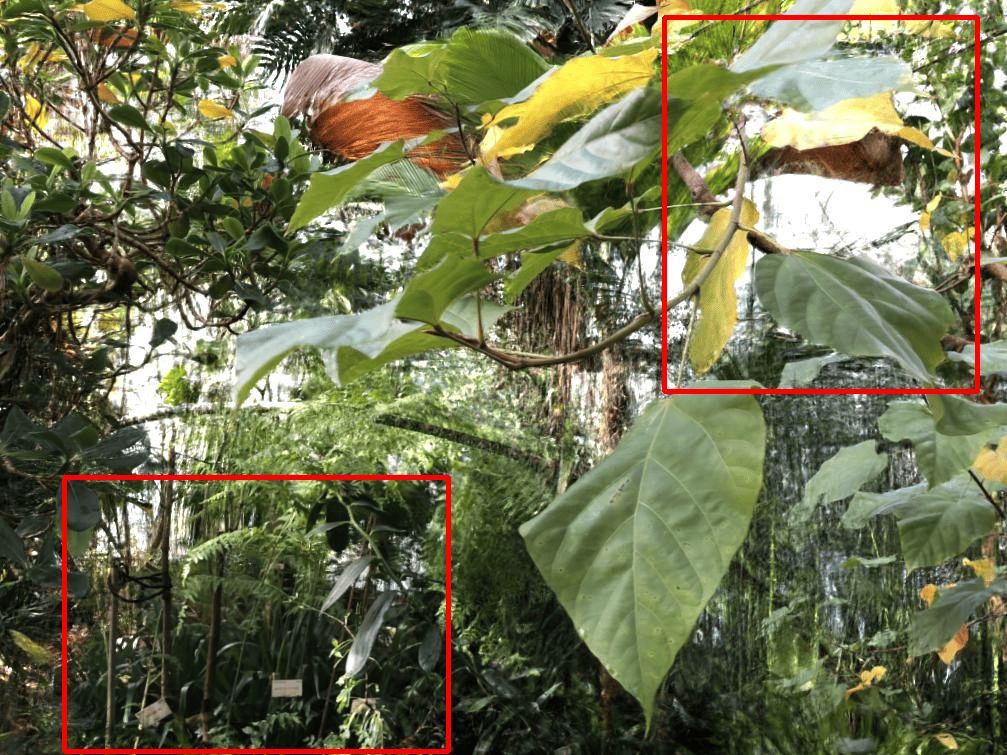} &
     \includegraphics[width=\mywidth]{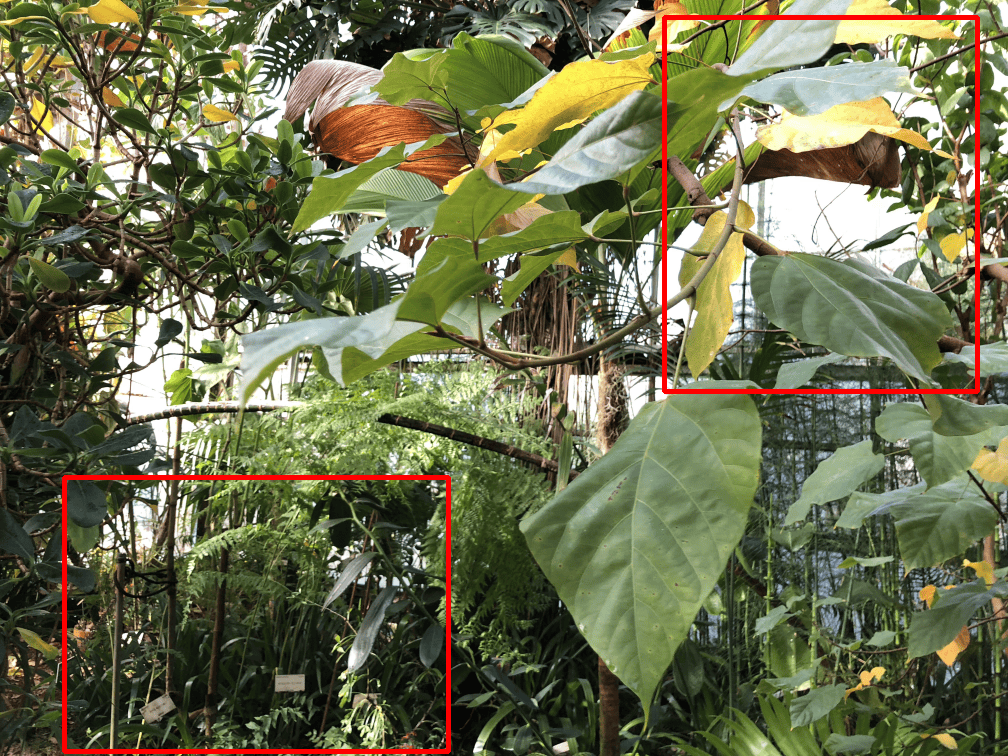} \\

     \rotatebox{90}{~~~~Mip / Kitchen} &
     \includegraphics[width=\mywidth]{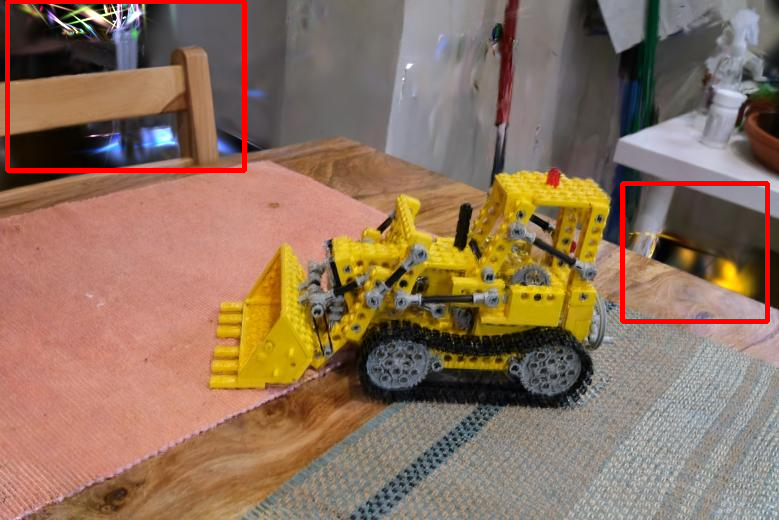}  &
     \includegraphics[width=\mywidth]{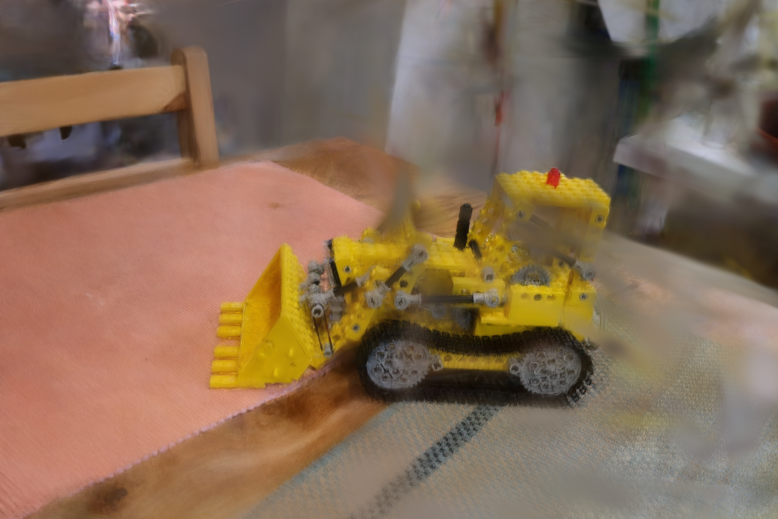} &
     \includegraphics[width=\mywidth]{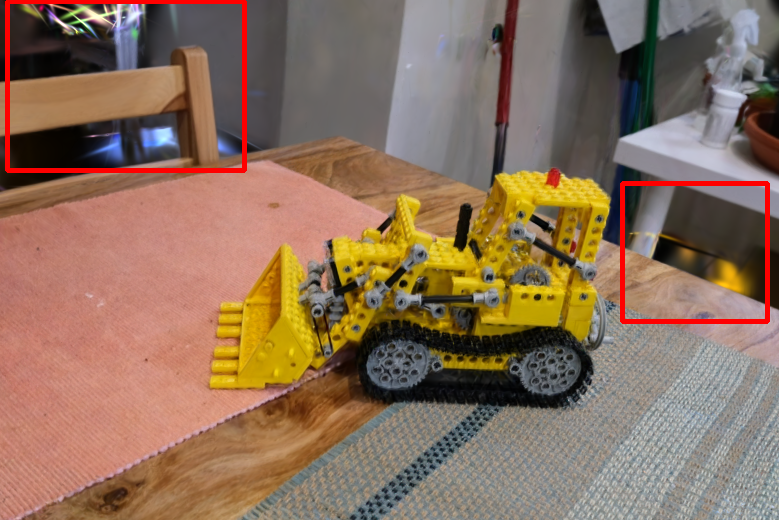} &
     \includegraphics[width=\mywidth]{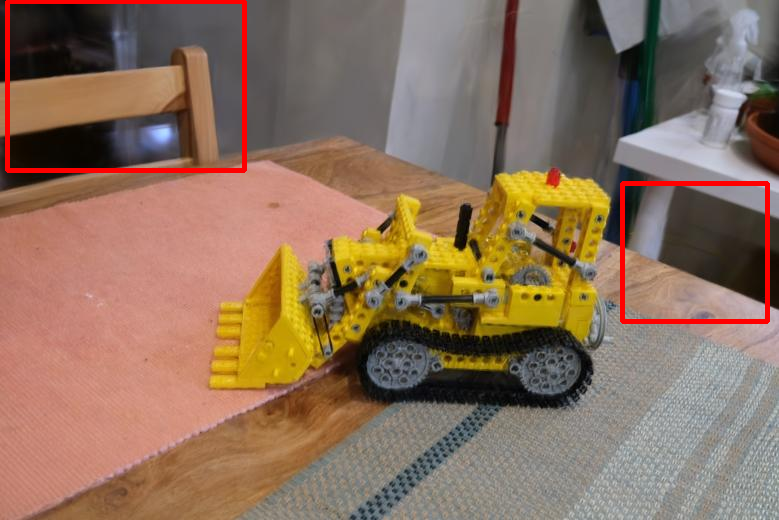} &
     \includegraphics[width=\mywidth]{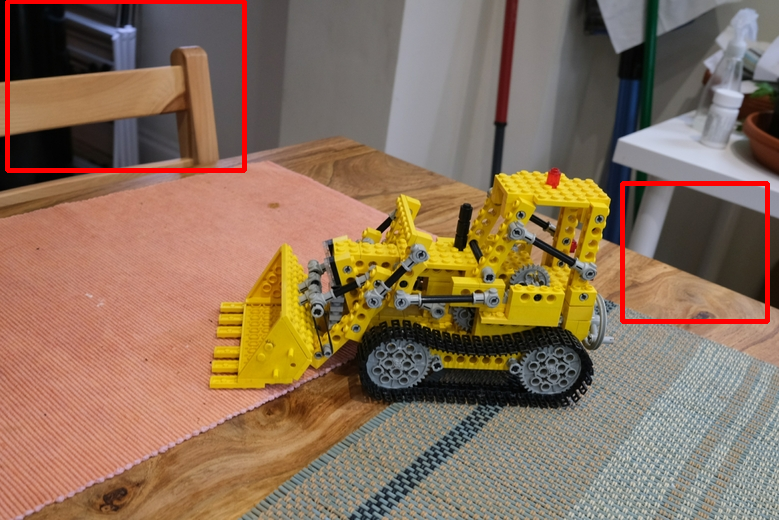} \\

     \rotatebox{90}{~~~~Mip / Garden} &
     \includegraphics[width=\mywidth]{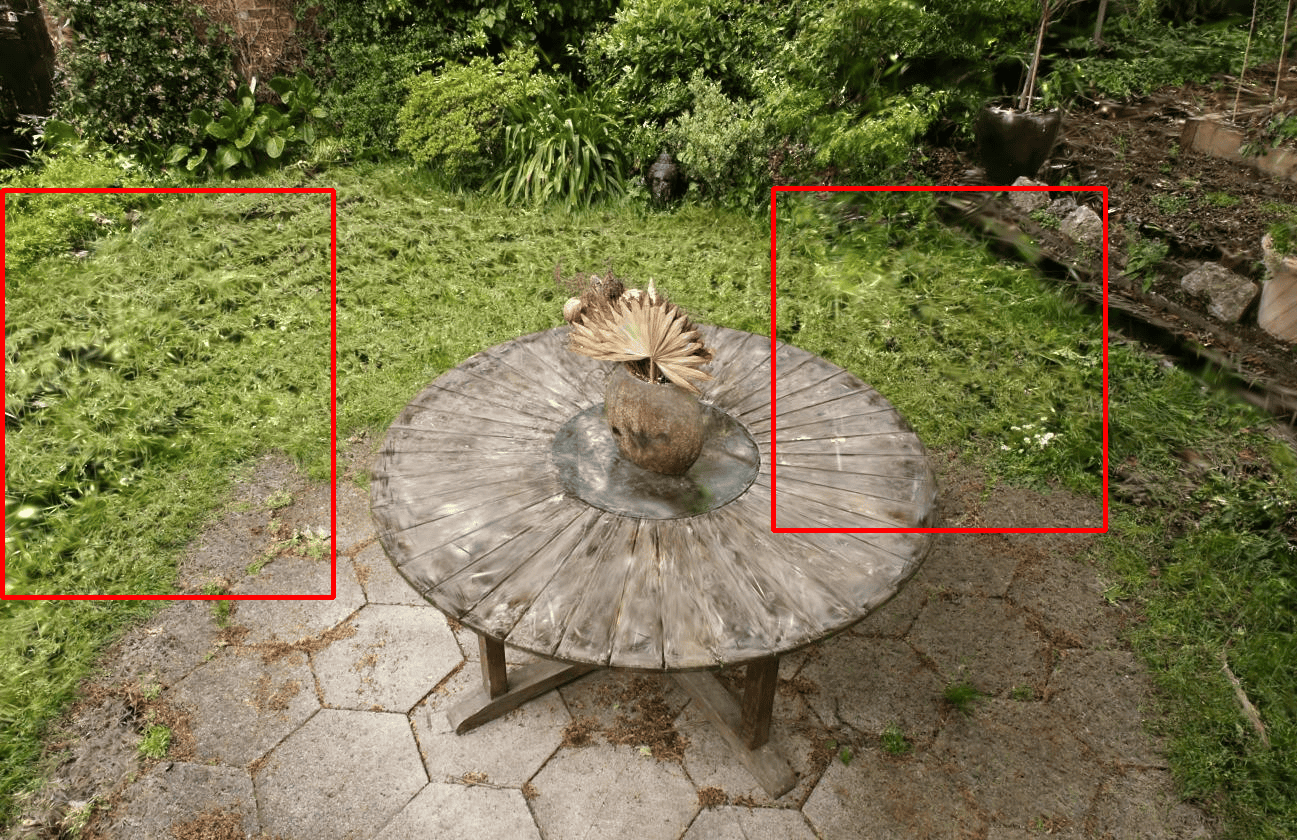}  &
     \includegraphics[width=\mywidth]{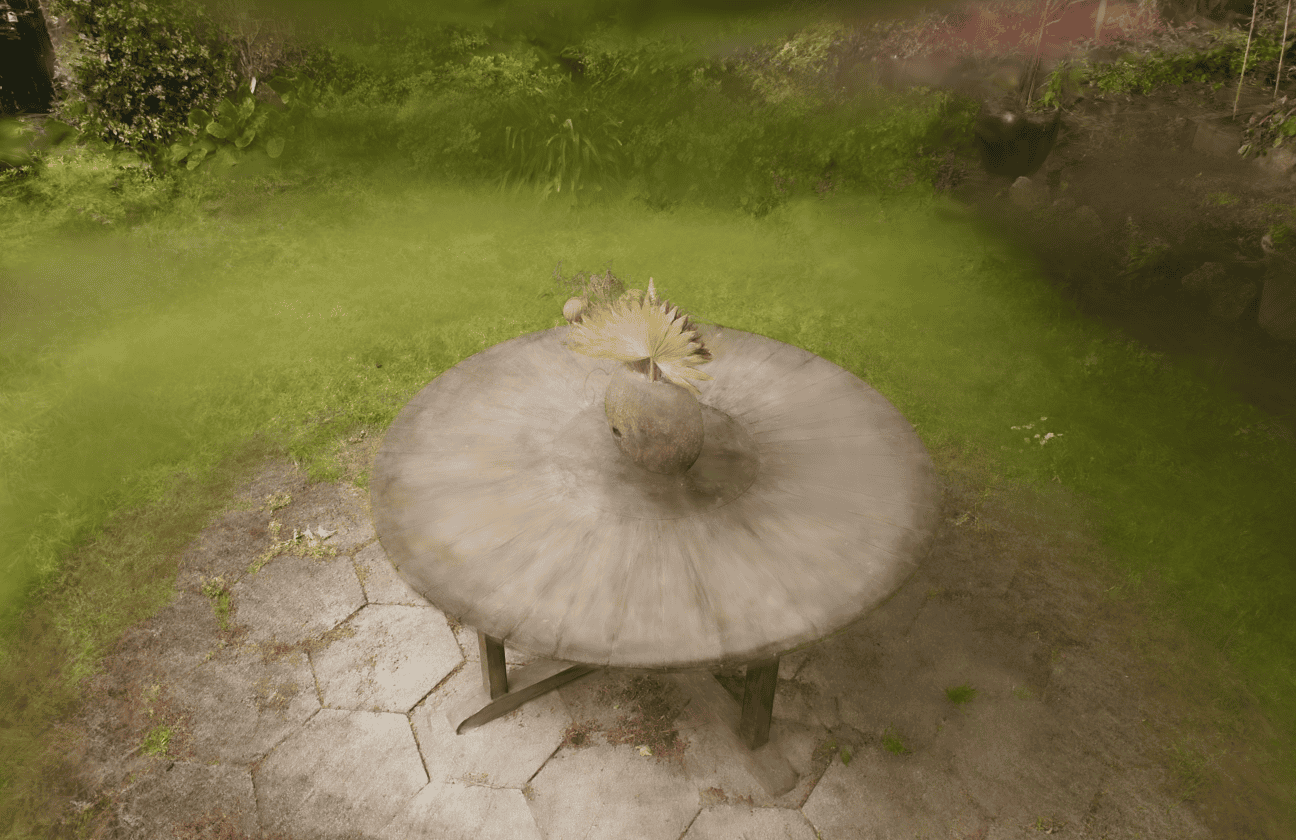} &
     \includegraphics[width=\mywidth]{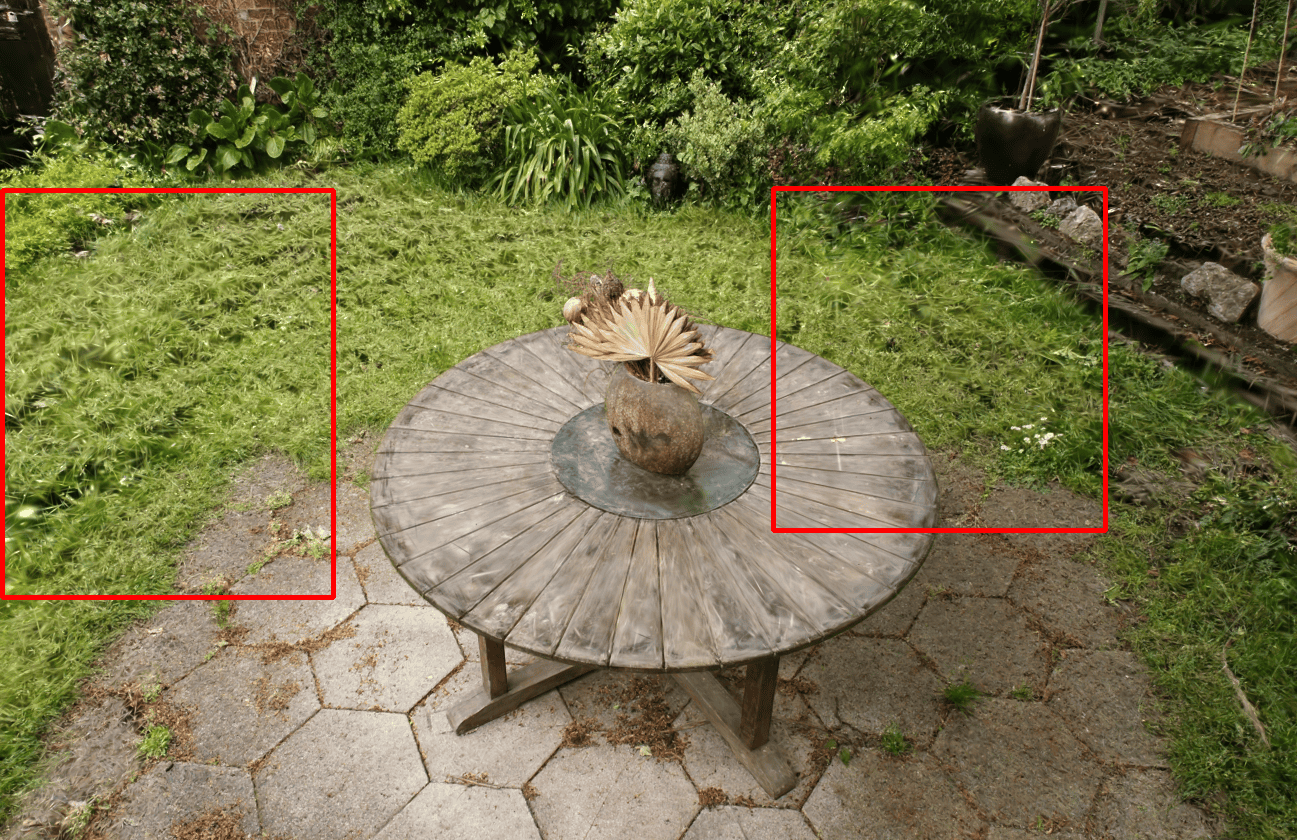} &
     \includegraphics[width=\mywidth]{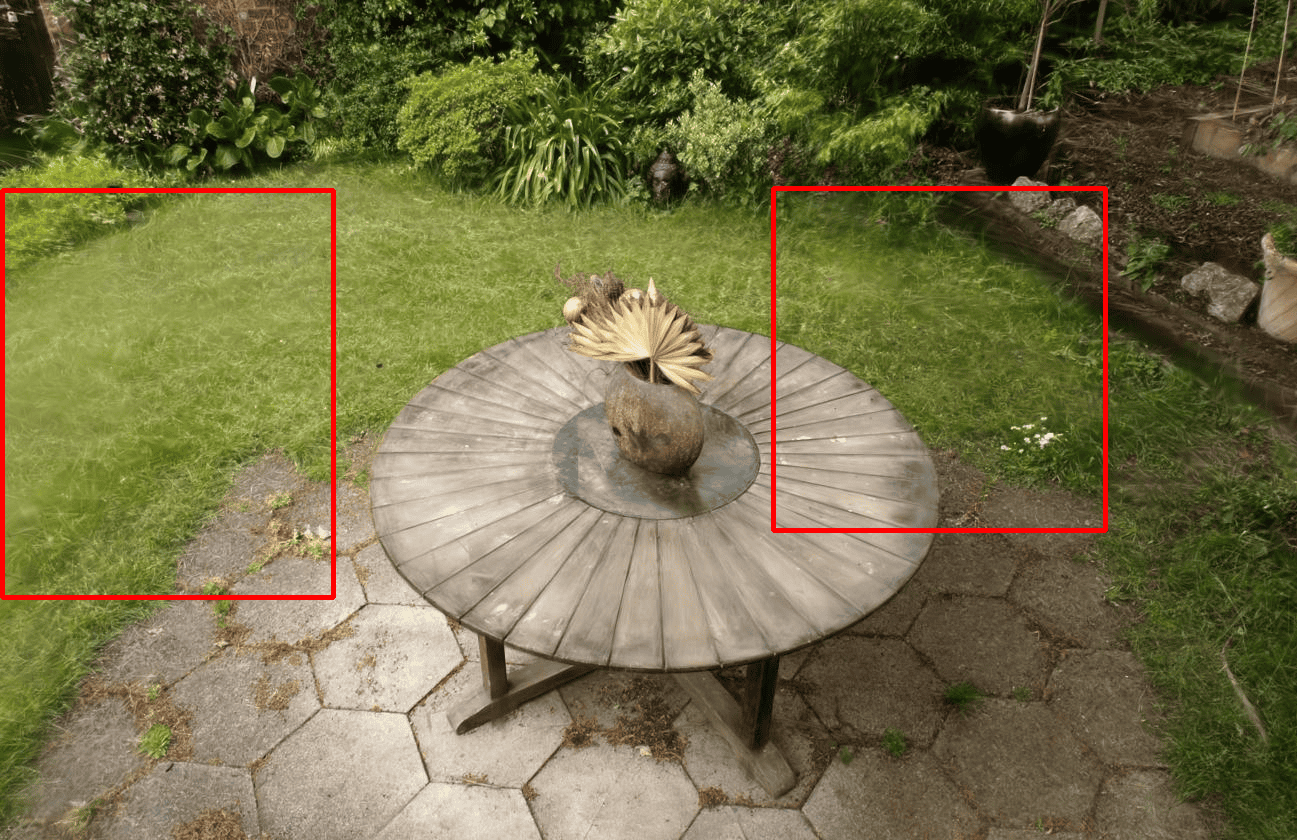} &
     \includegraphics[width=\mywidth]{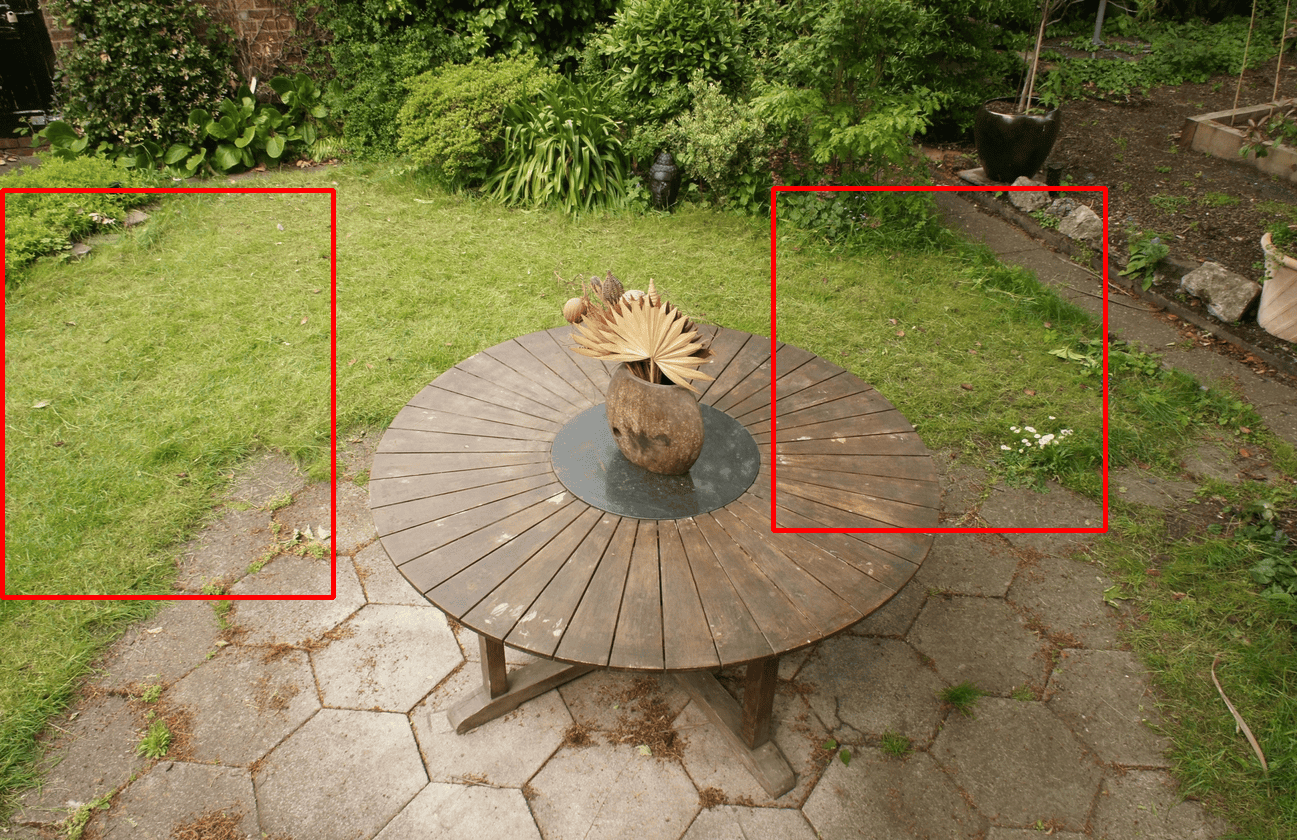} \\

     {} & {3DGS} & {ViewExtrapolator \cite{liu2024novel}} & {Difix3D+\cite{wu2025difix3d+}} & \paperName & Ground Truth
     
     \end{tabular}
     \vspace{-0.2cm}
     \caption{\textbf{Qualitative Comparisons on LLFF \cite{mildenhall2019llff} and Mip-NeRF 360 \cite{barron2021mip}.} \paperName demonstrates state-of-the-art performance on these two datasets. 
     }
     \vspace{-0.4cm}
\label{fig: llff_mip_comp}
\end{figure*}

\boldparagraph{Multi-Level Confidence Maps}
As shown in \cref{fig: multi_level_mask}, $\gamma_c$ is a hyperparameter that controls sensitivity to artifacts when rendering confidence maps. The larger the value of $\gamma_c$, the more sensitive the rendered confidence map becomes to artifacts. Selecting a single appropriate $\gamma_c$ is not trivial. Therefore, we apply multi-level confidence maps as guidance. Since DMs generate a coarse structure of image rather than detailed appearance in the early denoising stages \cite{shaulov2025flowmo}, we provide $\mc{M}_{\mc{V}; \mc{G}}^{\gamma_c}$ with a small $\gamma_c$ to offer more comprehensive guidance. In the later denoising stages, DMs tend to generate detailed appearances, so we provide $\mc{M}_{\mc{V}; \mc{G}}^{\gamma_c}$ with a large $\gamma_c$ to ensure that the guidance is sufficiently accurate.

\boldparagraph{Confidence Guidance}
Given the rendered image $\hat{I}_{\mc{V}; \mc{G}}$ and the corresponding confidence map $\mc{M}_{\mc{V}; \mc{G}}^{\gamma_c}$, we can provide denoising guidance to DMs.
We denote the rendered image after VAE encoding as $x_0^r$, and the resized confidence map that aligns with the shape of the latent space as $\mc{M}^c$. As illustrated in \cref{eq: pred_x0}, the predicted $x_0^t$ at $t$ timestep is given by $x_t - \sigma_t \mathbb{F}_\theta(x_t, t)$. We guide the model prediction as $x_0^{t,g}$ by blending the rendered image using confidence mask:
\begin{equation}
    x_0^{t,g} = \mc{M}^c \odot x_0^r + (1 - \mc{M}^c) \odot x_0^t
\end{equation}
However, the input for the next denoising step cannot be directly obtained using \cref{eq: t_t-1} since the model prediction $x_0^{t}$ has been changed. Instead, we derive the new $x_{t-1}$ by solving the following equations:
\begin{align}
    x_{t-1} &= x_0 + \sigma_{t-1} \mathbb{F}_\theta(x_t, t) \nonumber \\
    x_{t-1} &= x_t + (\sigma_{t-1} - \sigma_t) \mathbb{F}_\theta(x_t, t)
\end{align}
The representation of $x_{t-1}$ derived from $x_0^{t,g}$ and $x_t$ is:
\begin{equation}
    x_{t-1} = \frac{\sigma_{t-1}}{\sigma_t} x_t - \frac{\sigma_{t-1} - \sigma_t}{\sigma_t} x_0^{t,g}
\end{equation}

\boldparagraph{Overall Guidance} 
Although the interleaved refining strategy provides higher fidelity rendering results and ensures that the rendering is more consistent with the generated content, using IDMs may still encounter issues of inconsistency in areas with low confidence. Particularly in regions with weak textures like ground and sky, the confidence map tends to be low, and allowing denoising to proceed freely in these areas can result in high inconsistency and blurriness in 3DGS. To address this issue, we propose an overall guidance approach, which combines confidence guidance in the very early stages of denoising to provide structural hints for the images.
The combination of certainty and overall guidance is defined as follows:
\begin{align}
    x_0^{t,g} =& \mc{M}^c \odot x_0^r + \nonumber \\
    &(1- \mc{M}^c) \odot (\beta \mc{M}^\alpha x_0^r + (1-\beta \mc{M}^\alpha) x_0^t )
\end{align}
where $\beta$ is a hyperparameter that controls the strength of the overall guidance.

\definecolor{leafgreen}{rgb}{0.05, 0.54, 0.25}
\definecolor{orange}{rgb}{1.0, 0.6, 0.0}
\def \tick{\color{leafgreen}{\ding{52}}}
\def \cross{\color{red}{\ding{56}}}

\definecolor{tabfirst}{rgb}{1, 0.7, 0.7} %
\definecolor{tabsecond}{rgb}{1, 0.85, 0.7} %
\definecolor{tabthird}{rgb}{1, 1, 0.7} %

\begin{table*}[]
\centering
\small
\setlength{\tabcolsep}{2pt}
\begin{tabular}{@{\extracolsep{0pt}}lcccccccccccccc@{}} 
\toprule
\multicolumn{1}{c}{} & \multicolumn{3}{c}{LLFF \cite{mildenhall2019llff}} & \multicolumn{3}{c}{Mip-NeRF 360 \cite{barron2021mip}} & \multicolumn{1}{c}{Waymo} \cite{sun2020scalability} & \multirow{2}{*}{DM Type} & \multirow{2}{*}{w/o Finetune} & \multirow{2}{*}{Only RGBs} &
\multirow{2}{*}{3D Render} \\
& PSNR$\uparrow$  & SSIM$\uparrow$  & LPIPS$\downarrow$  & PSNR$\uparrow$  & SSIM$\uparrow$  & LPIPS$\downarrow$ & KID$\downarrow$ & & & \\
\midrule
3DGS \cite{kerbl20233d} & 18.10 & 0.633 & 0.265 & 21.83 & 0.643 & 0.239 & 0.155 & N/A & N/A & \tick & \tick \\

\midrule

\paperName + SDXL & \cellcolor{tabsecond}19.93 & \cellcolor{tabsecond}0.695 & \cellcolor{tabsecond}0.237 & \cellcolor{tabsecond}22.68 & \cellcolor{tabsecond}0.685 & \cellcolor{tabthird}0.213 & \cellcolor{tabthird}0.150 & Image & \tick & \tick & \tick \\
\paperName + Flux & \cellcolor{tabfirst}20.12 & \cellcolor{tabfirst}0.700 & \cellcolor{tabfirst}0.221 & \cellcolor{tabfirst}23.02 & \cellcolor{tabfirst}0.689 & \cellcolor{tabfirst}0.208 & \cellcolor{tabsecond}0.147 & Image & \tick & \tick & \tick \\
ViewExtrapolator \cite{liu2024novel} & 18.27 & 0.614 & 0.338 & 20.84 & 0.591 & 0.332 & 0.180 & Video & \tick & \tick & \tick \\
NVS-Solver \cite{you2024nvs} & 11.99 & 0.351 & 0.560 & 12.45 & 0.266 & 0.631 & 0.289 & Video & \tick & \tick & \cross \\

\midrule

Difix3D+ \cite{wu2025difix3d+} & \cellcolor{tabthird}18.86 & \cellcolor{tabthird}0.658 & \cellcolor{tabthird}0.239 & \cellcolor{tabthird}22.43 & \cellcolor{tabthird}0.661 & \cellcolor{tabsecond}0.210 & \cellcolor{tabfirst}0.143 & Image & \cross & \tick & \tick \\
StreetCrafter \cite{yan2025streetcrafter} & N/A & N/A & N/A & N/A & N/A & N/A & 0.157 & Video & \cross & \cross & \tick \\

\bottomrule
\end{tabular}
\vspace{-0.2cm}
\caption{\textbf{Quantitative Comparison with Baselines.} \paperName demonstrates superior performance among baselines without fine-tuning. Compared to models that require fine-tuning, \paperName providing better results on LLFF and Mip-NeRF 360, while achieving comparable performance on Waymo. {\colorbox{tabfirst}{First}}, {\colorbox{tabsecond}{second}}, and {\colorbox{tabthird}{third}} performances in each column are indicated by their respective colors.}

\vspace{-0.3cm}
\label{tab: main_comp}
\end{table*}

\section{Experiments}

\begin{figure}
     \centering
     \small 
     \setlength{\tabcolsep}{0pt}
     \def\mywidth{2.9cm}
     \begin{tabular}{P{\mywidth}P{\mywidth}P{\mywidth}}

     \includegraphics[width=\mywidth]{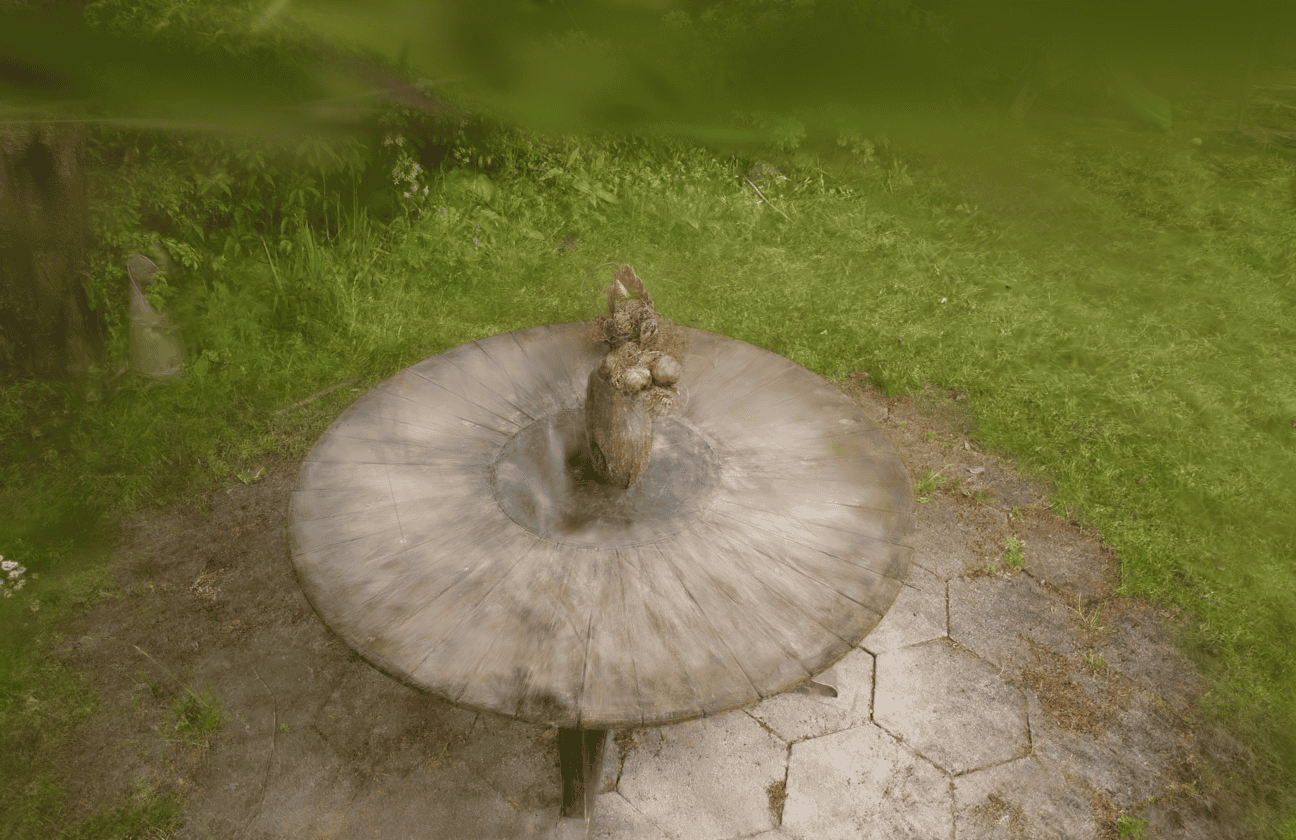}  &
     \includegraphics[width=\mywidth]{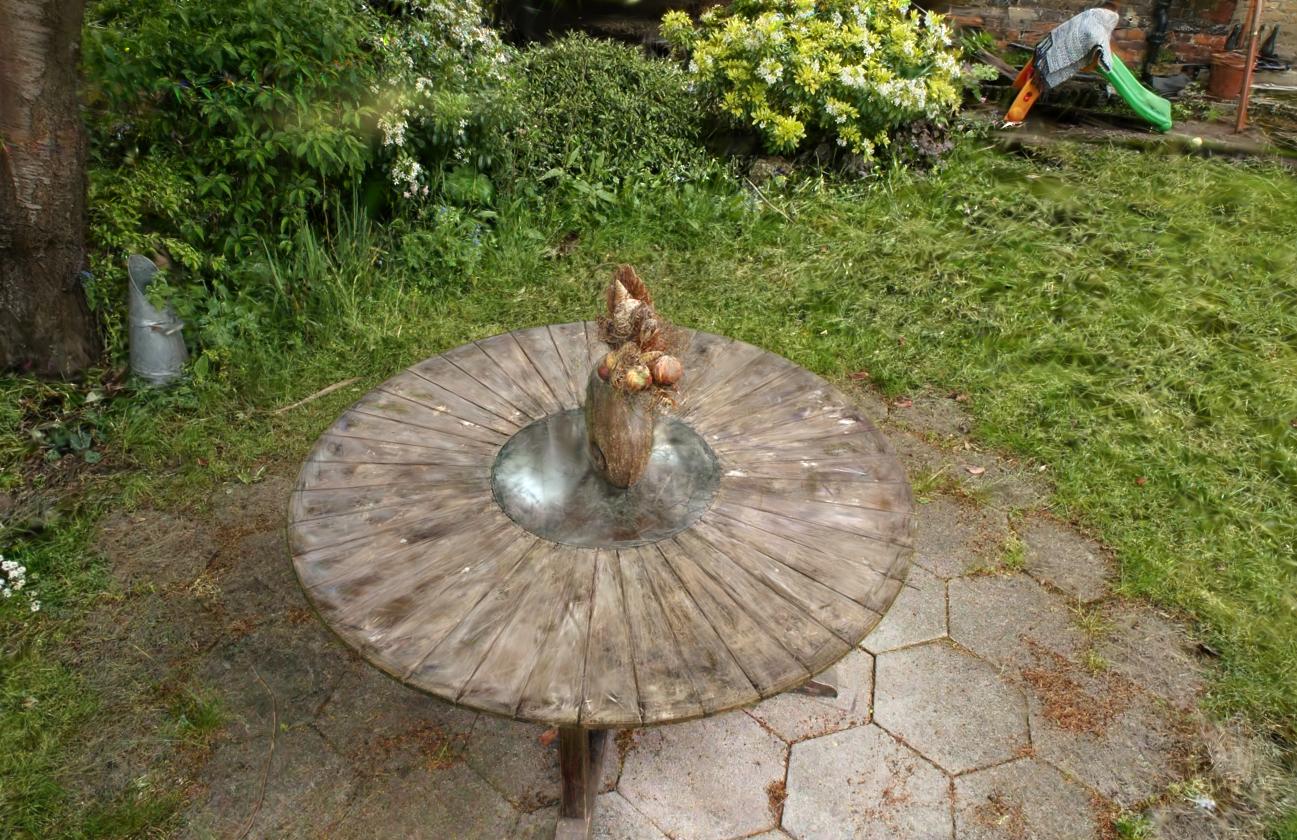}  &
     \includegraphics[width=\mywidth]{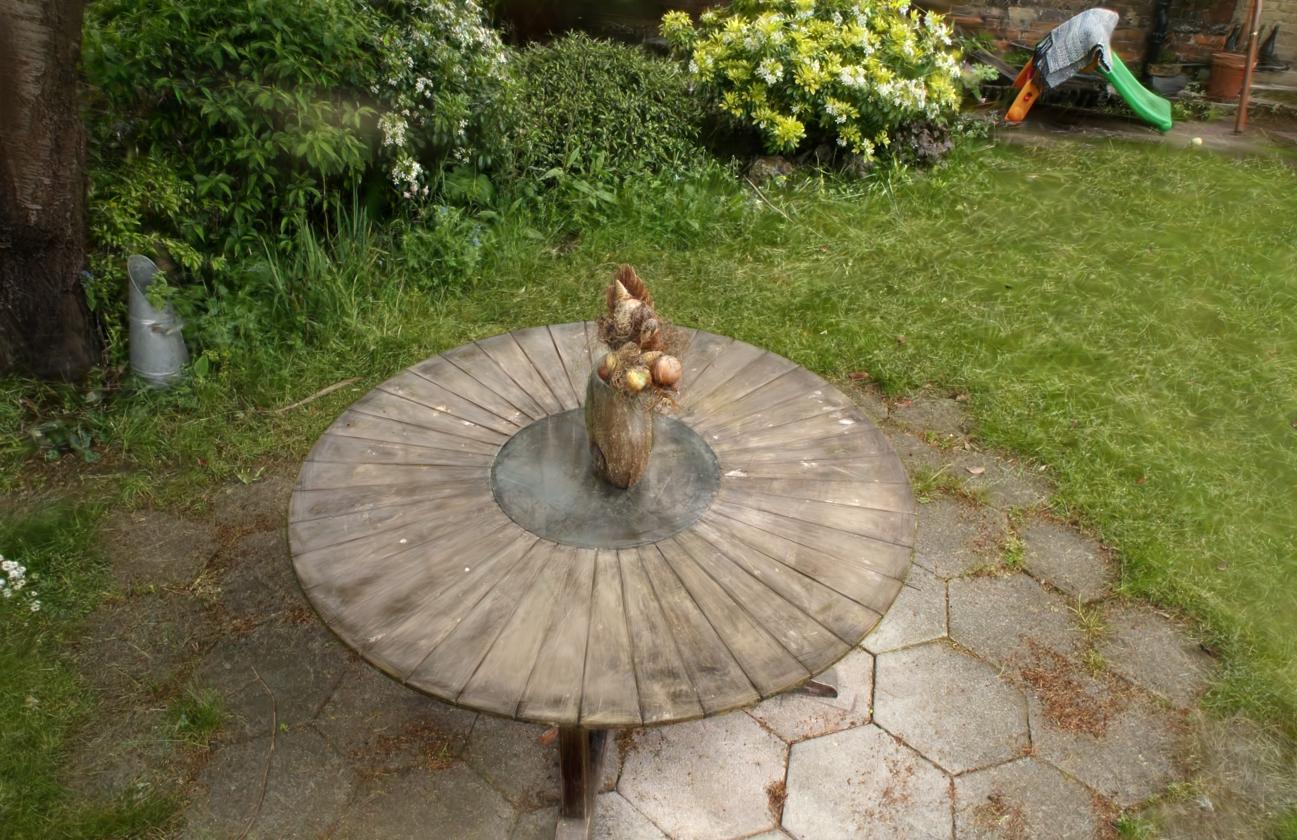} \\

     \includegraphics[width=\mywidth]{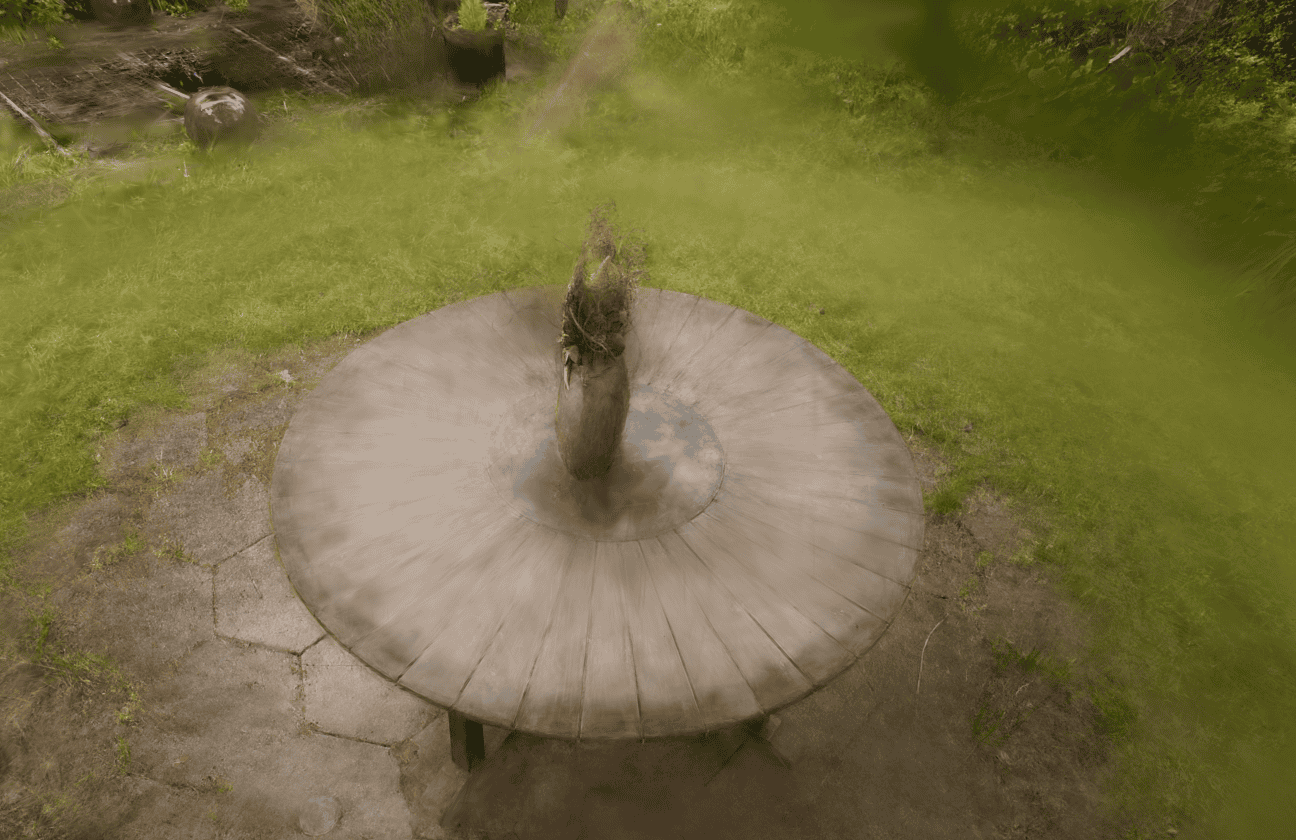}  &
     \includegraphics[width=\mywidth]{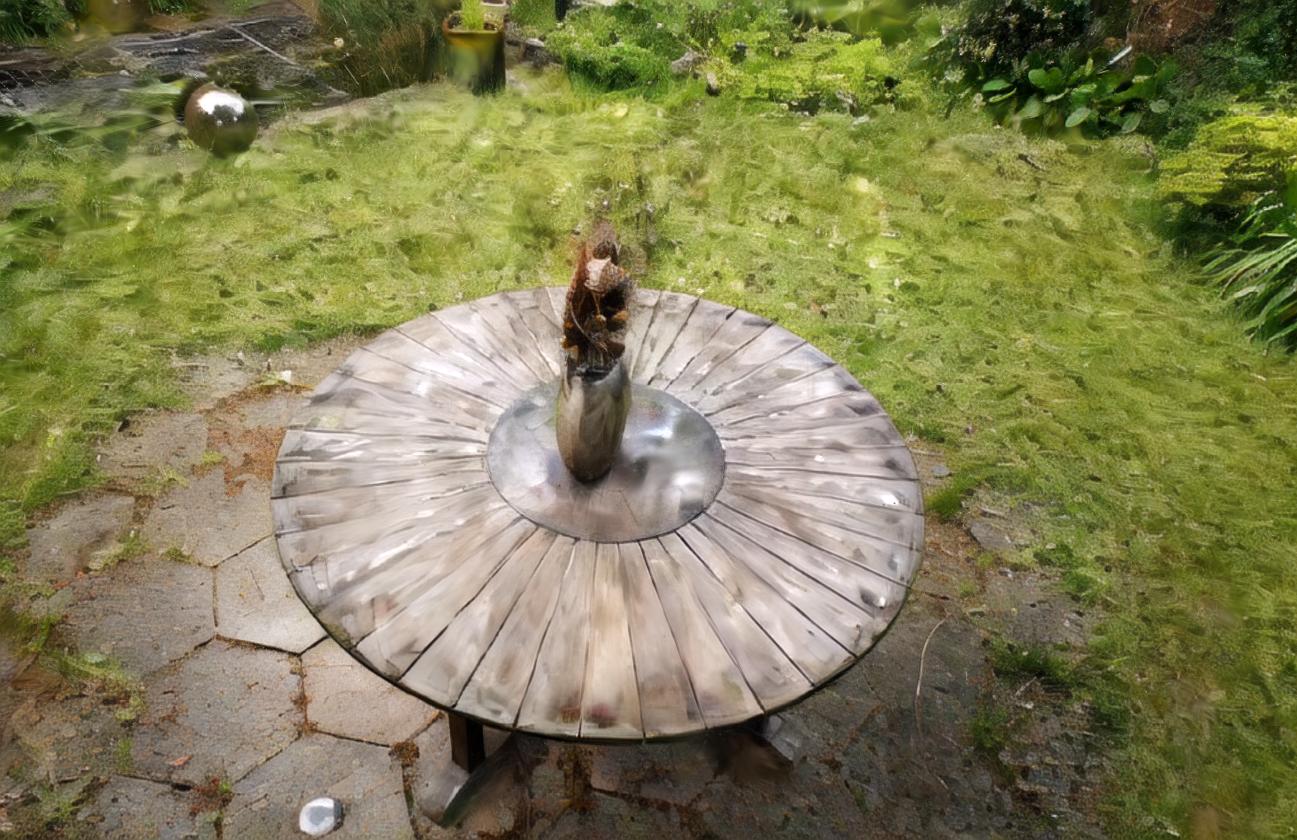}
     &
     \includegraphics[width=\mywidth]{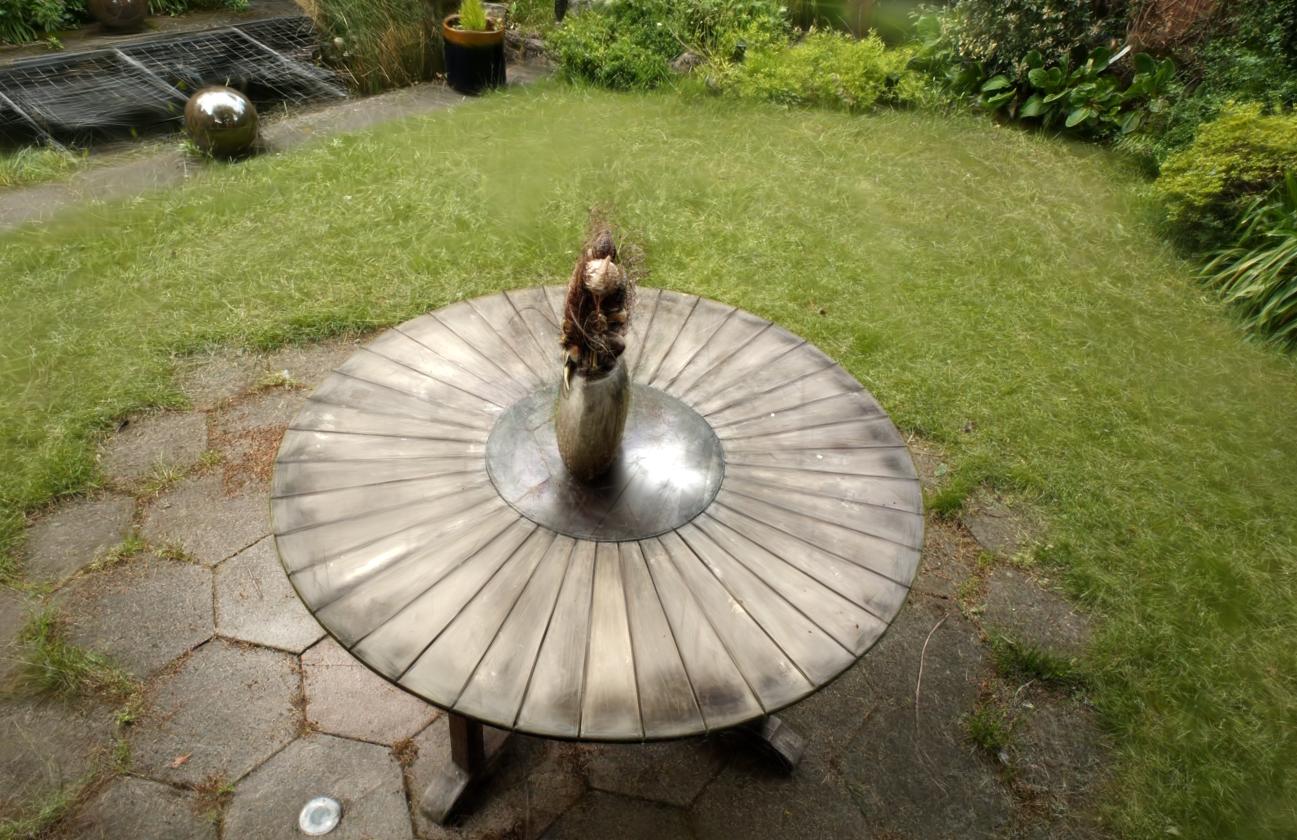}\\

      VE \cite{liu2024novel} + SVD & \paperName+ SVD & \paperName+ Flux
     
     \end{tabular}
     \vspace{-0.2cm}
     \caption{\textbf{Qualitative Ablation on Diffusion Models Selection.} 
     \paperName + Flux yields results with higher fidelity than \paperName + SVD. Additionally, the improved results of \paperName + SVD compared to ViewExtrapolator + SVD highlight the effectiveness of confidence guidance.
     }
     \vspace{-0.4cm}
\label{fig: idm_vs_vdm}
\vspace{-0.4cm}
\end{figure}

\boldparagraph{Datasets} 
We conduct a series of experiments to evaluate the performance of \paperName across multiple datasets with varying settings. We select LLFF \cite{mildenhall2019llff} as the evaluation dataset for forward-facing scenes, Mip-NeRF 360 \cite{barron2021mip} for object-centric scenes, and Waymo \cite{sun2020scalability} for driving scenes.
For the LLFF and MipNeRF datasets, which contain relatively dense captured images, we select sparse or partially observed views as the training set and choose an extrapolated view trajectory that is distant from the views in the training set. The Waymo dataset only provides captured images from a single pass down the street, making it relatively sparse. We only utilize the front cameras as the training set and then translate or rotate the training cameras to create the test views. Details on the design of the training and testing views are provided in the supplementary materials.

\begin{figure*}[t!]
     \centering
     \small 
     \setlength{\tabcolsep}{0pt}
     \def\mywidth{3.55cm}
     \begin{tabular}{P{0.5cm}P{\mywidth}P{\mywidth}P{\mywidth}P{\mywidth}P{\mywidth}}
     
     \rotatebox{90}{~Waymo / 143481} &
     \includegraphics[width=\mywidth]{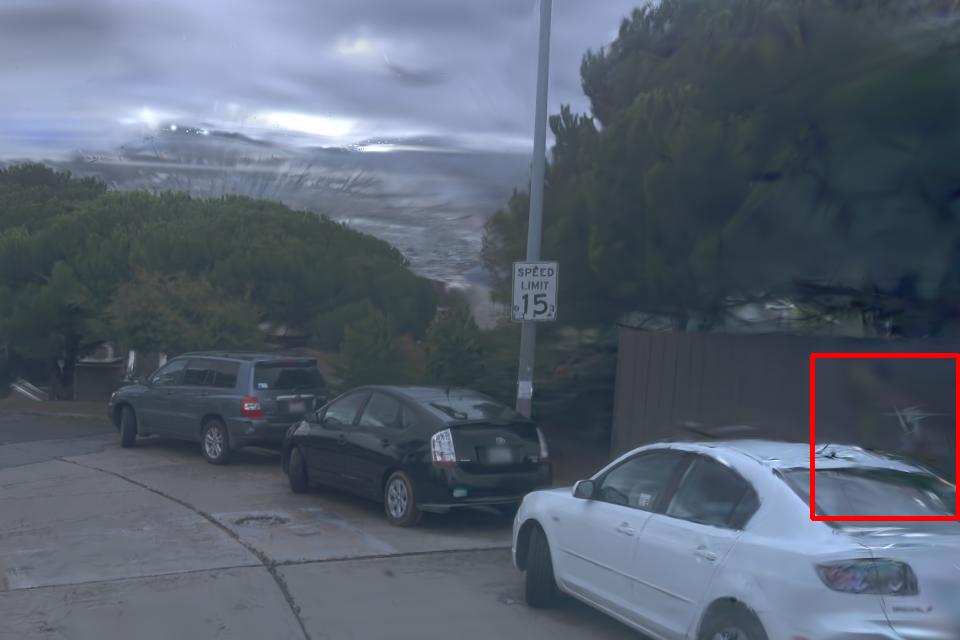}  &
     \includegraphics[width=\mywidth]{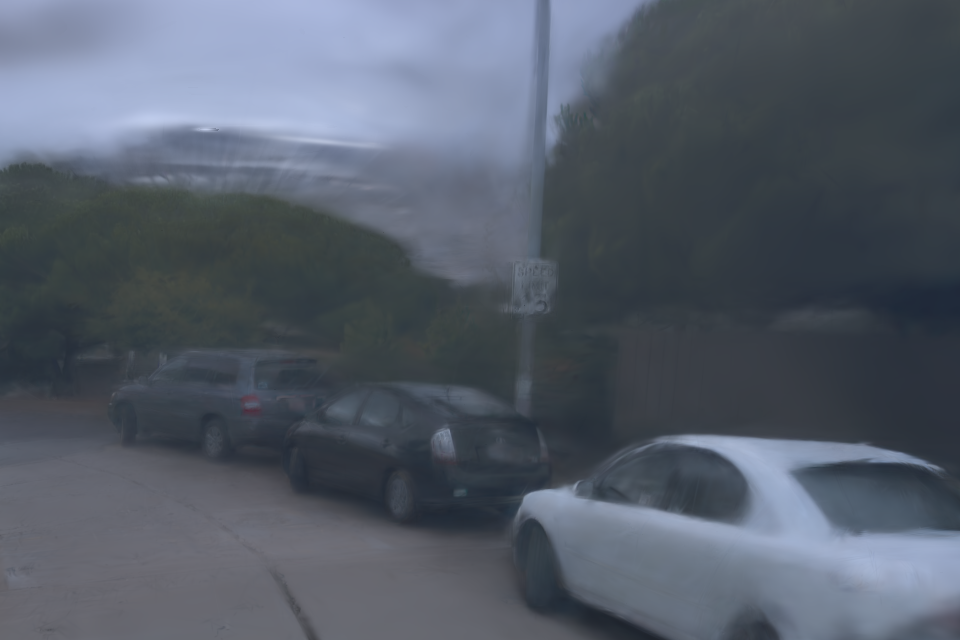} &
     \includegraphics[width=\mywidth]{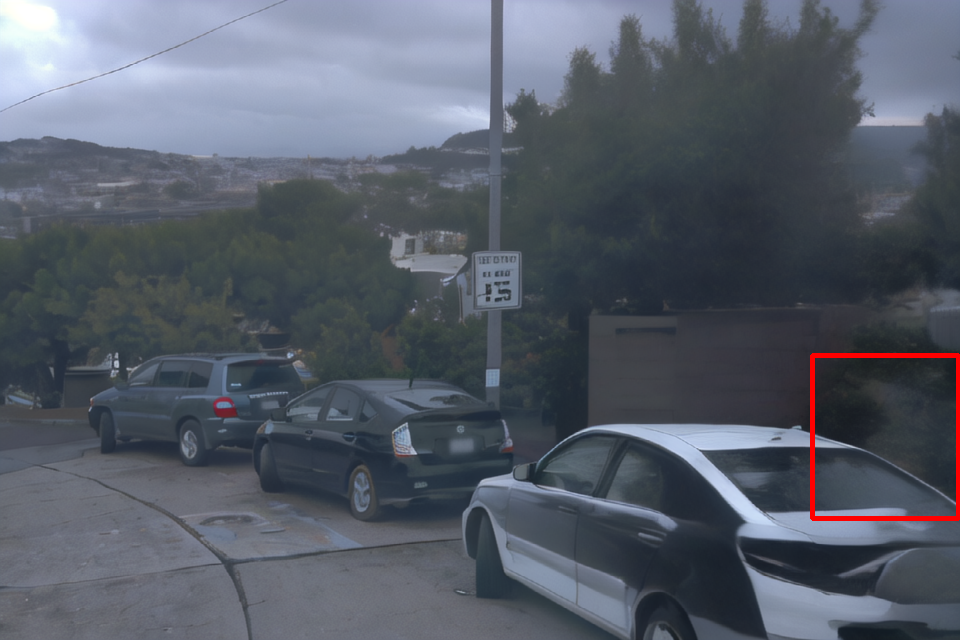} &
     \includegraphics[width=\mywidth]{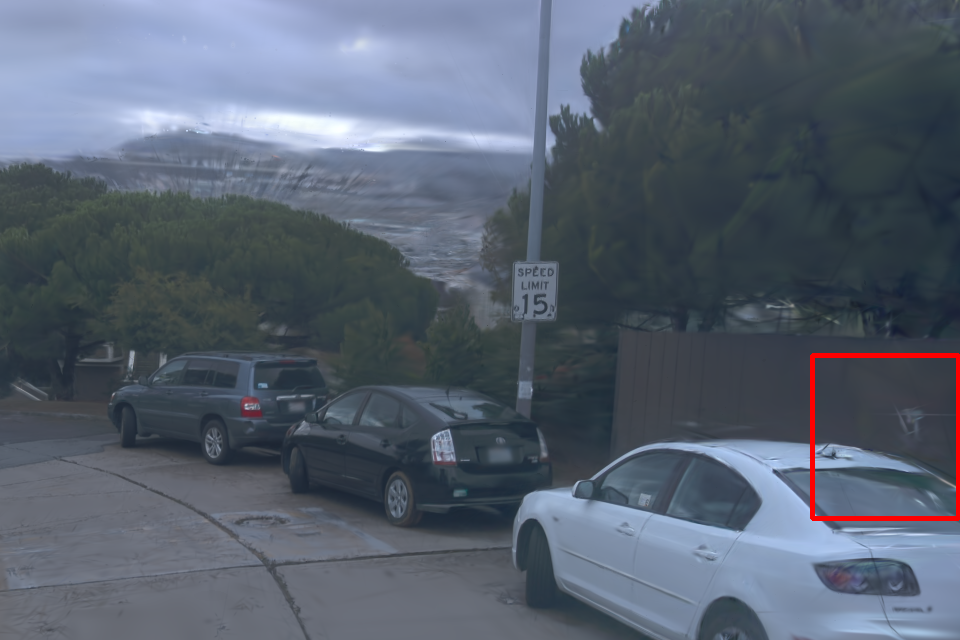} &
     \includegraphics[width=\mywidth]{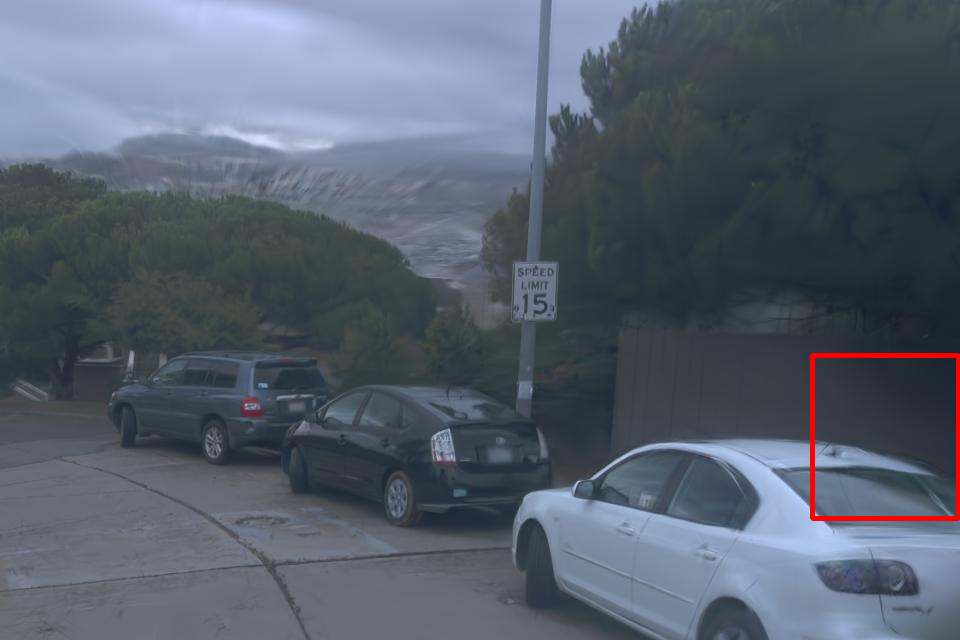} \\

     \rotatebox{90}{~Waymo / 177619} &
     \includegraphics[width=\mywidth]{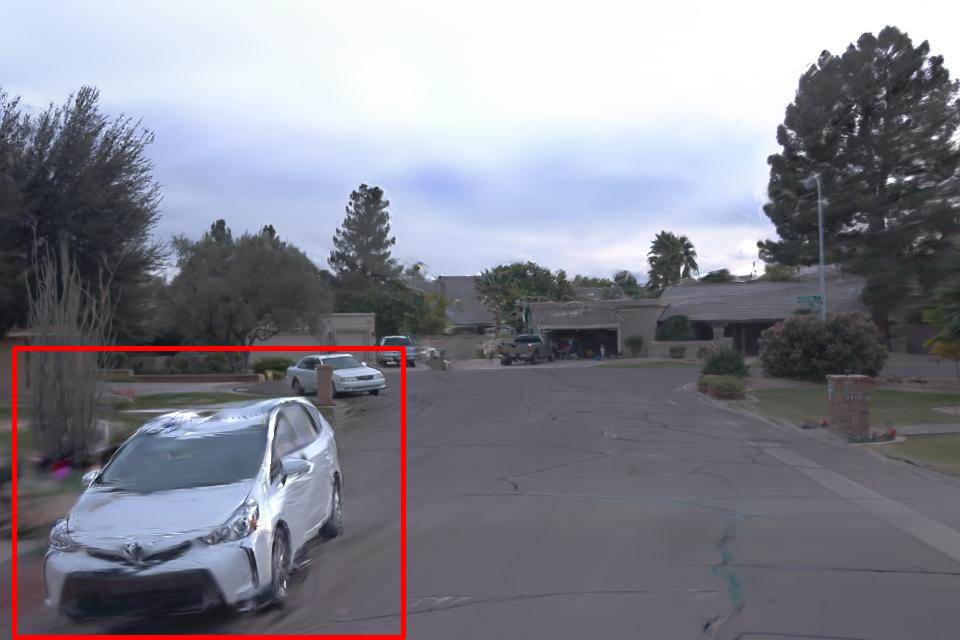}  &
     \includegraphics[width=\mywidth]{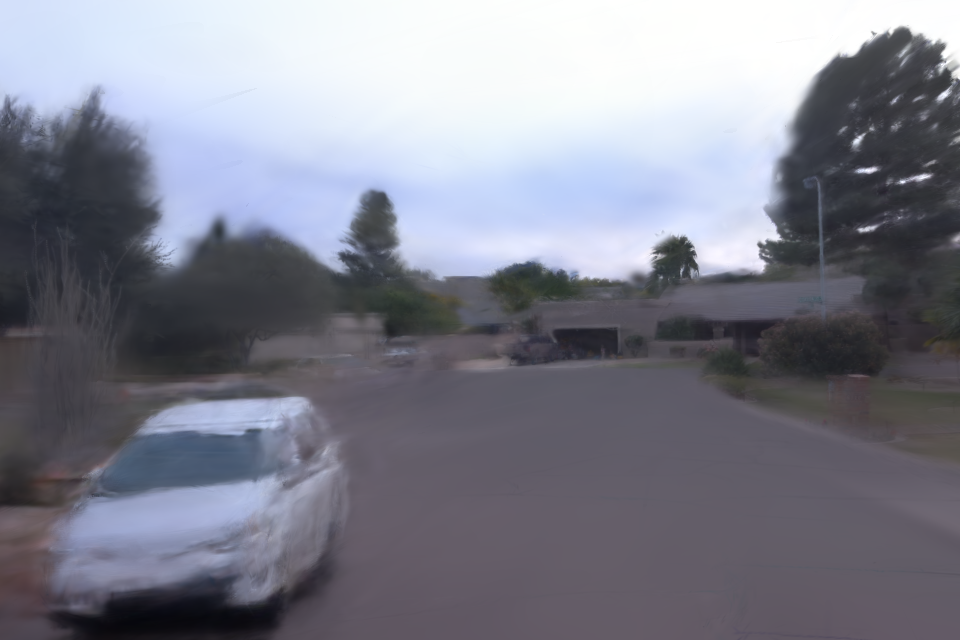} &
     \includegraphics[width=\mywidth]{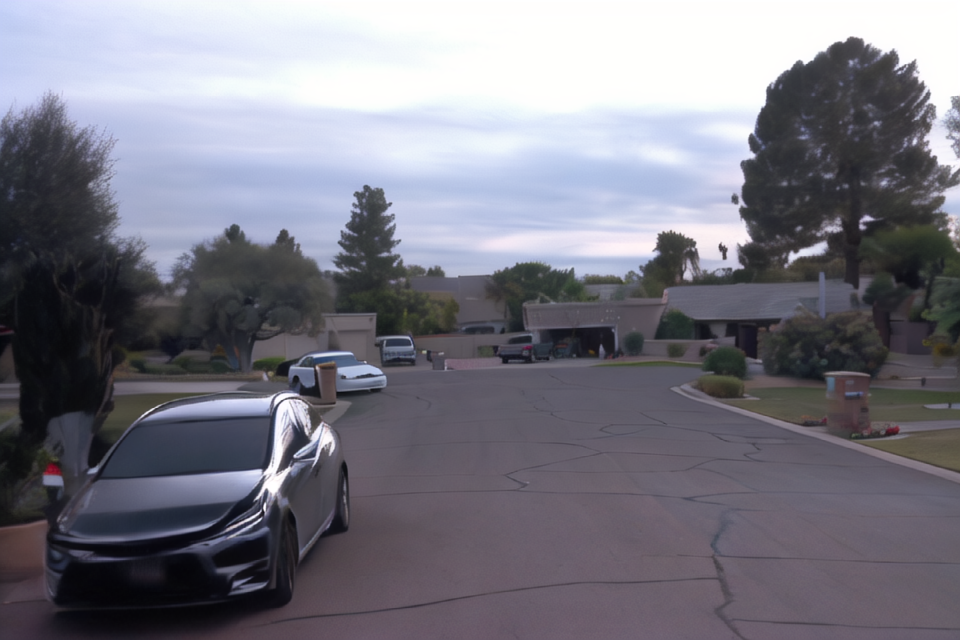} &
     \includegraphics[width=\mywidth]{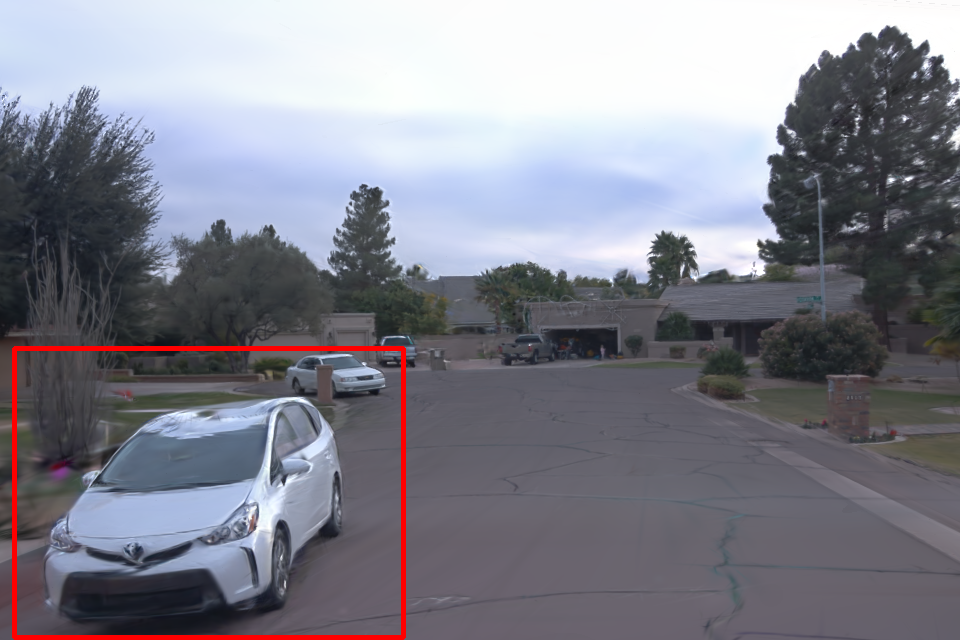} &
     \includegraphics[width=\mywidth]{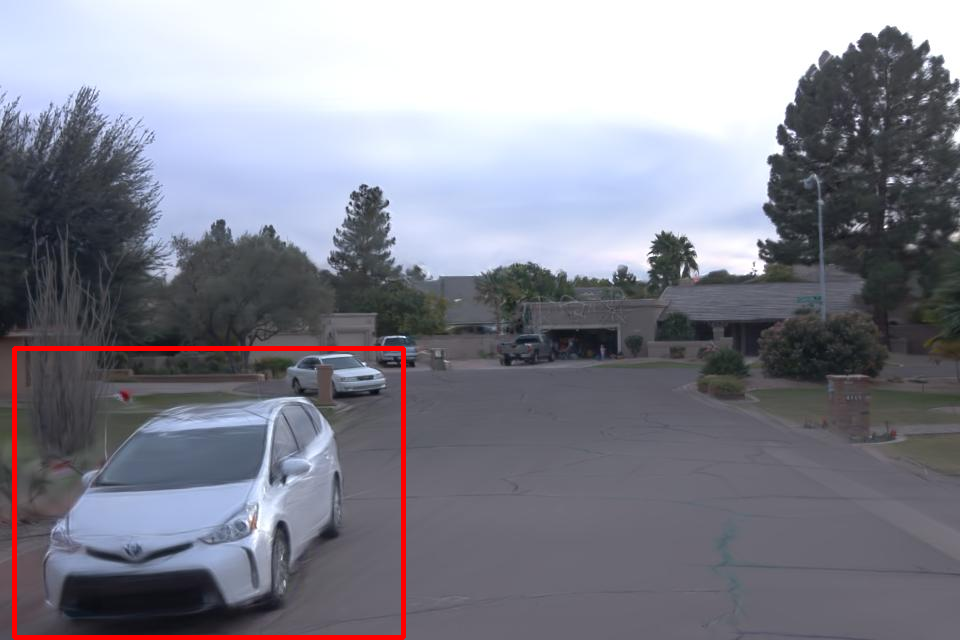} \\

     {} & {3DGS} & {ViewExtrapolator \cite{you2024nvs}} & {StreetCrafter \cite{yan2025streetcrafter}} & Difix3D+ \cite{wu2025difix3d+} & \paperName
     
     \end{tabular}
     \vspace{-0.3cm}
     \caption{\textbf{Qualitative Comparisons on Waymo \cite{sun2020scalability}}. \paperName provide superior performance compared to ViewExtrapolator and StreetCrafter, and is comparable to Difix3D+ in the Waymo dataset. In some cases, \paperName refines the scene even better than Difix3D+.
     }
     \vspace{-0.3cm}
\label{fig: waymo_comp}
\vspace{-0.2cm}
\end{figure*}

\boldparagraph{Model Settings and Baselines}
\paperName utilizes two powerful IDMs as its backbone: SDXL \cite{podell2023sdxl} and Flux \cite{flux2024}, to showcase the capabilities of our method. 
For baseline selection, we consider various methods with different settings. For fine-tuning-free methods, we select ViewExtrapolator \cite{liu2024novel}, and NVS-Solver \cite{you2024nvs} as the baseline. While ViewExtrapolator refines 3DGS with generated views like ours, NVS-Solver employs VDMs as the final renderer, without using 3D renderers, which consumes more computational resources during rendering.
For methods that require fine-tuning of DMs, we choose Difix3D+ \cite{wu2025difix3d+} and StreetCrafter \cite{yan2025streetcrafter} as baselines. StreetCrafter focuses on urban scenes and requires both LiDAR and RGB observations as input, while Difix3D+ is more generalizable and only requires RGB images. For all methods with a 3D renderer, we apply nearly the same 3D refining steps, ensuring that there are sufficient refining steps for the models to converge.

\boldparagraph{Evaluation Metrics}
For the experiments on LLFF and MipNeRF, we adopt the most common settings for quantitative assessments, which include the evaluation of PSNR, SSIM, and LPIPS \cite{zhang2018unreasonable}. In the case of the Waymo dataset, where no ground truth is available for the test images, we utilize KID \cite{binkowski2018demystifying} for quantitative assessments.

\subsection{Comparison with Baselines}

We evaluate \paperName using SDXL \cite{podell2023sdxl} and Flux \cite{flux2024} as the diffusion backbone on the LLFF, Mip-NeRF 360, and Waymo datasets. This includes a quantitative comparison in \cref{tab: main_comp} and qualitative comparisons in \cref{fig: llff_mip_comp} and \cref{fig: waymo_comp} against baseline methods. Although \paperName utilizes only IDMs as the backbone and does not require fine-tuning of the DMs, it still demonstrates performance that is comparable to, or even surpasses, methods that use VDMs or require fine-tuning, both in quantitative and qualitative assessments. 

Specifically, ViewExtrapolator \cite{liu2024novel}, which uses opacity masks as guidance, shows slight improvements in LLFF, although the improvement is less significant compared to our confidence-guided solution.
Moreover, it fails to provide improvements in Mip-NeRF 360 and Waymo. 
This is due to the fact that ViewExtrapolator uses the nearest view from a set of training views as the reference view to generate the test views in a video diffusion model.
While using the nearest training view as the reference view in SVD performs well in the forward-facing scenes in LLFF, where the test views are closer to the training views, this is usually not the case for Mip-NeRF 360 and Waymo, hence ViewExtrapolator yields degraded performance.

Difix3D+ demonstrates the most generalizability and powerful performance across our baselines. \paperName surpasses Difix3D+ \cite{wu2025difix3d+} in LLFF and Mip-NeRF 360, while providing comparable performance in Waymo.
We attribute this to the generalizability of DMs. Although Difix3D+ is finetuned on DLV3D \cite{ling2024dl3dv} and may have encountered similar scenes to those in LLFF and Mip-NeRF 360, the domain gap between datasets still weakens the generalizability of Difix3D+. In contrast, our method maintains the original generalizability of DMs learned from web-scale datasets. Regarding the Waymo dataset, Difix3D+ is fine-tuned on a large-scale in-house driving dataset, where driving scenes are highly structured and exhibit relatively small inter-class differences, making them easier for models to learn.

StreetCrafter \cite{yan2025streetcrafter} is tailored for urban scenes and requires LiDAR as input; for this reason, we only conduct experiments with this model on the Waymo dataset. In contrast to the original setting in StreetCrafter, our setup only provides the front camera to color the LiDAR points, which highlights the limitations of StreetCrafter in this context.
NVS-Solver produces less satisfying results compared to other methods, which may be attributed to inaccurate depth estimation and warping results. We provide NVS-Solver results in supplementary materials.

Please note that we compute the average score across scenes for each dataset. We provide a quantitative comparison for each scene, along with additional qualitative comparisons in the supplementary materials.

\subsection{Ablation Study}

\boldparagraph{Image Diffusion Models vs Video Diffusion Models} 
\paperName can also be applied to VDMs without temporal down-sampling, such as SVD \cite{blattmann2023stable}. Although SVD offers inherent consistency across frames, it suffers from blurriness compared to more advanced IDMs. We conduct an ablation study on the scene from MipNeRF-360/Garden to provide quantitative and qualitative comparisons in \cref{tab: idm_vs_vdm} and \cref{fig: idm_vs_vdm}. Additionally, we include the results from ViewExtrapolator \cite{liu2024novel} on the same scene. While ViewExtrapolator also uses SVD as its backbone, it employs an opacity mask as guidance, which disentangles the effects of the differences in diffusion model backbones and helps demonstrate the effectiveness of our confidence guidance.

\begin{table}[t]
\centering
\small 
\begin{tabular}{lcccc}
\toprule 
& PSNR$\uparrow$ & SSIM$\uparrow$  & LPIPS$\downarrow$ & Guidance \\
\midrule
3DGS
& 18.38 & 0.415 & 0.357 & N/A \\ 
\midrule
VE \cite{liu2024novel} + SVD
& 17.86 & 0.409 & 0.505 & Opacity \\ 
\paperName + SVD
& 19.03 & 0.453 & 0.331 & Certainty \\ 
\midrule
\paperName + SDXL
& 19.41 & 0.517 & 0.294 & Certainty \\ 
\paperName + Flux
& \textbf{19.72} & \textbf{0.520} &\textbf{ 0.287} & Certainty \\
\bottomrule
\end{tabular}
\vspace{-0.3cm}
\caption{\textbf{Quantitative Ablation on Diffusion Models Selection.} }
\label{tab: idm_vs_vdm}
\vspace{-0.1cm}
\end{table}

\begin{table}[t]
\centering
\small 
\begin{tabular}{lcccc}
\toprule 
& PSNR$\uparrow$ & SSIM$\uparrow$  & LPIPS$\downarrow$ \\
\cline{2-4}
Raw Flux \cite{flux2024} & 19.23 & 0.390 & 0.389 \\
+ Confidence Guidance & 19.32 & 0.435 & 0.349 \\ 
+ Interleave Strategy & 19.65 & 0.517 & 0.293 \\
+ Overall Guidance & \textbf{19.72} & \textbf{0.520} & \textbf{0.287} \\ 
\bottomrule
\end{tabular}
\vspace{-0.3cm}
\caption{\textbf{Ablation Study on Modules of \paperName}. We incorporate each module from the raw Flux model to illustrate its necessity. }
\label{tab: module_abl}
\vspace{-0.6cm}
\end{table}

\boldparagraph{Effectiveness of Interleaved 2D-3D Refinement}
The interleaved refining strategy, confidence guidance, and overall guidance are crucial for ensuring that the generation aligns with the original scenes and enhances consistency across frames. We conduct an ablation study of these modules on the scene from MipNeRF-360/Garden, as shown in \cref{tab: module_abl}. We perform experiments starting from a raw Flux model, which we slightly modify to function as an image-to-image model. We progressively add components from \paperName to demonstrate the necessity of these techniques.

\section{Conclusion} 

In this paper, we present \paperName, a method for fixing artifacts and improving the quality of 3DGS without fine-tuning DMs. \paperName demonstrates state-of-the-art performance across various datasets and possesses strong capabilities for deployment with future, more advanced DMs. 
However, \paperName still has certain limitations.  It may encounter failure cases when extrapolated views lead to excessive artifacts with minimal credible guidance.  Additionally, the updating process for 3DGS is relatively slow and challenging to converge over dozens of refining steps. These challenges suggest opportunities for future work on designing more robust and efficient methods for integrating 3D reconstruction with 2D generative models.

\vspace{0.22cm}
{\small
\noindent\textbf{Acknowledgements:}
This work is supported by NSFC under grant 62202418, U21B2004, and 62441223, the National Key R\&D Program of China under Grant 2021ZD0114501, and Scientific Research Fund of Zhejiang University grant XY2025028. 
}

\newpage

{
    \small
    \bibliographystyle{ieeenat_fullname}
    \bibliography{main}
}
\clearpage
\setcounter{page}{1}
\maketitlesupplementary

\section{3DGS Fisher Information Derivation}
The uncertainty attribute of 3DGS in this paper is defined as:
\begin{equation}
    \mc{\bar{C}}_{\mc{V}; \mc{G}} = -\log p_f(\pi | \mc{V}; \mc{G})
\end{equation}
Under the following regularity conditions, $-\log p_f(\pi | \mc{V}; \mc{G})$ can be viewed as a loss term for Fisher information. It can also be expressed as an expectation term to represent Fisher information:  $-\mathbb{E}_{\log p_f}[\frac{\partial^2 \log p_f(\pi | \mc{V}; \mc{G})}{\partial\mc{G}^2}]$:
\begin{itemize}
    \item The partial derivative of $p_f(\pi | \mc{V}; \mc{G})$ with respect to $\mc{G}$ exists almost everywhere.
    \item The integral of $p_f(\pi | \mc{V}; \mc{G})$ can be differentiated under the integral sign with respect to $\mc{G}$.
    \item The \textit{support} of $p_f(\pi | \mc{V}; \mc{G})$ does not depend on $\mc{G}$. In mathematics, the \textit{support} of a real-valued function $p_f$ is the subset of the function domain of elements that are not mapped to zero.
\end{itemize}
The volume rendering of 3D Gaussians meets these regularity conditions. With the consideration of $-\log p_f(\pi | \mc{V}; \mc{G})$ can be regarded as the loss term of $\mc{L}$, the uncertain attribute of 3DGS can be represented as:
\begin{align}
    \mc{\bar{C}}_{\mc{V}; \mc{G}} &= -\mathbb{E}_{\log p_f}[\frac{\partial^2 \log p_f(\pi | \mc{V}; \mc{G})}{\partial\mc{G}^2}] \nonumber \\
    & = \mathbb{E}_{\log p_f}[\frac{-\partial^2 \log p_f(\pi | \mc{V}; \mc{G})}{\partial\mc{G}\partial\mc{G}^T}] \nonumber \\
    & = \mathbb{E}_{\log p_f}[\frac{\partial^2 \cL(\mc{G})}{\partial\mc{G}\partial\mc{G}^T}] \nonumber \\
    & = \bH^{''}[\pi|\mc{V}; \mc{G}] \nonumber \\ 
    & = \nabla_{\mc{G}}\pi(\mc{V};\mc{G})^T\nabla_{\mc{G}}\pi(\mc{V};\mc{G})
\label{eq: uncertainty}
\end{align}

\section{Extrapolated Views Design}

We design extrapolated testing views for the LLFF \cite{mildenhall2019llff}, Mip-NeRF 360 \cite{barron2021mip}, and Waymo \cite{sun2020scalability} datasets. The process for generating testing views in the Waymo dataset is straightforward; we translate the camera by 2 to 3 meters or rotate it by 10 to 15 degrees horizontally. However, the design for LLFF and Mip-NeRF 360 is not as straightforward, as we aim to construct extrapolated views that have ground truth images. For this reason, we cannot generate trajectories freely; instead, we need to create partitions for the testing and training sets. We present visualizations of the training and testing cameras in \cref{fig: partition_view} from these scenes to illustrate the design of the extrapolated views. For some scenes where obvious extrapolated trajectories cannot be directly extracted, we aim to make the training views sparse in order to produce relative extrapolated trajectories.

\definecolor{mygray}{rgb}{0.77, 0.77, 0.77}
\definecolor{myred}{rgb}{1.0, 0.57, 0.57}

\begin{figure*}
     \centering
     \small 
     \setlength{\tabcolsep}{0pt}
     \def\mywidth{3.55cm}
     \begin{tabular}{P{0.5cm}P{\mywidth}P{\mywidth}P{\mywidth}P{\mywidth}P{\mywidth}}
     
     \rotatebox{90}{~~~~~~~~~~~~~LLFF} &
     \includegraphics[width=\mywidth]{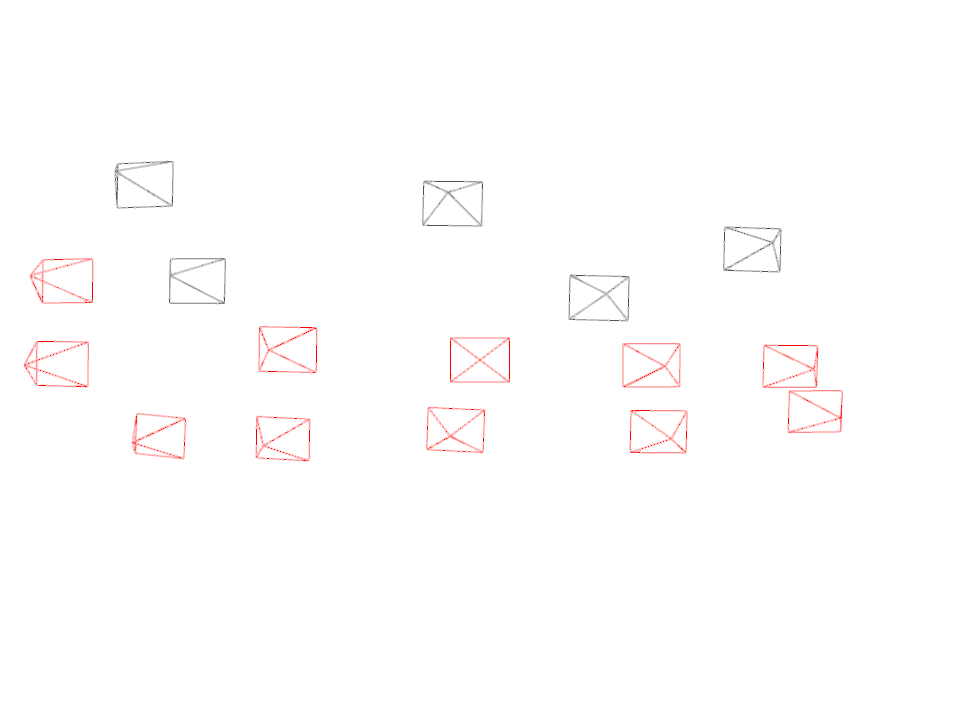}  &
     \includegraphics[width=\mywidth]{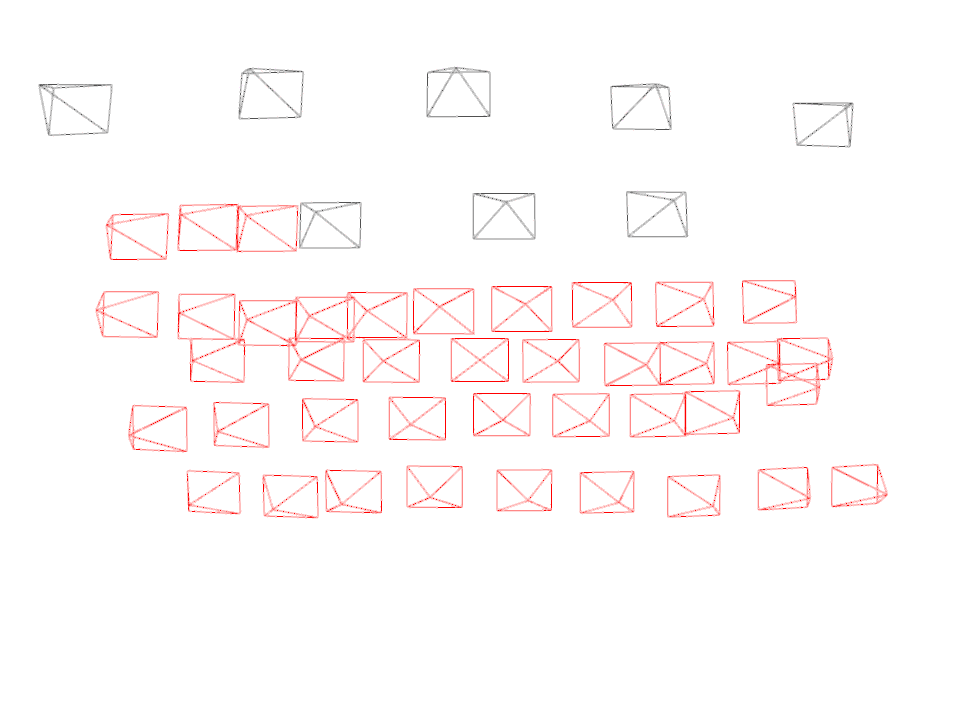} &
     \includegraphics[width=\mywidth]{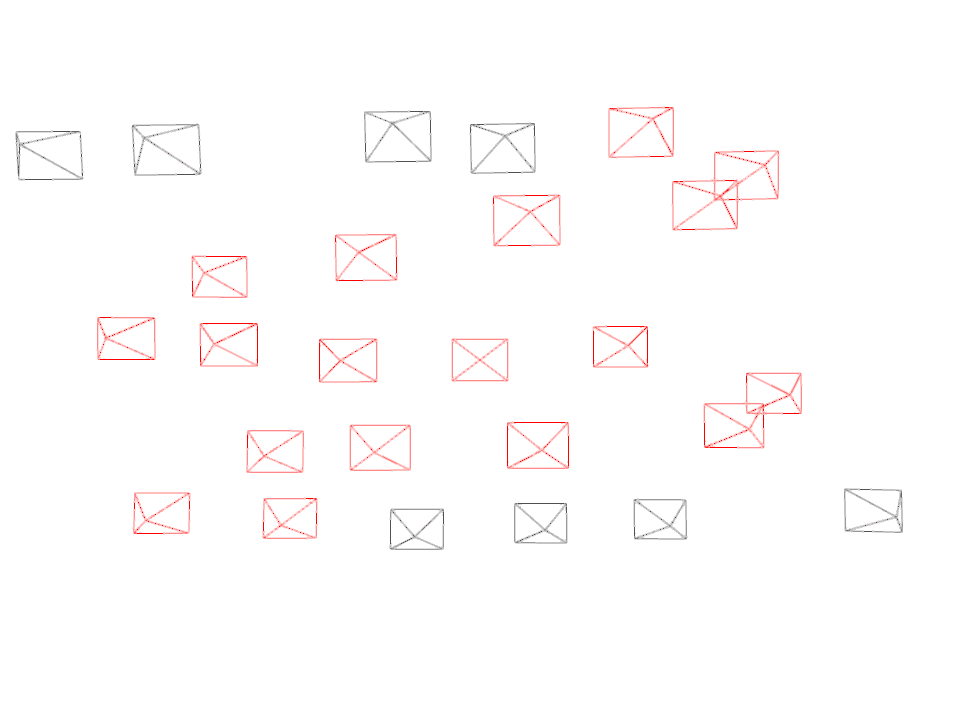} &
     \includegraphics[width=\mywidth]{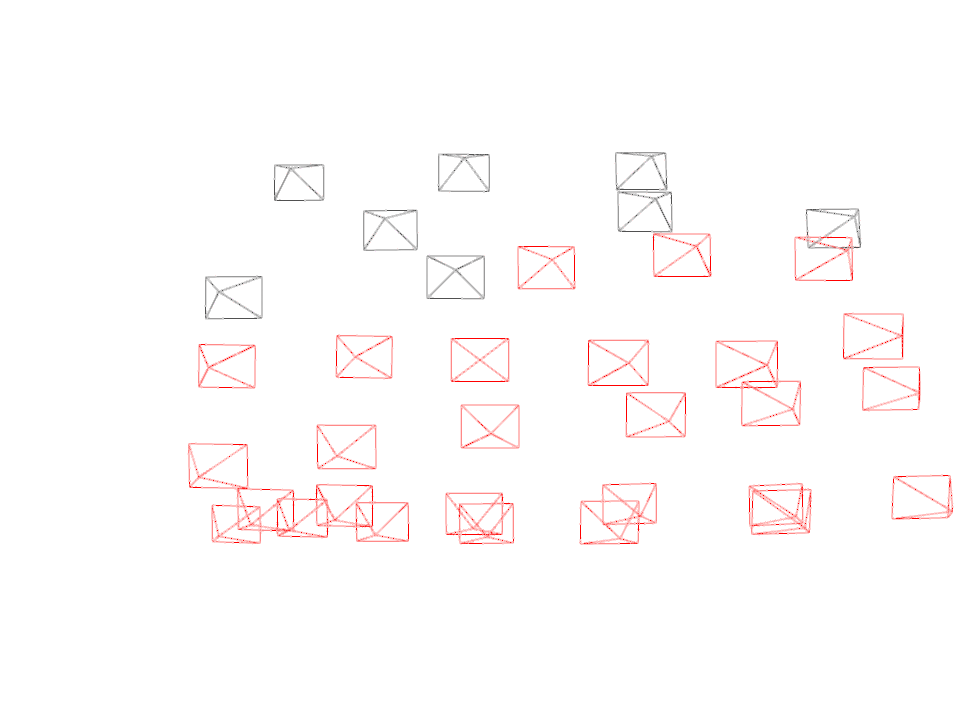} &
     \includegraphics[width=\mywidth]{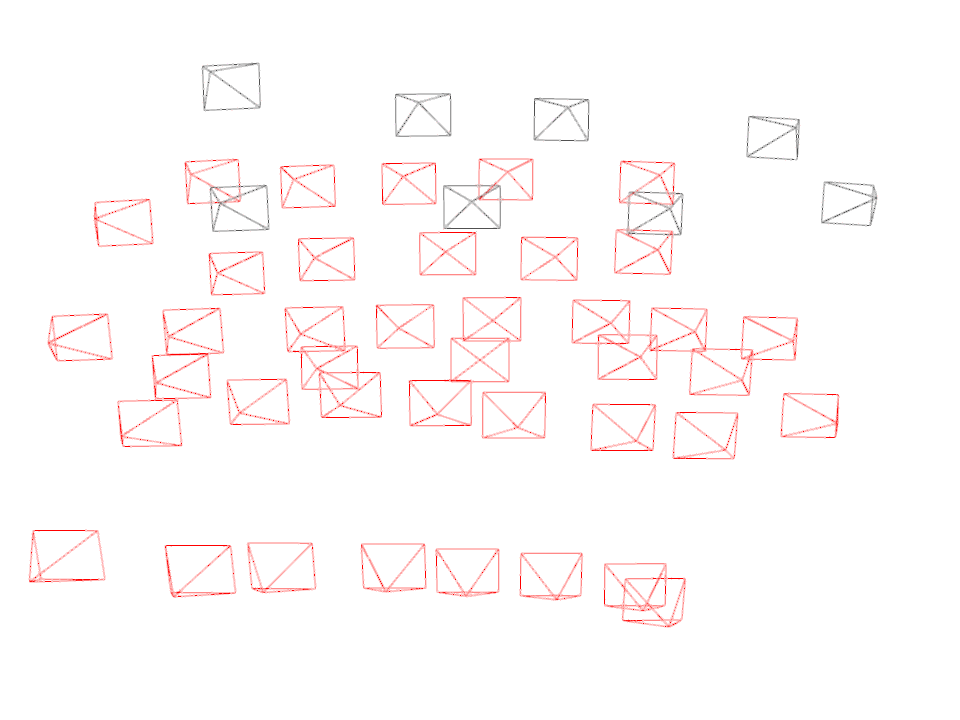} \\

     & fern & horns & leaves & fortress & trex \\

     \\
     \\

     \rotatebox{90}{~~~~~~Mip-NeRF 360} &
     \includegraphics[width=\mywidth]{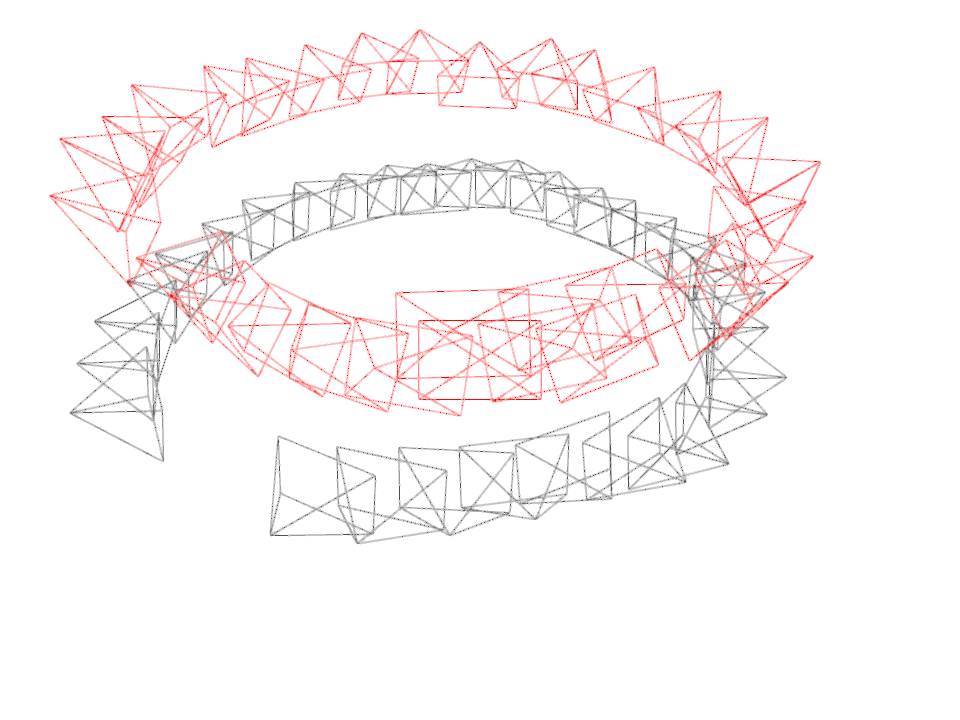}  &
     \includegraphics[width=\mywidth]{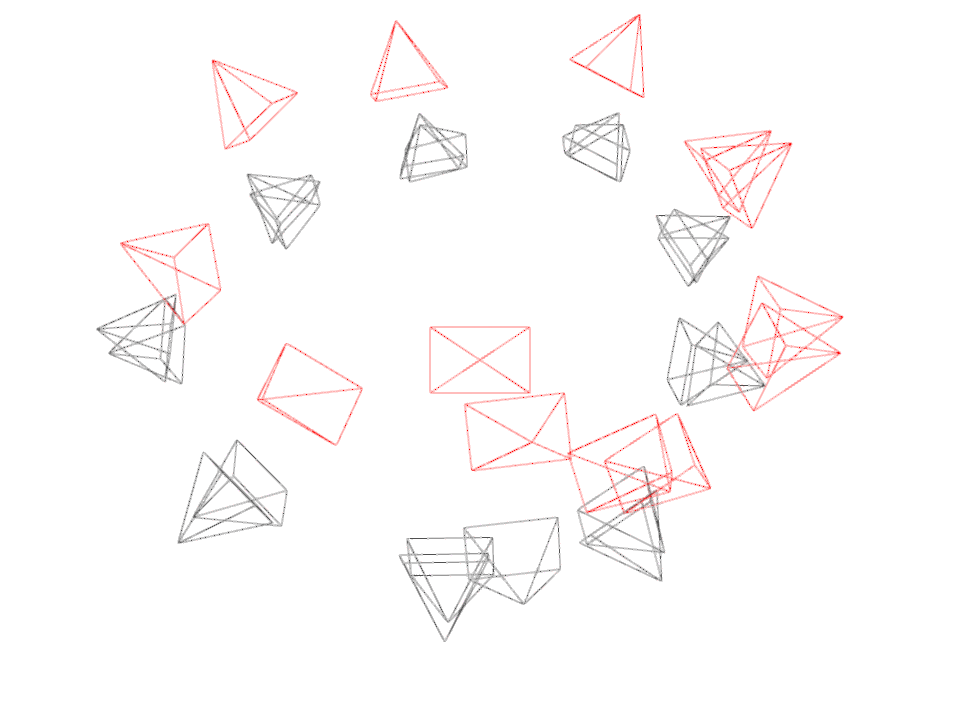} &
     \includegraphics[width=\mywidth]{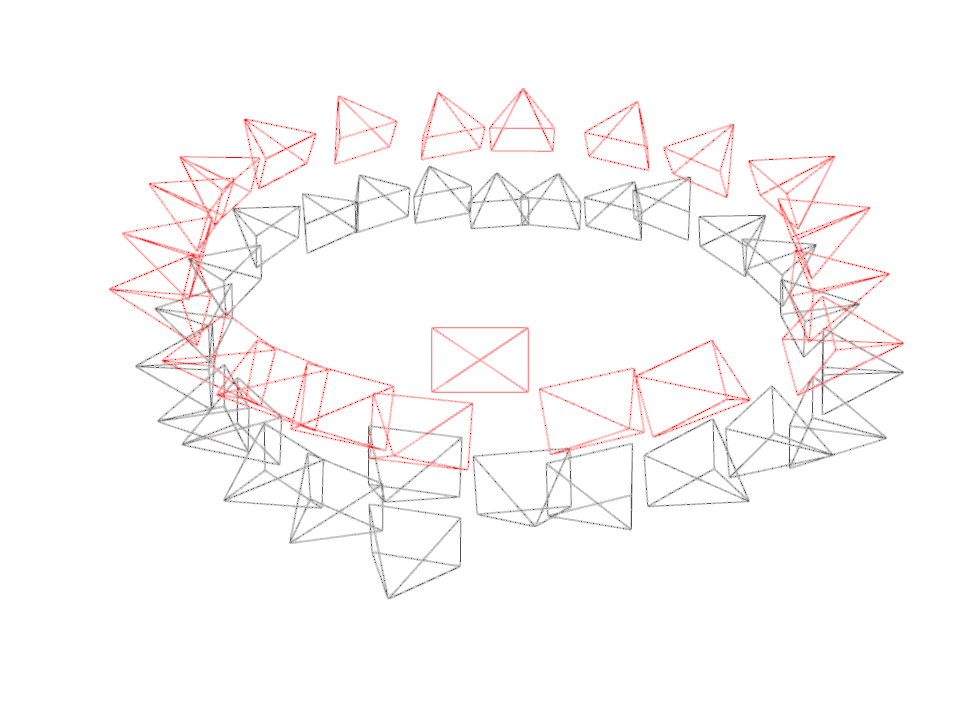} &
     \includegraphics[width=\mywidth]{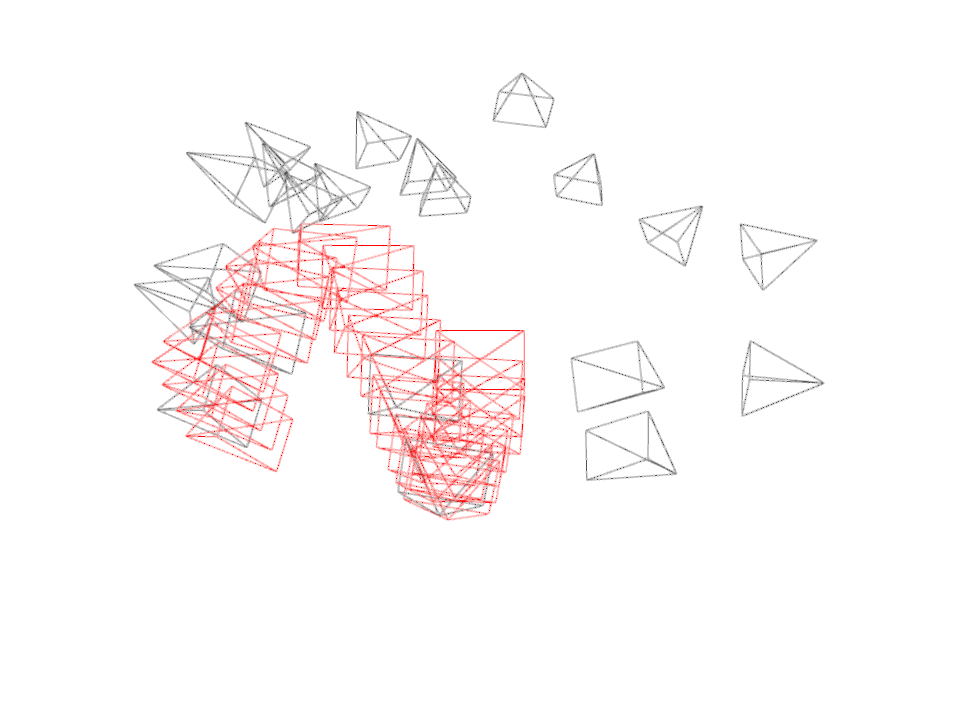} &
     \includegraphics[width=\mywidth]{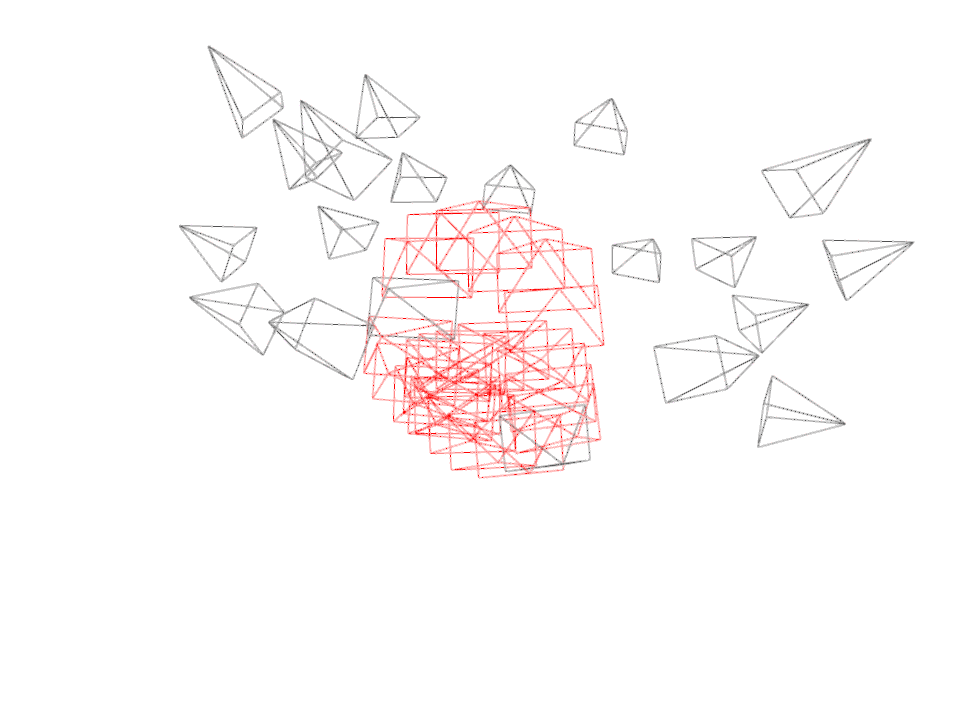} \\

     & garden & stump & bicycle & counter & kitchen

     \end{tabular}
     \vspace{-0.2cm}
     \caption{\textbf{Design of Training and Testing Views Design.} We design partitions to conduct experiments on extrapolated testing views. \colorbox{mygray}{Training views} and \colorbox{myred}{Testing views} are highlighted with their respective colors.
     }
     \vspace{-0.2cm}
\label{fig: partition_view}
\end{figure*}

\section{Sampling Strategy}

The supervisions during 3D refinement for $\mc{G}_i$ are sampled from current refined view $(\mc{V}_{i}^e, \hat{\mc{I}}^{e,f}_{i})$, previous refined views $\mc{F}_{i-1}$ and training views $S_{train}$. Each stage of 3D refinement aims to fit the newly refined 2D image while preserving rendering ability in the original training and previously refined views.
The sampling strategy for training is structured as follows. During the first third of the 3D refinement steps, every three steps are designated as current-refine steps, using the current refine image $\hat{\mc{I}}^{e,f}_i$ to refine 3DGS. In the subsequent third of the 3D refinement steps, every five steps are defined as current-refine steps, and in the final third of the 3D refinement steps, every eight steps are designated as current-refine steps. For the remaining non-current-refine steps, we randomly select views from the training set $\mc{S}_{train}$ and the previous refined set $\mc{F}_{i-1}$, but with different selection weights. The probability of selecting views from $\mc{F}_{i-1}$ is lower compared to that of selecting views from $\mc{S}_{train}$.

\vspace{-0.2cm}

\section{Additional Experiments}

\subsection{More Comparisons with Baselines}
\begin{figure*}[t!]
     \centering
     \small 
     \setlength{\tabcolsep}{0pt}
     \def\mywidth{3.6cm}
     \begin{tabular}{P{\mywidth}P{\mywidth}P{\mywidth}P{\mywidth}P{\mywidth}}
     
     \includegraphics[width=\mywidth]{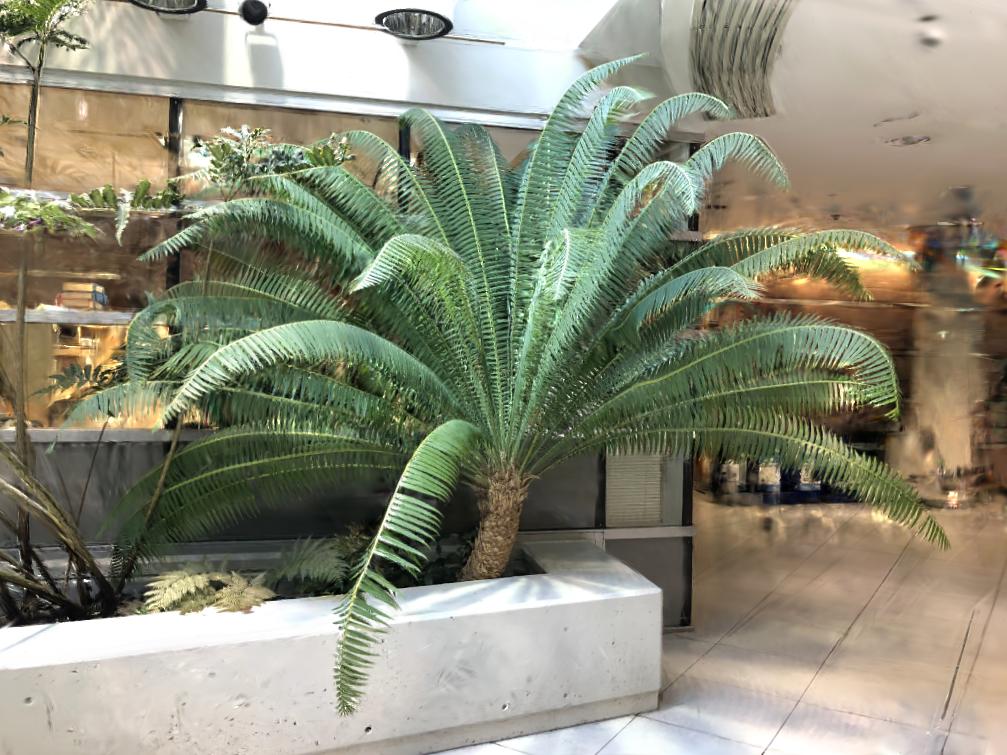}  &
     \includegraphics[width=\mywidth]{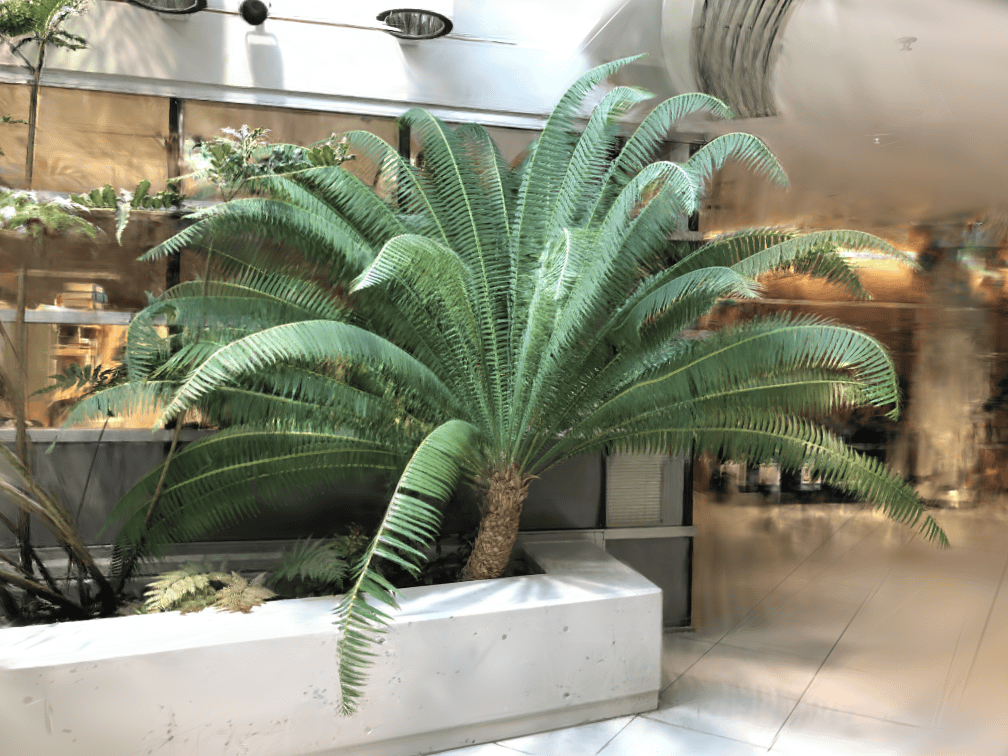} &
     \includegraphics[width=\mywidth]{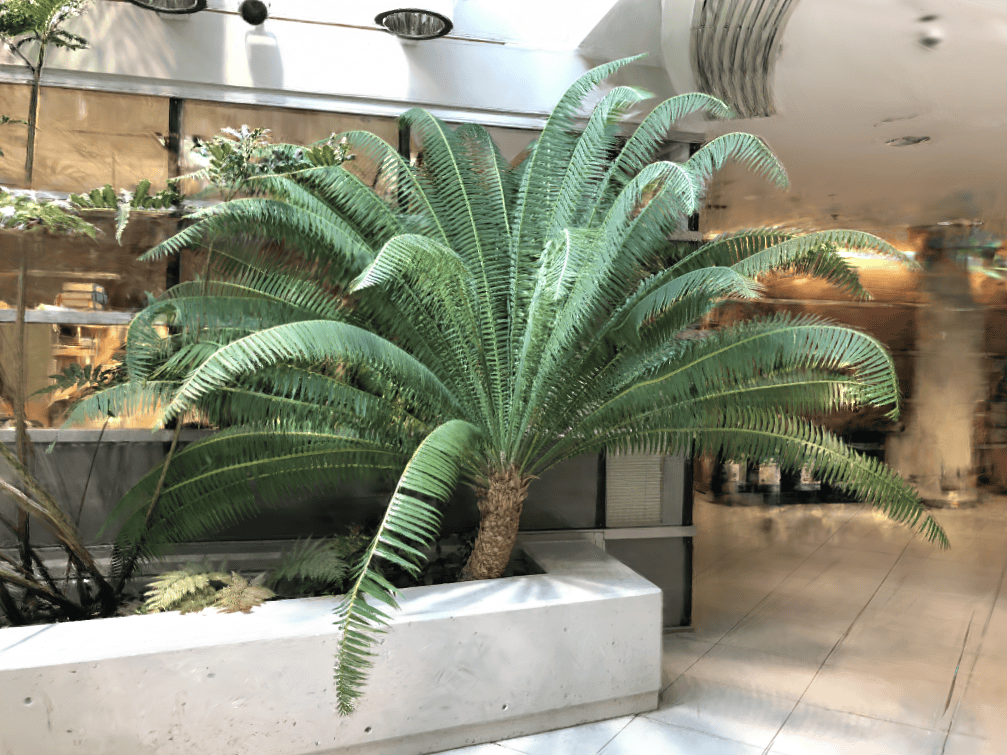} &
     \includegraphics[width=\mywidth]{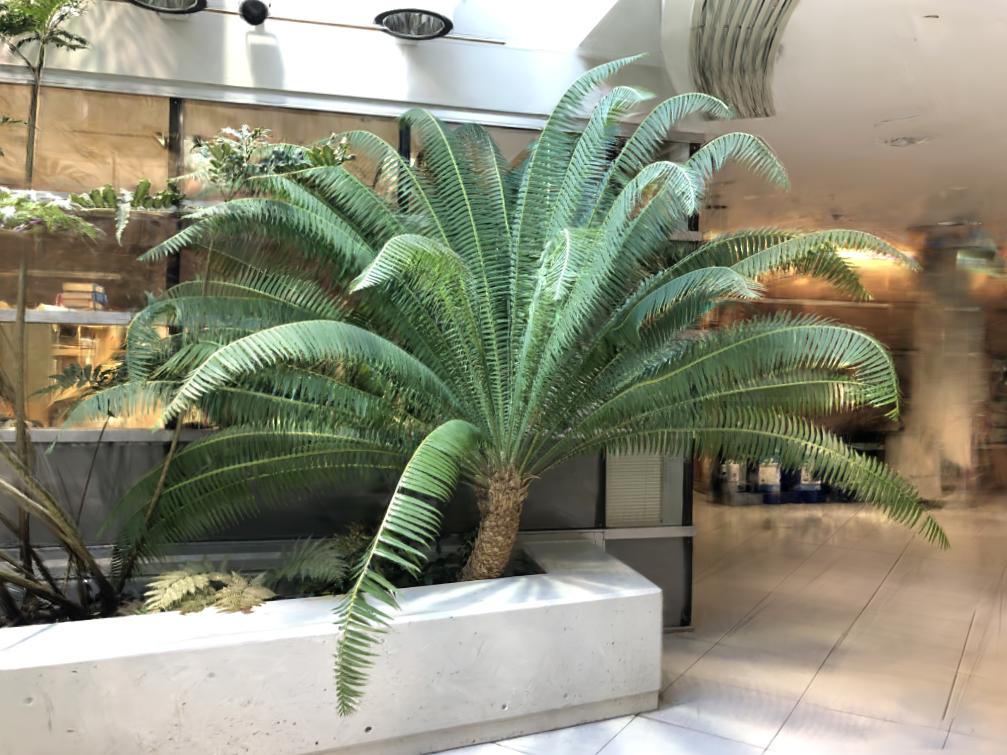} &
     \includegraphics[width=\mywidth]{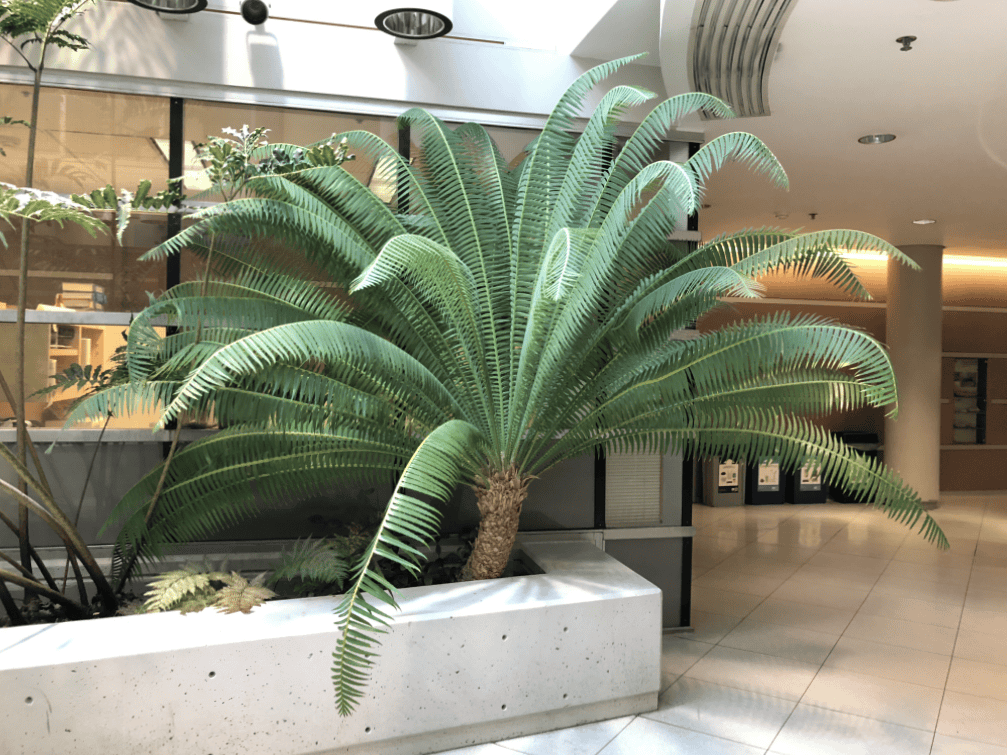} \\

     \includegraphics[width=\mywidth]{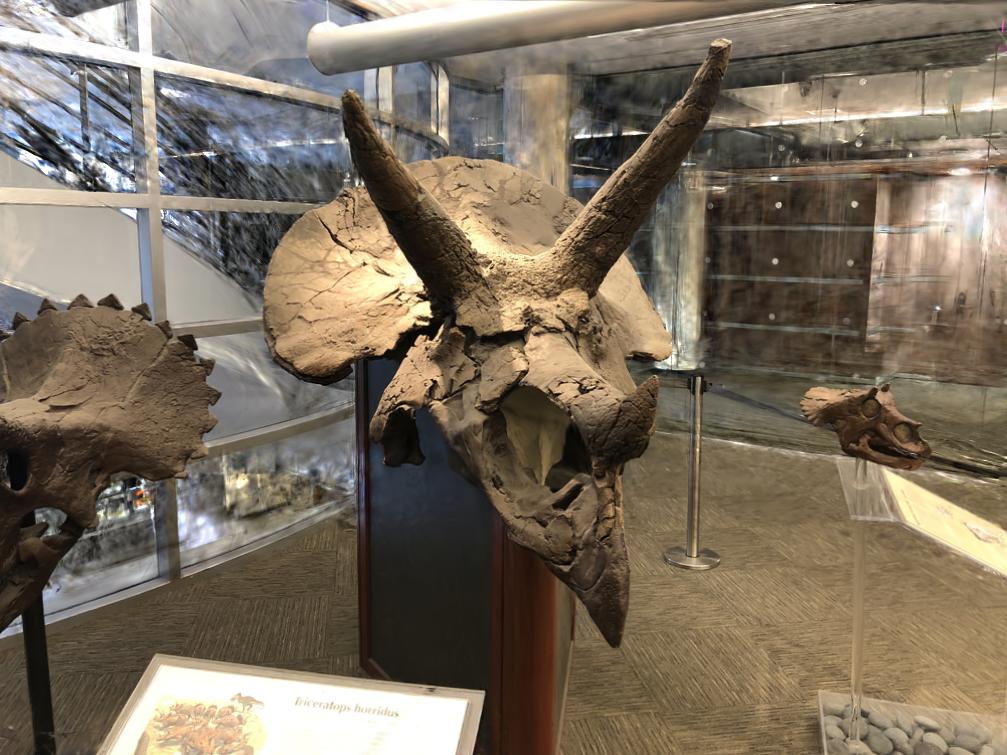}  &
     \includegraphics[width=\mywidth]{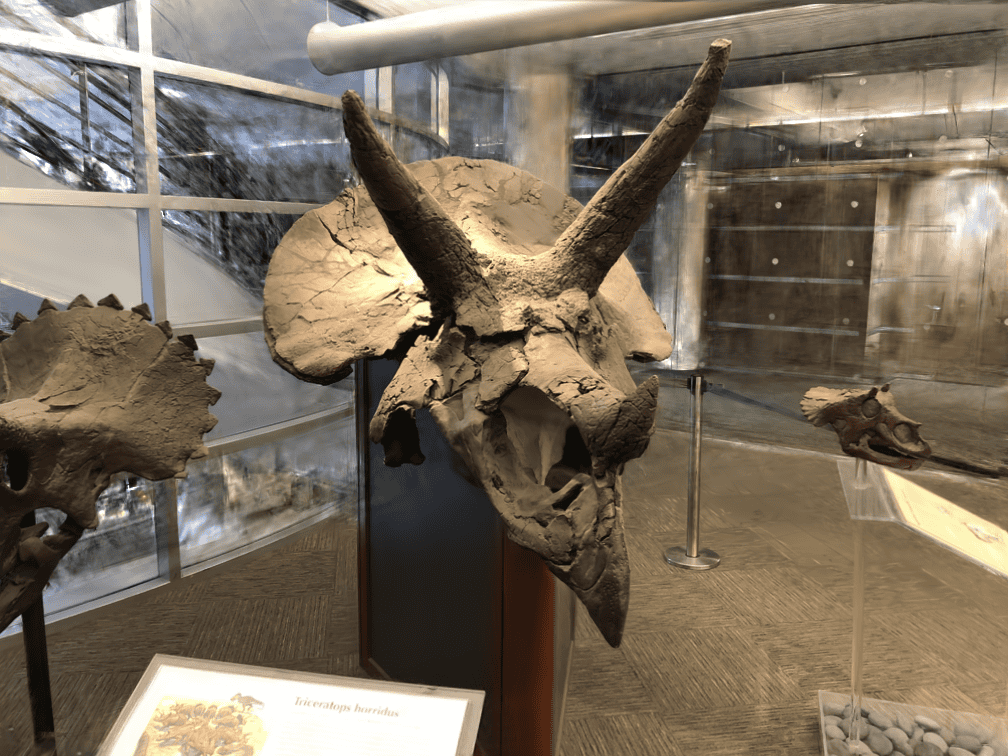} &
     \includegraphics[width=\mywidth]{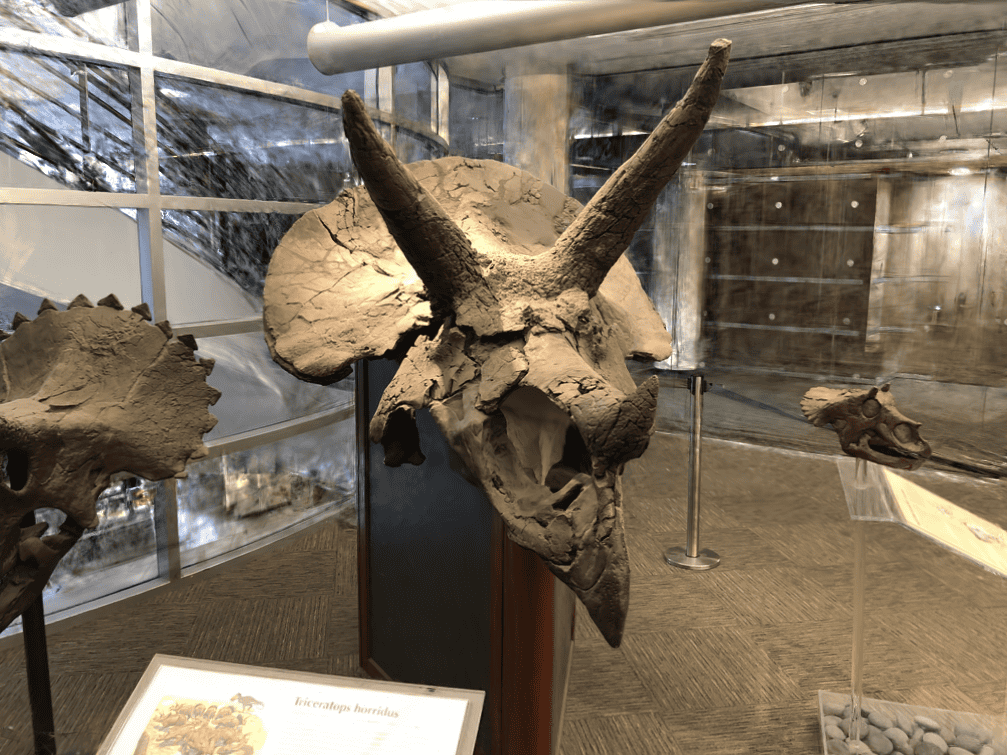} &
     \includegraphics[width=\mywidth]{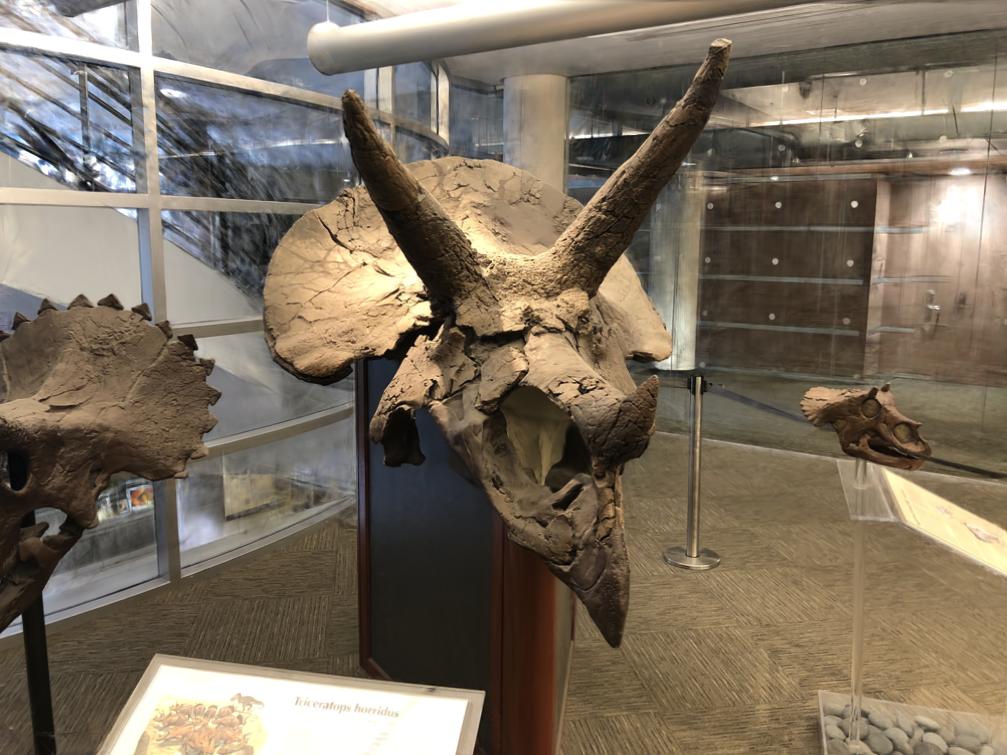} &
     \includegraphics[width=\mywidth]{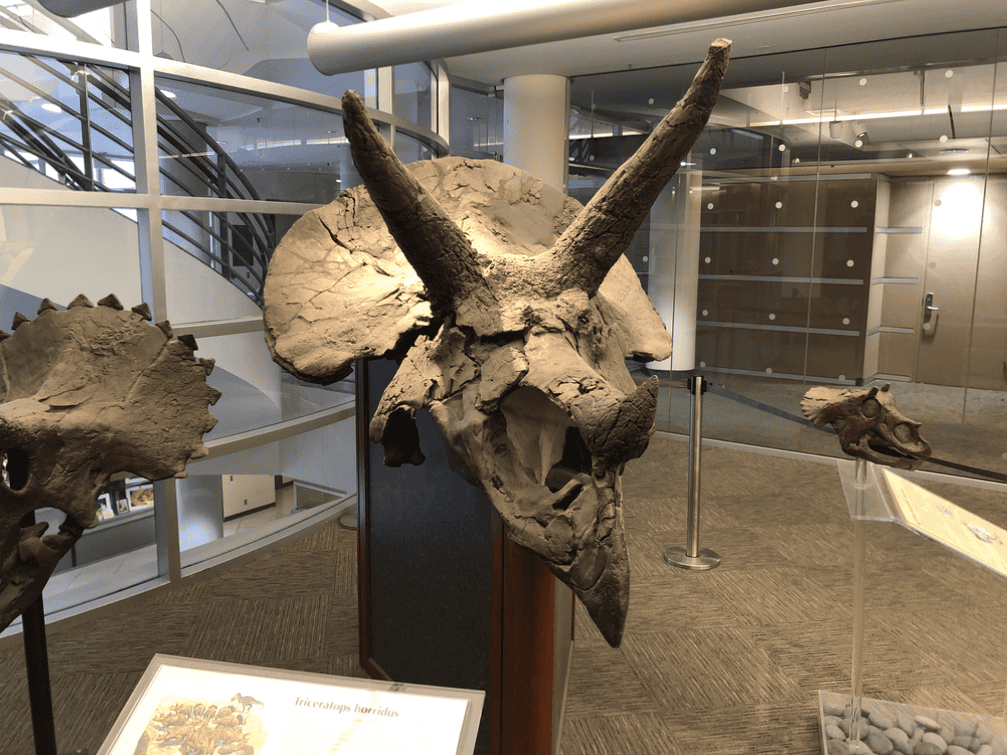} \\

     \includegraphics[width=\mywidth]{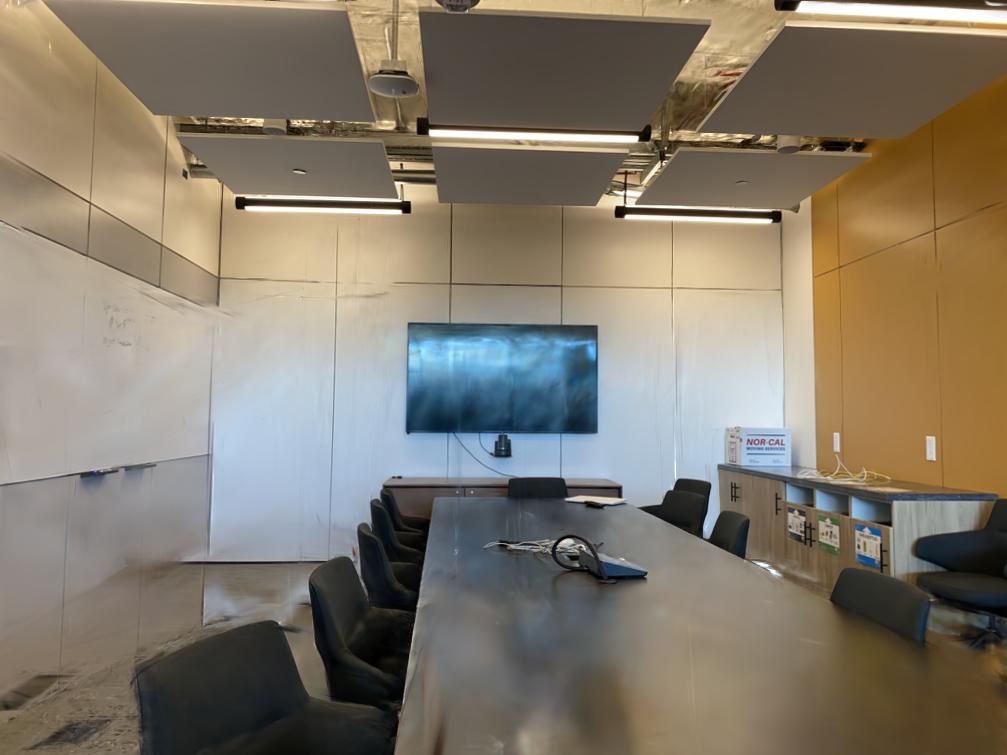}  &
     \includegraphics[width=\mywidth]{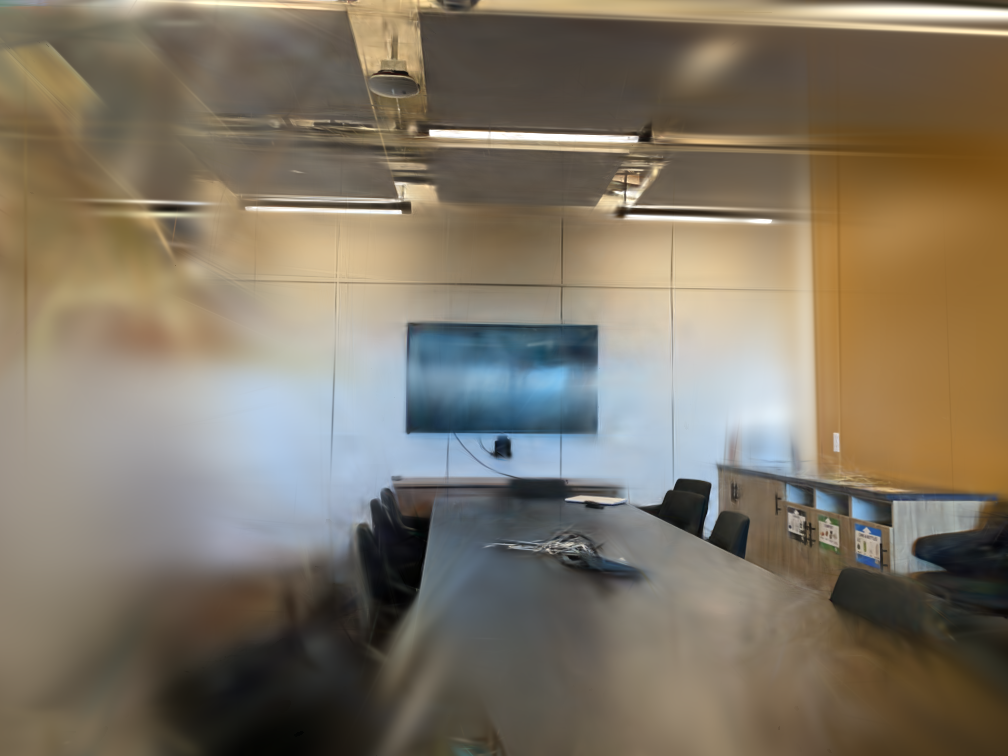} &
     \includegraphics[width=\mywidth]{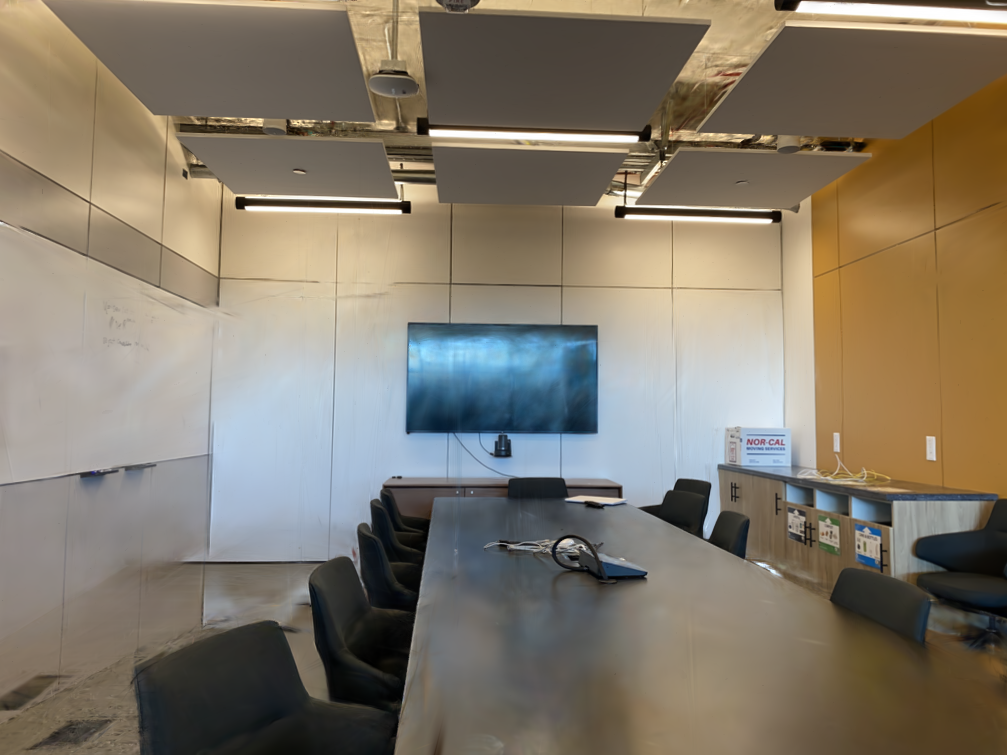} &
     \includegraphics[width=\mywidth]{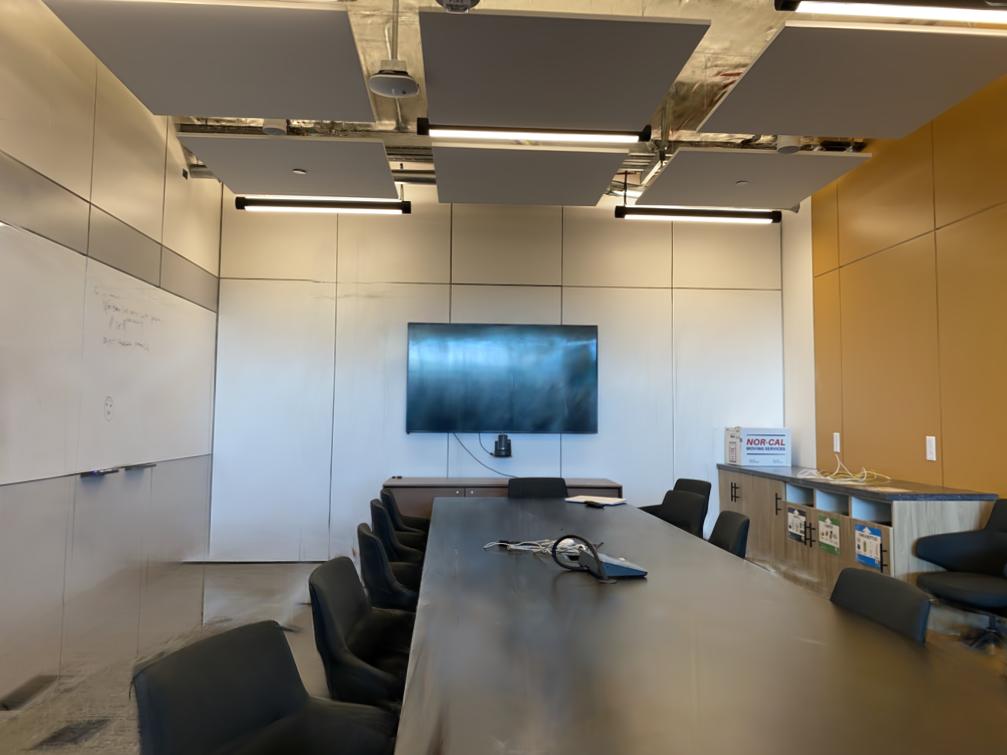} &
     \includegraphics[width=\mywidth]{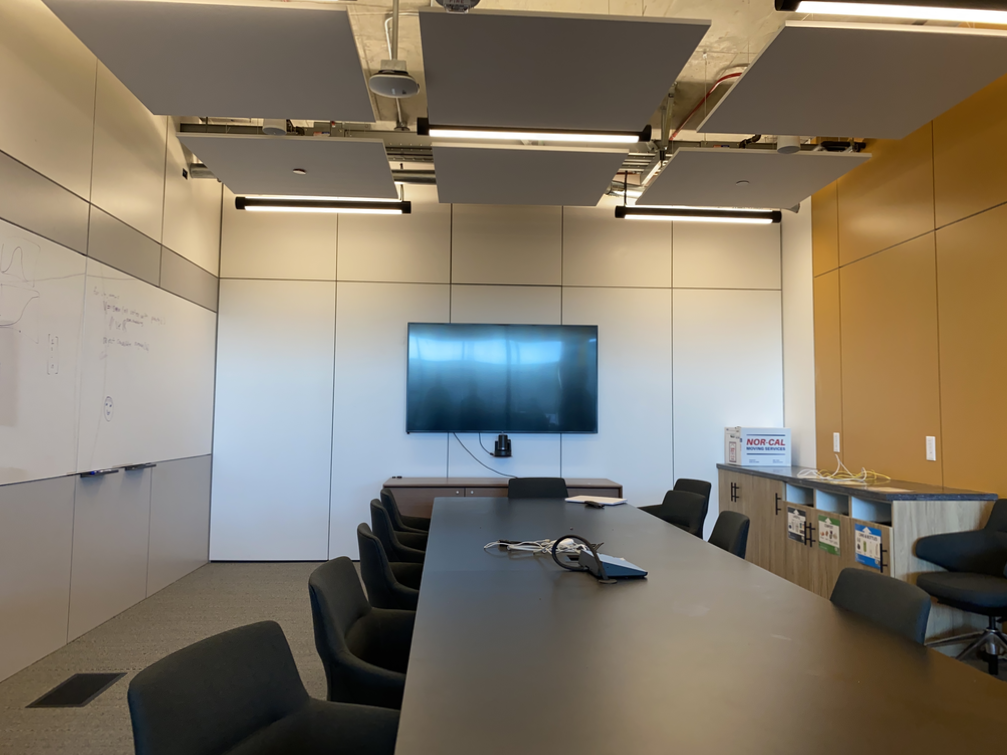} \\

     \includegraphics[width=\mywidth]{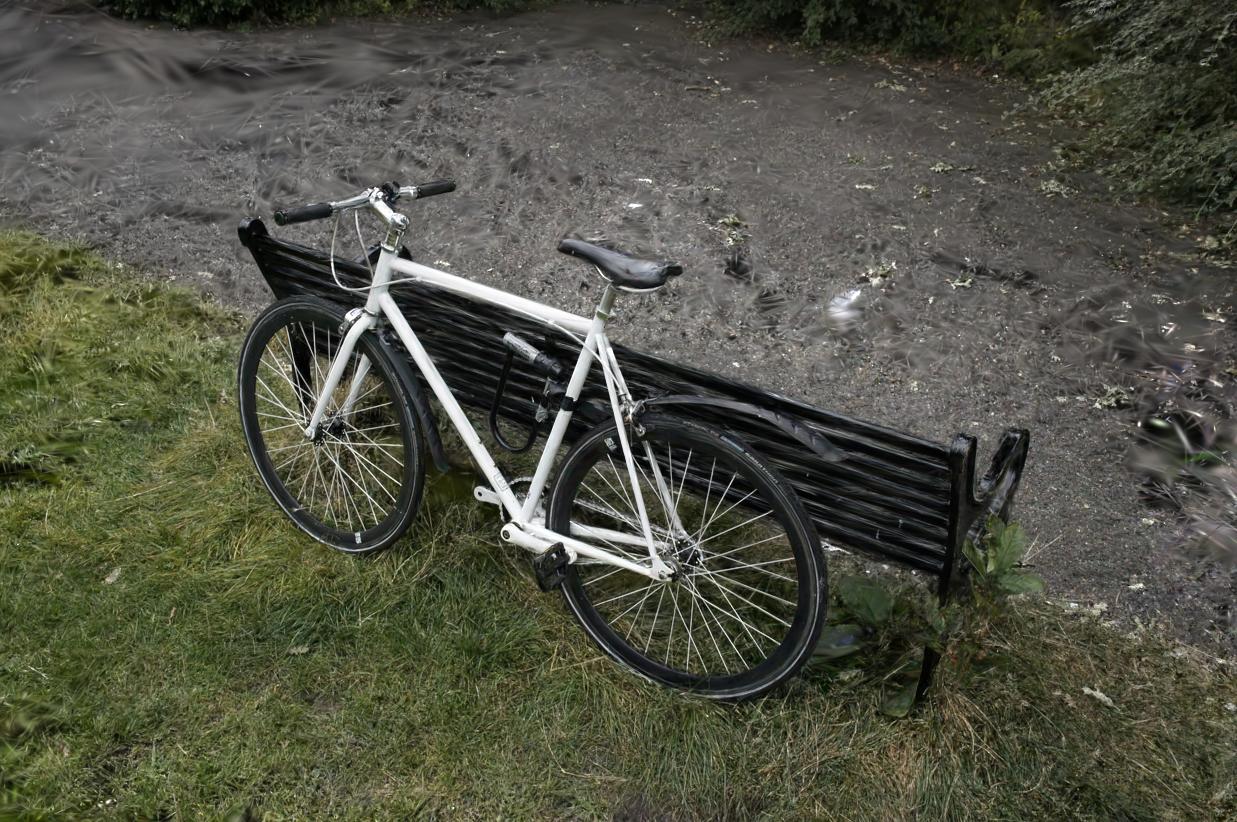}  &
     \includegraphics[width=\mywidth]{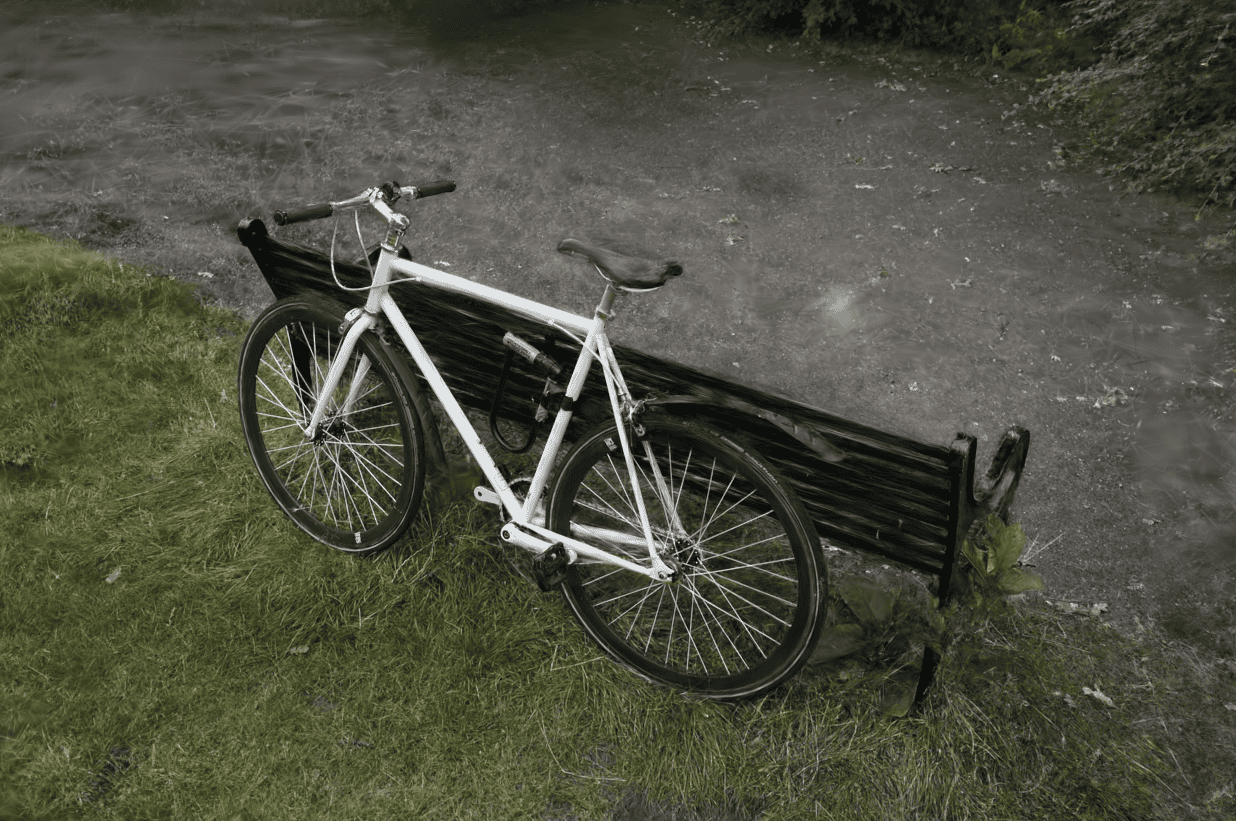} &
     \includegraphics[width=\mywidth]{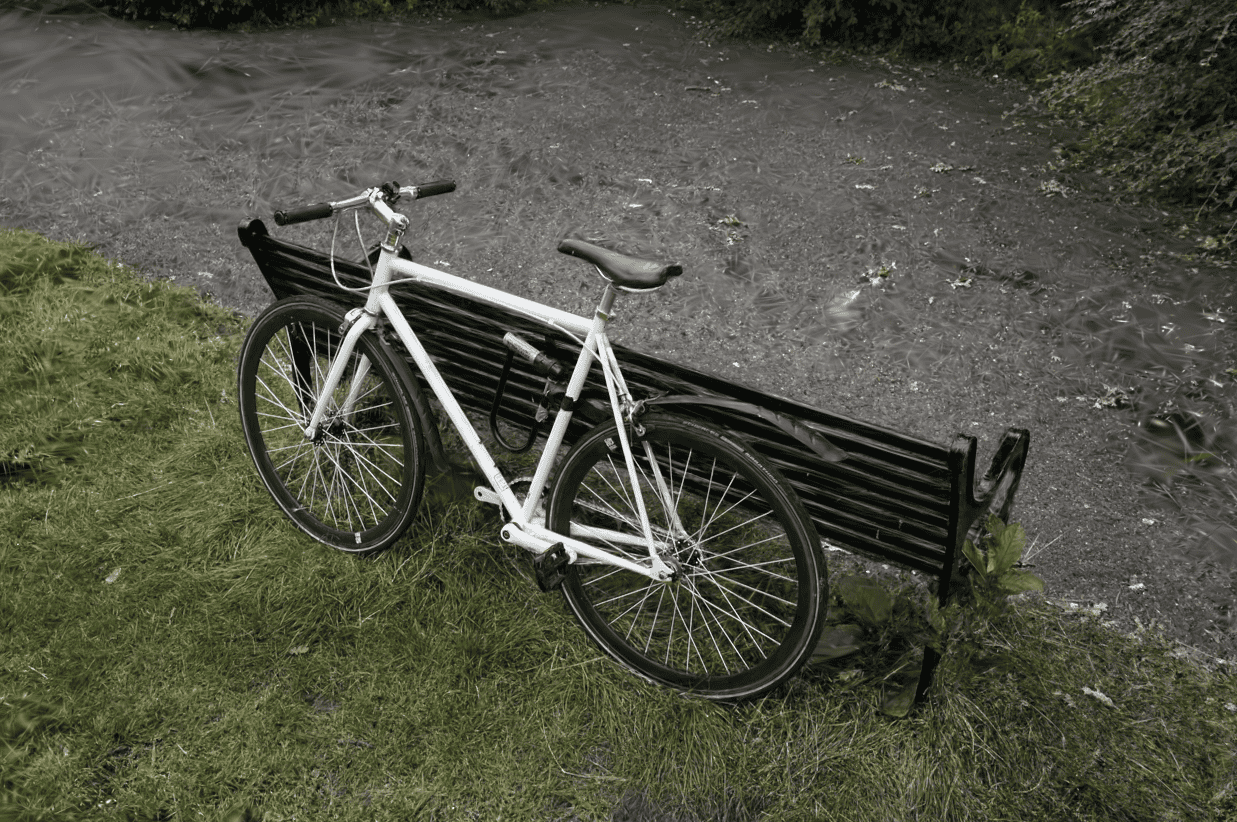} &
     \includegraphics[width=\mywidth]{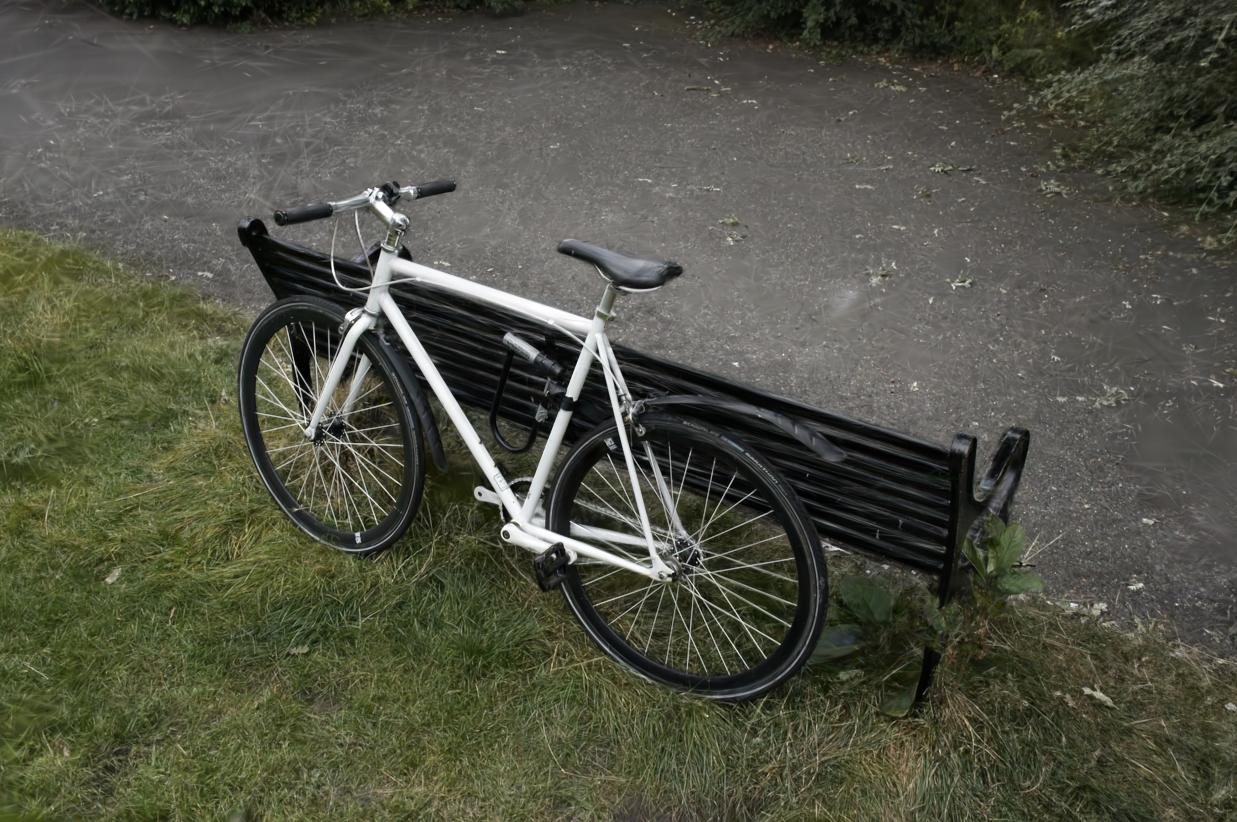} &
     \includegraphics[width=\mywidth]{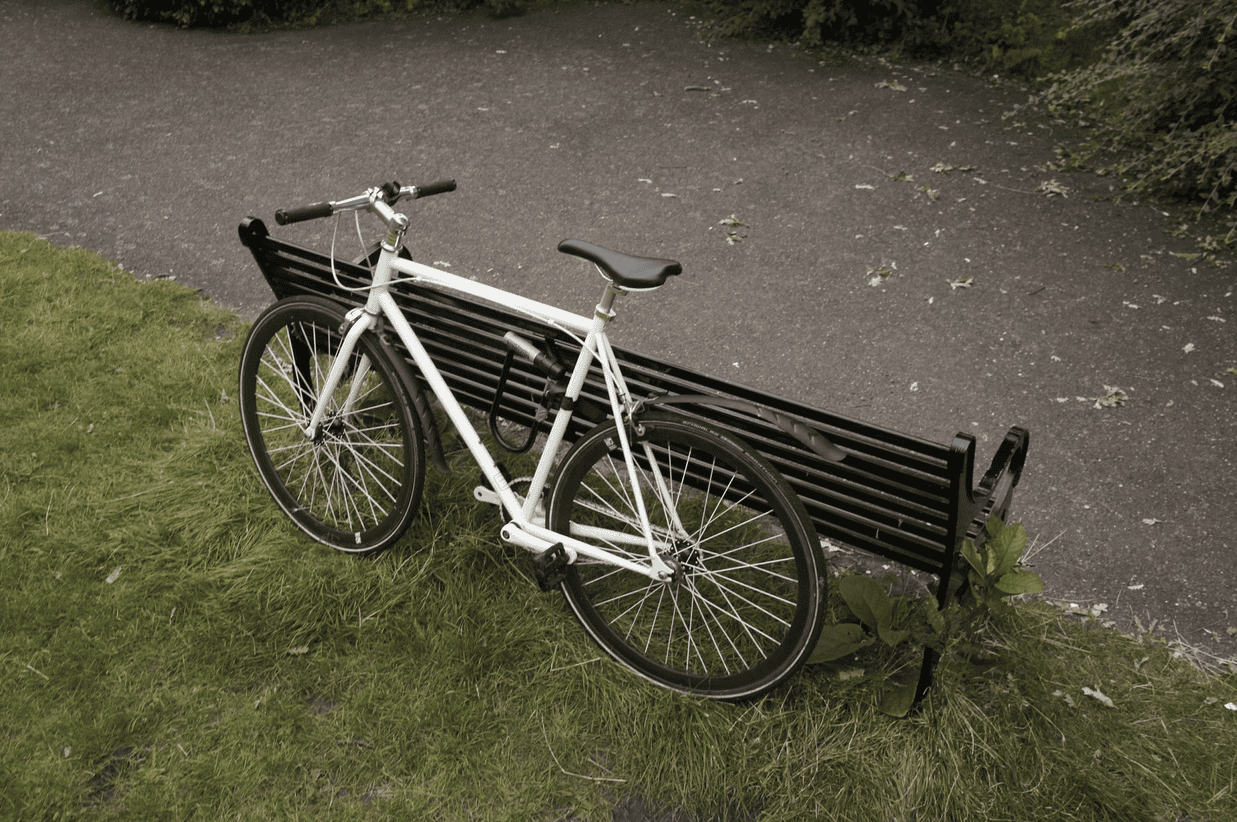} \\

     \includegraphics[width=\mywidth]{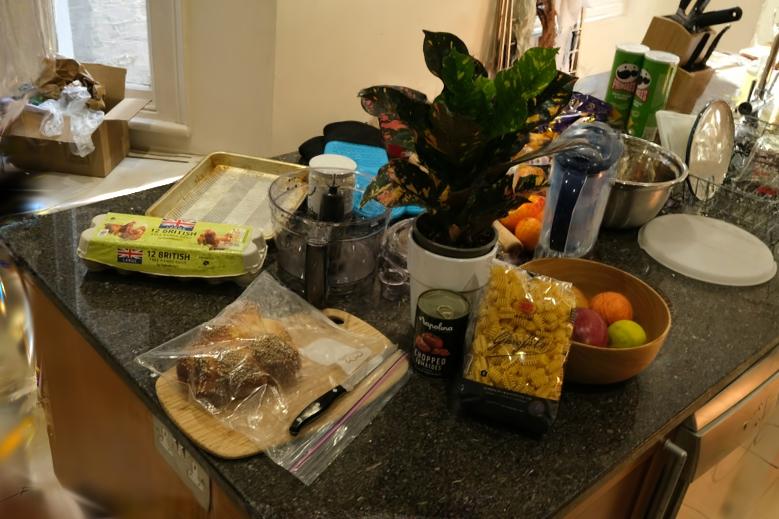}  &
     \includegraphics[width=\mywidth]{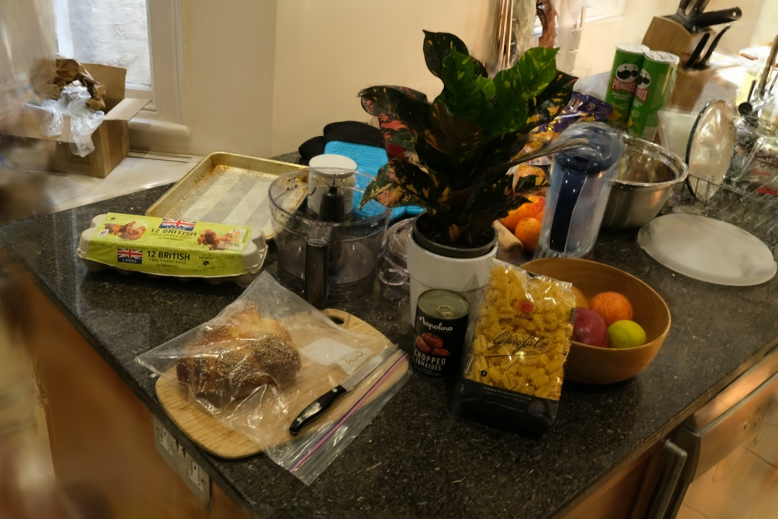} &
     \includegraphics[width=\mywidth]{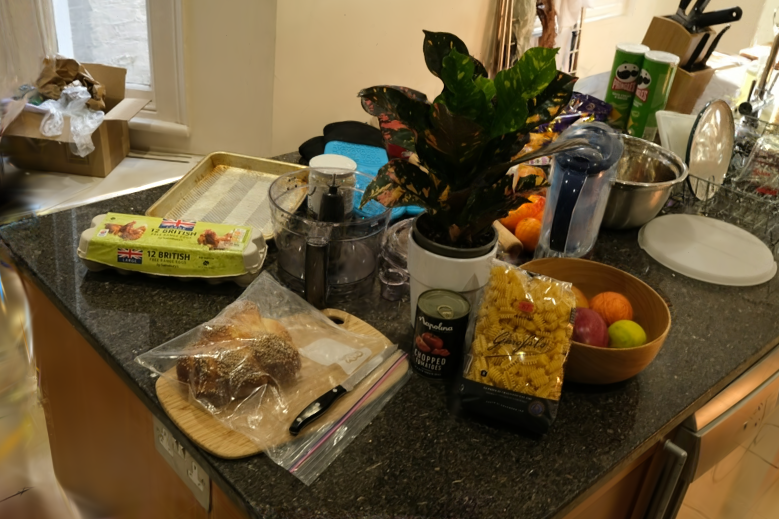} &
     \includegraphics[width=\mywidth]{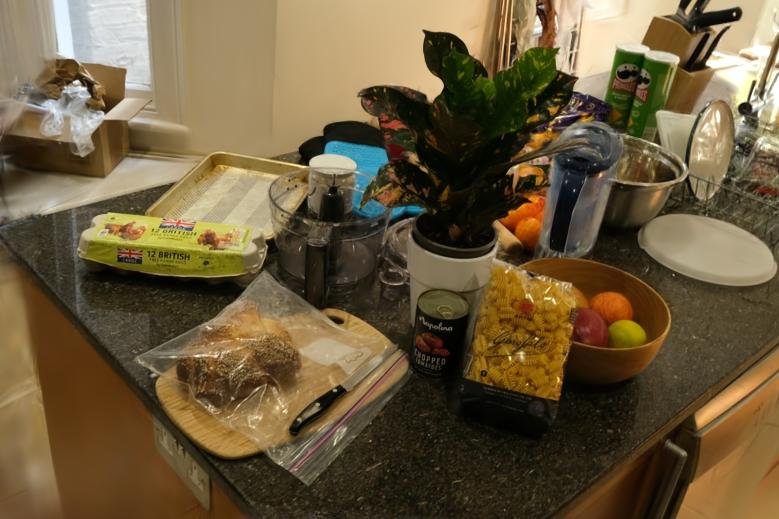} &
     \includegraphics[width=\mywidth]{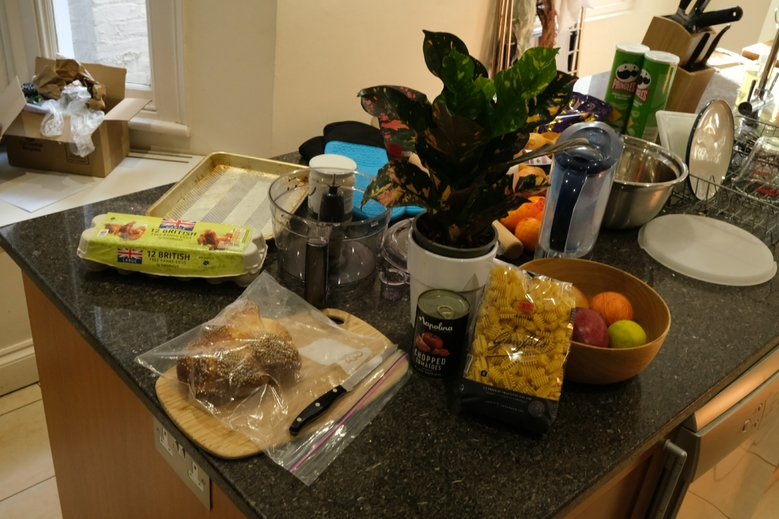} \\

     \includegraphics[width=\mywidth]{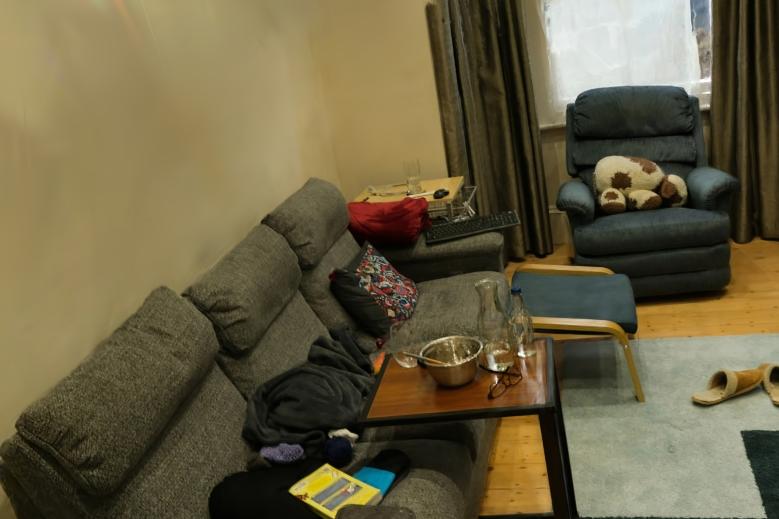}  &
     \includegraphics[width=\mywidth]{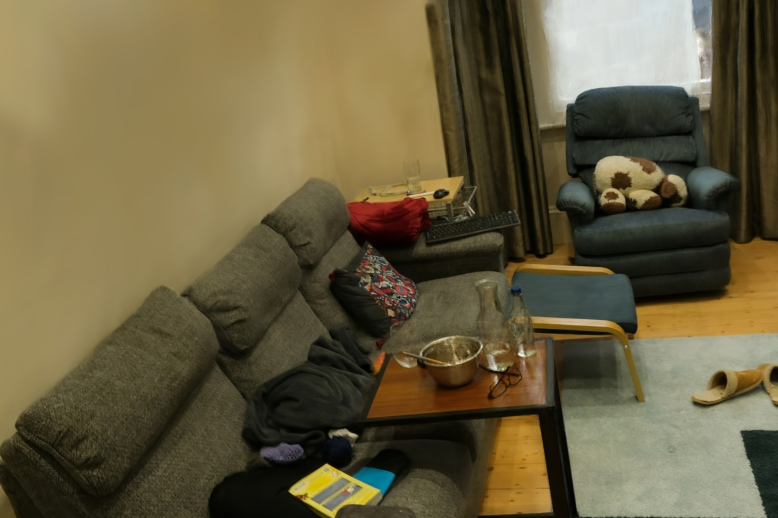} &
     \includegraphics[width=\mywidth]{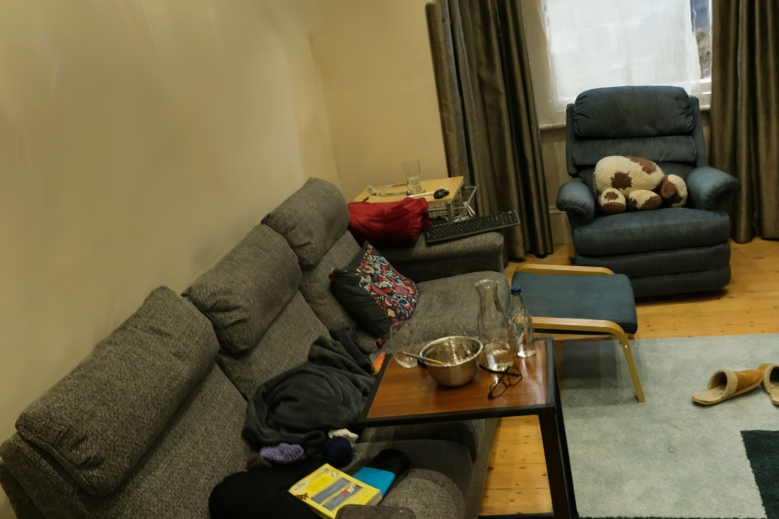} &
     \includegraphics[width=\mywidth]{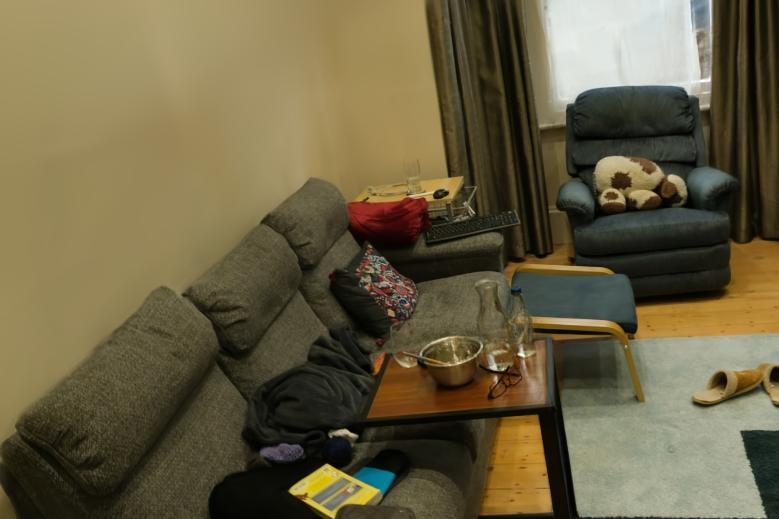} &
     \includegraphics[width=\mywidth]{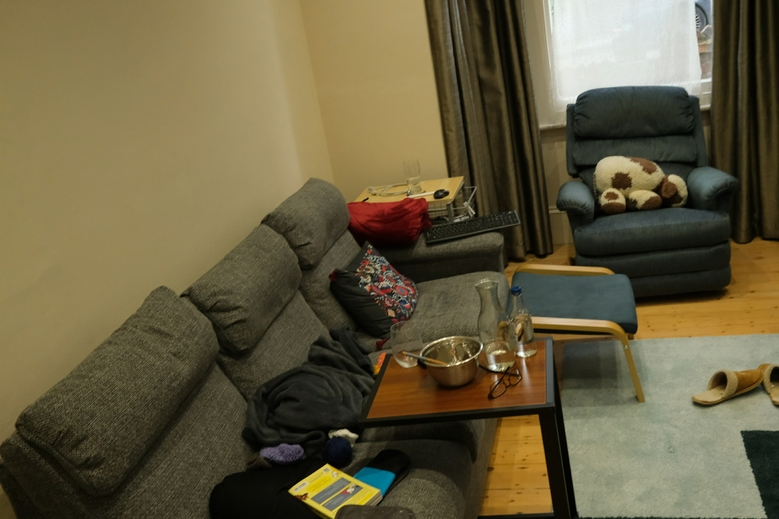} \\

     {3DGS} & {ViewExtrapolator \cite{liu2024novel}} & {Difix3D+\cite{wu2025difix3d+}} & \paperName & Ground Truth \\

     \includegraphics[width=\mywidth]{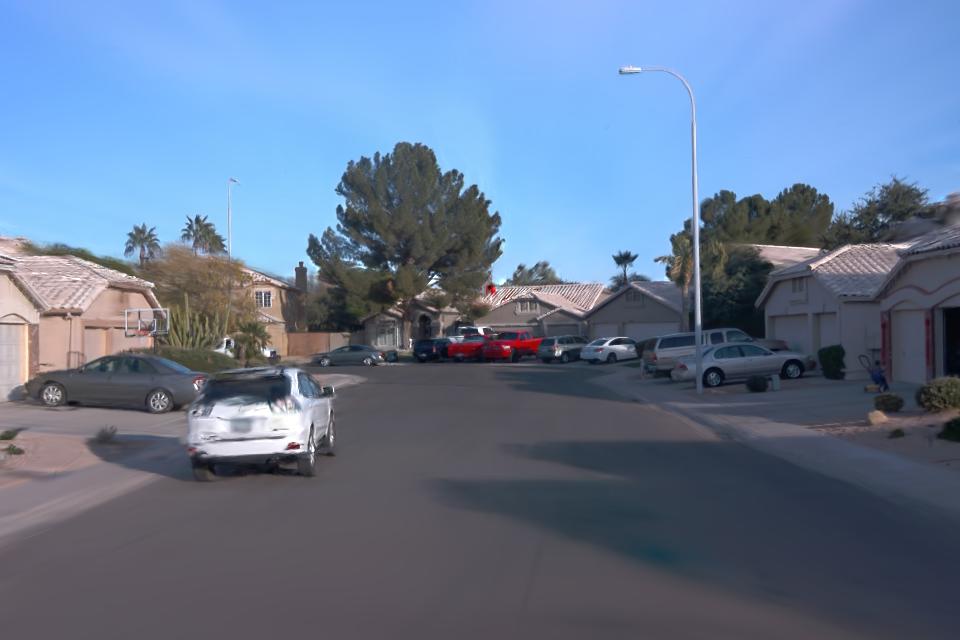}  &
     \includegraphics[width=\mywidth]{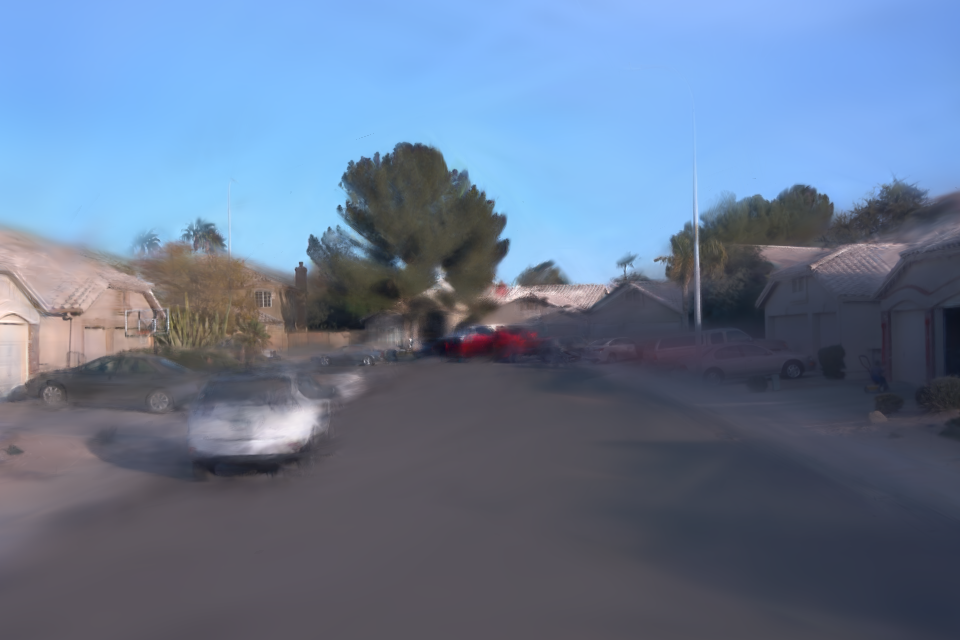} &
     \includegraphics[width=\mywidth]{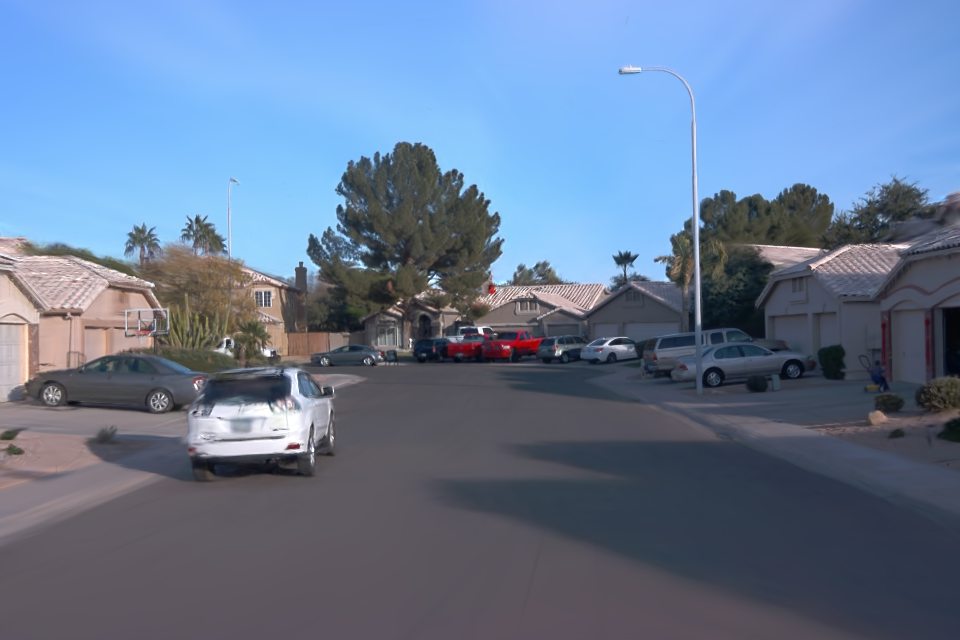} &
     \includegraphics[width=\mywidth]{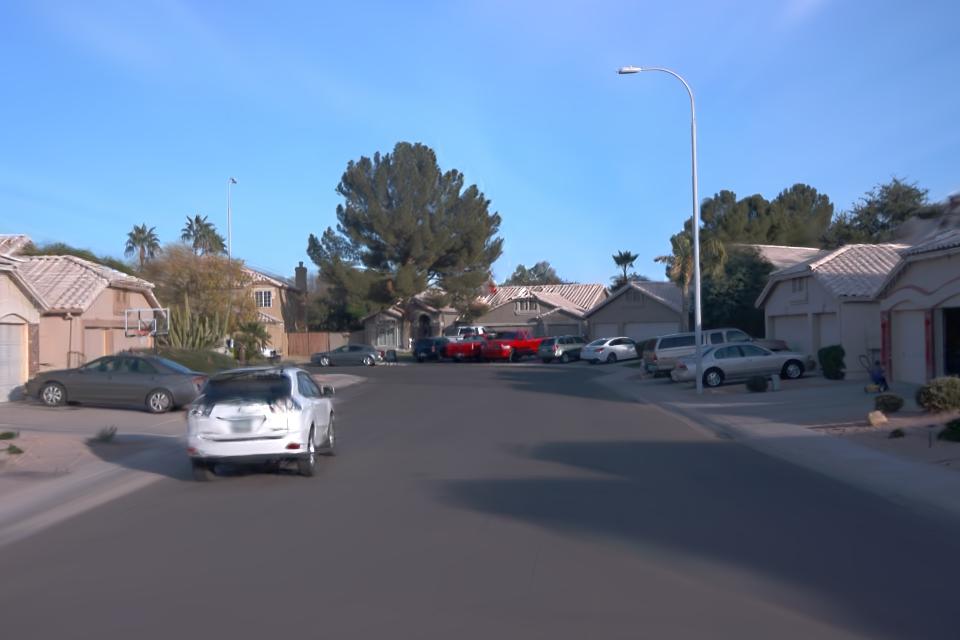} &
     \includegraphics[width=\mywidth]{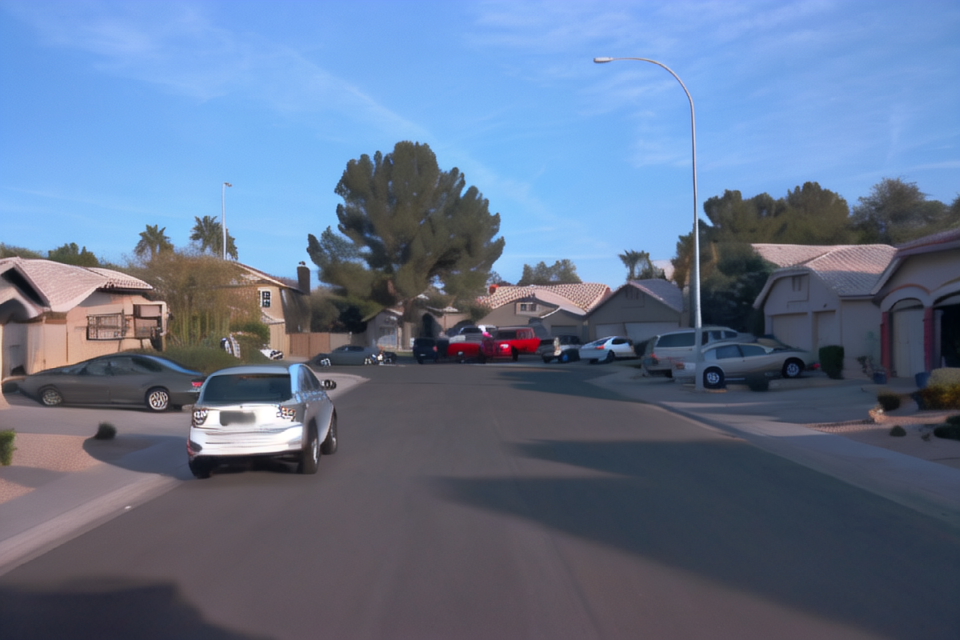} \\

     \includegraphics[width=\mywidth]{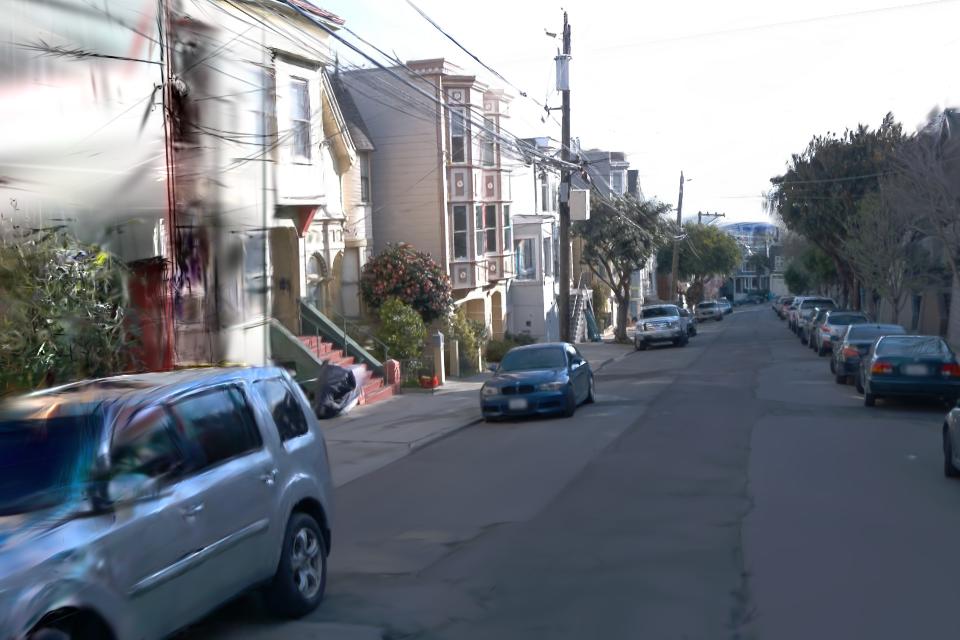}  &
     \includegraphics[width=\mywidth]{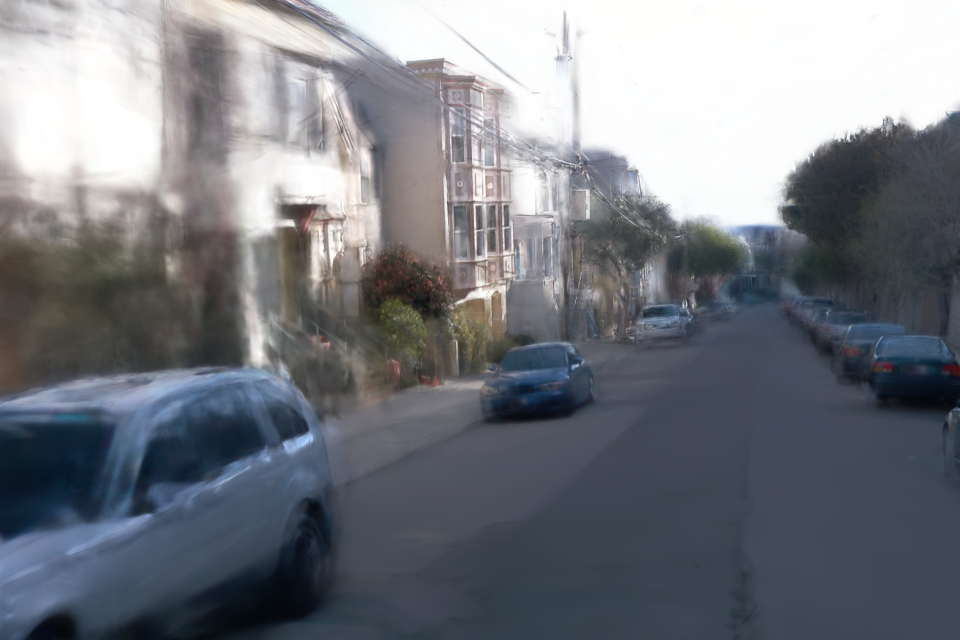} &
     \includegraphics[width=\mywidth]{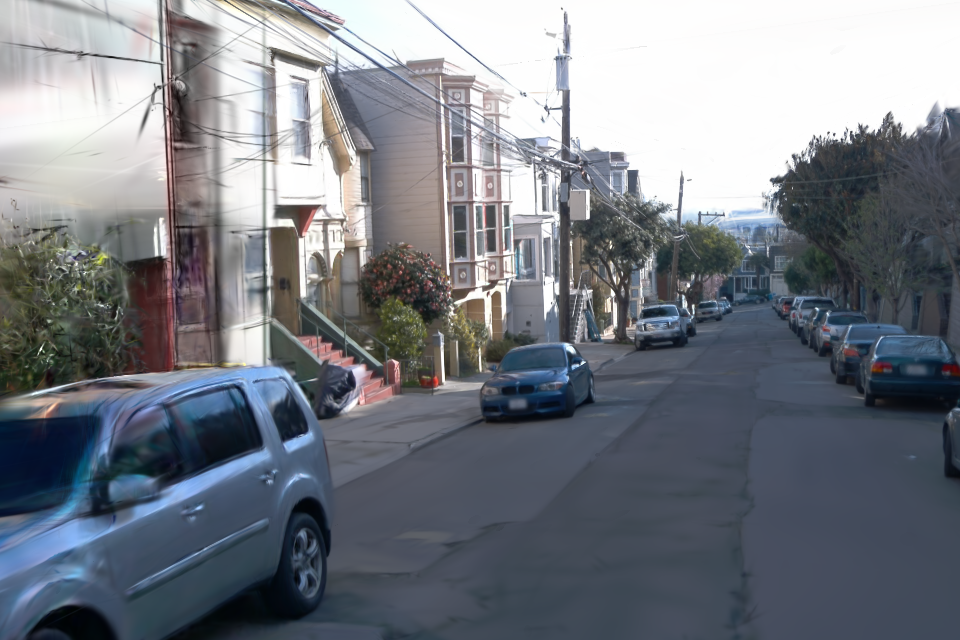} &
     \includegraphics[width=\mywidth]{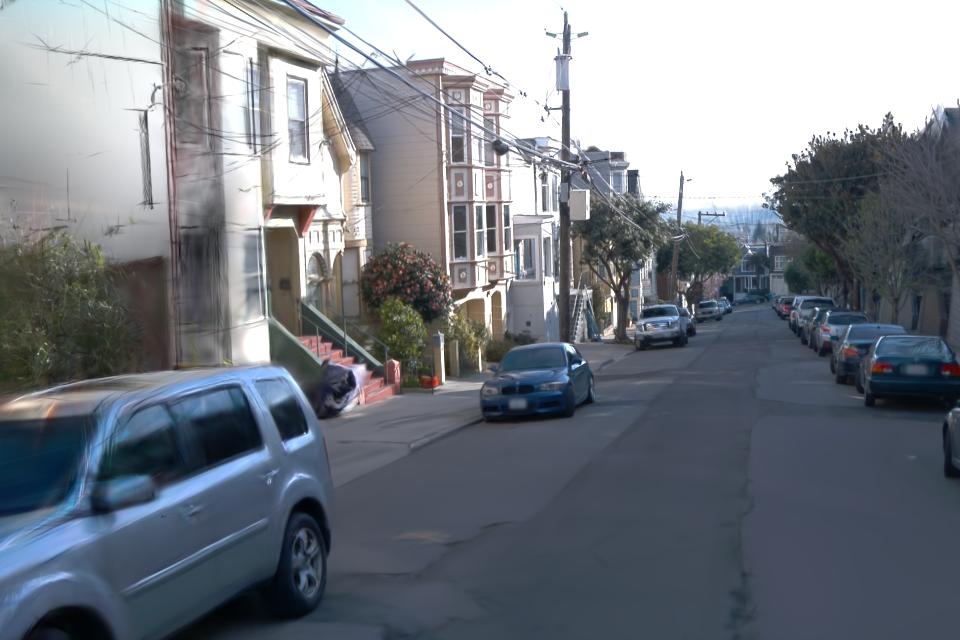} &
     \includegraphics[width=\mywidth]{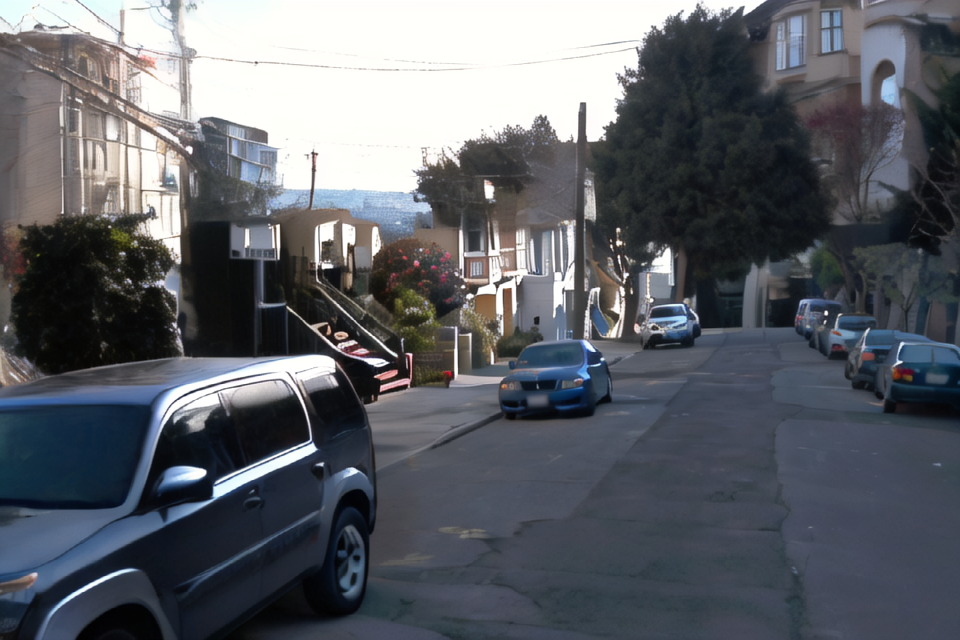} \\

     {3DGS} & {ViewExtrapolator \cite{liu2024novel}} & {Difix3D+\cite{wu2025difix3d+}} & \paperName & StreetCrafter \cite{yan2025streetcrafter}
     
     \end{tabular}
     \vspace{-0.2cm}
     \caption{\textbf{Additional Qualitative Comparisons}
     }
     \vspace{-0.4cm}
\label{fig: more_comp}
\end{figure*}

\vspace{-0.2cm}
We provide more qualitative comparisons in \cref{fig: more_comp}. The quantitative comparisons on each scene are shown in \cref{tab: llff_per_scene}, \cref{tab: mip_per_scene}, and \cref{tab: waymo_per_scene}. Additionally, \cref{fig: nvs} shows the quantitative comparisons between FreeFix and NVS-Solver \cite{you2024nvs}.

\definecolor{leafgreen}{rgb}{0.05, 0.54, 0.25}
\definecolor{orange}{rgb}{1.0, 0.6, 0.0}
\def \tick{\color{leafgreen}{\ding{52}}}
\def \cross{\color{red}{\ding{56}}}

\definecolor{tabfirst}{rgb}{1, 0.7, 0.7} %
\definecolor{tabsecond}{rgb}{1, 0.85, 0.7} %
\definecolor{tabthird}{rgb}{1, 1, 0.7} %

\begin{table*}[]
\centering
\small
\setlength{\tabcolsep}{2pt}
\begin{tabular}{@{\extracolsep{8pt}}llcccccc@{}} 
\toprule

& & 3DGS & FreeFix + Flux & FreeFix + SDXL & ViewExtrapolator \cite{liu2024novel} & NVS-Solver \cite{you2024nvs} & Difix3D+ \cite{wu2025difix3d+} \\

\midrule
\multirow{3}{*}{Fern} &
PSNR $\uparrow$ & 17.78 & \cellcolor{tabsecond}19.3 & \cellcolor{tabfirst}19.39 & \cellcolor{tabthird}18.63 & 12.65 & 18.5 \\

& SSIM $\uparrow$ & 0.603 & \cellcolor{tabsecond}0.656 & \cellcolor{tabfirst}0.658 & 0.619 & 0.375 & \cellcolor{tabthird}0.631 \\

& LPIPS $\downarrow$ & 0.289 & \cellcolor{tabfirst}0.243 & \cellcolor{tabsecond}0.245 & 0.3 & 0.551 & \cellcolor{tabthird}0.265 \\

\midrule
\multirow{3}{*}{Flower} &
PSNR $\uparrow$ & \cellcolor{tabthird}18.64 & \cellcolor{tabsecond}18.95 & 18.54 & 17.59 & 11.04 & \cellcolor{tabfirst}19.07 \\

& SSIM $\uparrow$ & 0.575 & \cellcolor{tabfirst}0.612 & \cellcolor{tabsecond}0.605 & 0.527 & 0.253 & \cellcolor{tabthird}0.594 \\

& LPIPS $\downarrow$ & \cellcolor{tabthird}0.265 & \cellcolor{tabsecond}0.254 & \cellcolor{tabthird}0.265 & 0.367 & 0.654 & \cellcolor{tabfirst}0.244 \\

\midrule
\multirow{3}{*}{Fortress} &
PSNR $\uparrow$ & 16.97 & \cellcolor{tabsecond}21.33 & \cellcolor{tabthird}20.32 & \cellcolor{tabfirst}21.97 & 12.8 & 17.87 \\

& SSIM $\uparrow$ & 0.689 & \cellcolor{tabfirst}0.751 & \cellcolor{tabsecond}0.729 & 0.702 & 0.387 & \cellcolor{tabthird}0.712 \\

& LPIPS $\downarrow$ & \cellcolor{tabthird}0.205 & \cellcolor{tabsecond}0.194 & 0.255 & 0.25 & 0.473 & \cellcolor{tabfirst}0.166 \\

\midrule
\multirow{3}{*}{Horns}
& PSNR$\uparrow$ & 16.76 & \cellcolor{tabfirst}19.06 & \cellcolor{tabsecond}18.95 & \cellcolor{tabthird}18.17 & 11.81 & 17.78 \\
& SSIM$\uparrow$ & 0.588 & \cellcolor{tabfirst}0.69 & \cellcolor{tabsecond}0.685 & 0.615 & 0.336 & \cellcolor{tabthird}0.63 \\
& LPIPS$\downarrow$ & 0.322 & \cellcolor{tabfirst}0.28 & \cellcolor{tabthird}0.3 & 0.36 & 0.588 & \cellcolor{tabsecond}0.294 \\

\midrule
\multirow{3}{*}{Leaves}
& PSNR$\uparrow$ & 14.6 & \cellcolor{tabsecond}16.51 & \cellcolor{tabfirst}16.63 & 14.49 & 9.94 & \cellcolor{tabthird}14.82 \\
& SSIM$\uparrow$ & 0.432 & \cellcolor{tabsecond}0.525 & \cellcolor{tabfirst}0.53 & 0.382 & 0.115 & \cellcolor{tabthird}0.438 \\
& LPIPS$\downarrow$ & \cellcolor{tabthird}0.303 & \cellcolor{tabsecond}0.222 & \cellcolor{tabfirst}0.22 & 0.333 & 0.636 & \cellcolor{tabthird}0.303 \\

\midrule
\multirow{3}{*}{Room}
& PSNR$\uparrow$ & 23.68 & \cellcolor{tabsecond}25.02 & \cellcolor{tabfirst}25.22 & 18.47 & 13.53 & \cellcolor{tabthird}24.67 \\
& SSIM$\uparrow$ & 0.868 & \cellcolor{tabfirst}0.9 & \cellcolor{tabfirst}0.9 & 0.782 & 0.609 & \cellcolor{tabthird}0.883 \\
& LPIPS$\downarrow$ & 0.196 & \cellcolor{tabfirst}0.143 & \cellcolor{tabsecond}0.146 & 0.457 & 0.465 & \cellcolor{tabthird}0.173 \\

\midrule
\multirow{3}{*}{Trex}
& PSNR$\uparrow$ & 18.27 & \cellcolor{tabfirst}20.7 & \cellcolor{tabsecond}20.45 & 18.53 & 12.15 & \cellcolor{tabthird}19.33 \\
& SSIM$\uparrow$ & 0.676 & \cellcolor{tabfirst}0.763 & \cellcolor{tabsecond}0.758 & 0.674 & 0.382 & \cellcolor{tabthird}0.721 \\
& LPIPS$\downarrow$ & 0.275 & \cellcolor{tabfirst}0.212 & \cellcolor{tabsecond}0.228 & 0.3 & 0.553 & \cellcolor{tabthird}0.229 \\

\bottomrule
\end{tabular}
\vspace{-0.2cm}
\caption{\textbf{Quantitative Comparison with Baselines for each scene in LLFF.} }

\vspace{-0.3cm}
\label{tab: llff_per_scene}
\end{table*}

\begin{figure*}
     \centering
     \small 
     \setlength{\tabcolsep}{0pt}
     \def\mywidth{4.3cm}
     \begin{tabular}{P{\mywidth}P{\mywidth}P{\mywidth}P{\mywidth}}
     
     \includegraphics[width=\mywidth]{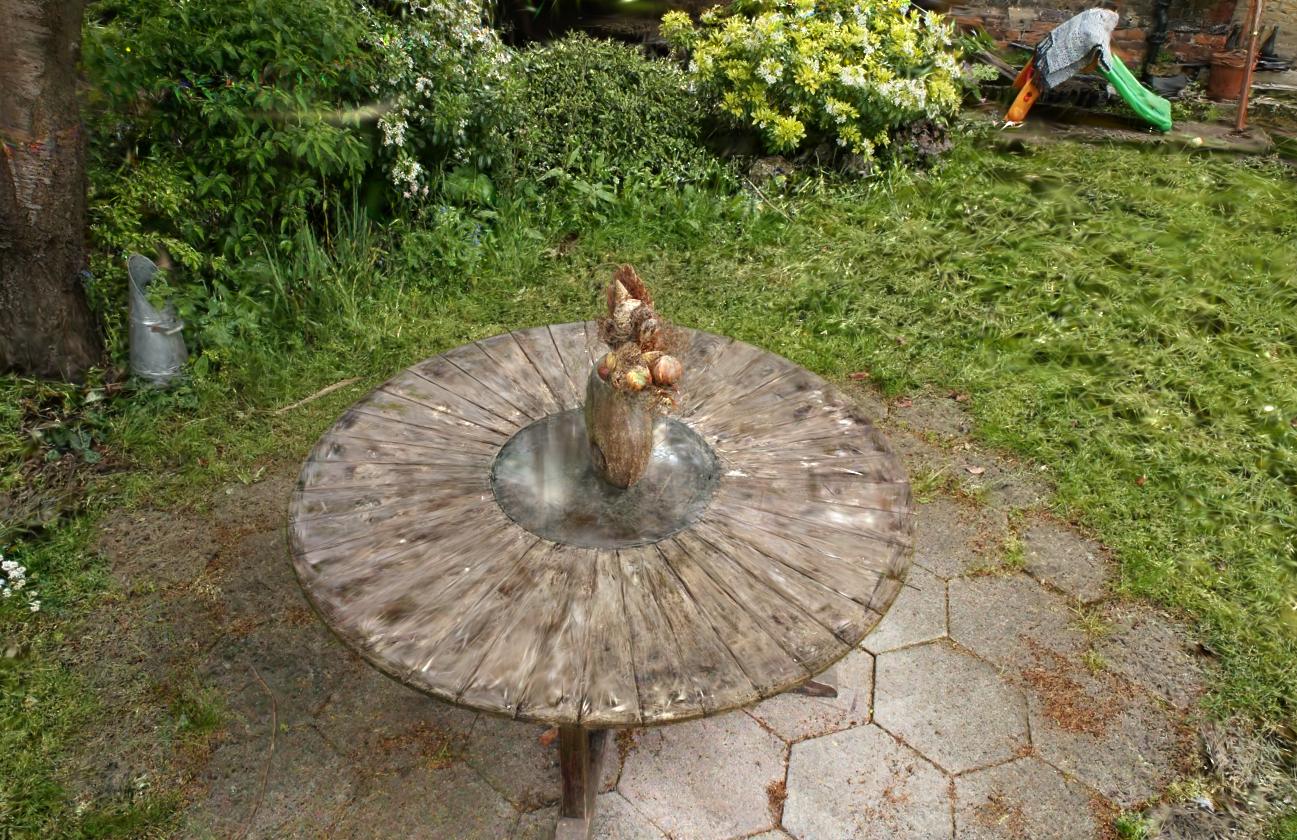}  &
     \includegraphics[width=\mywidth]{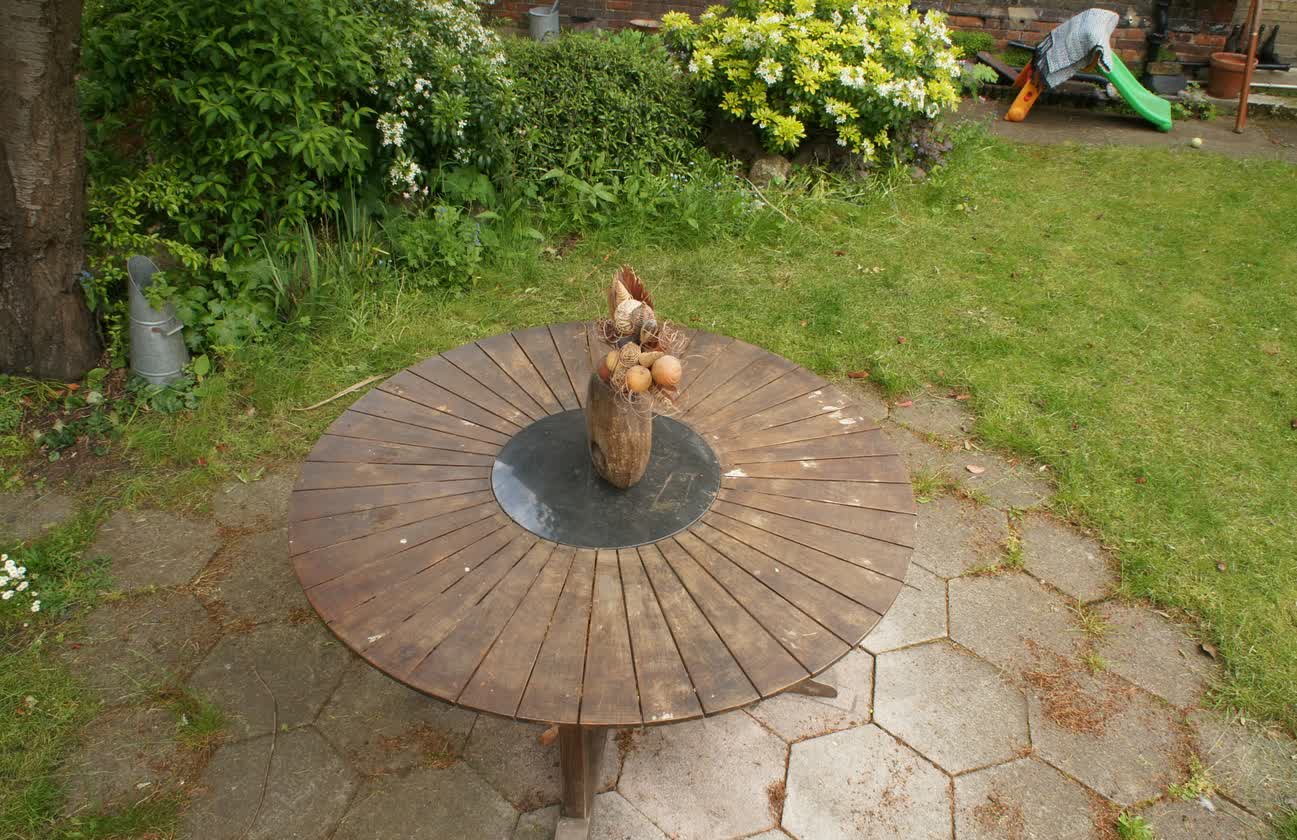}  &
     \includegraphics[width=\mywidth]{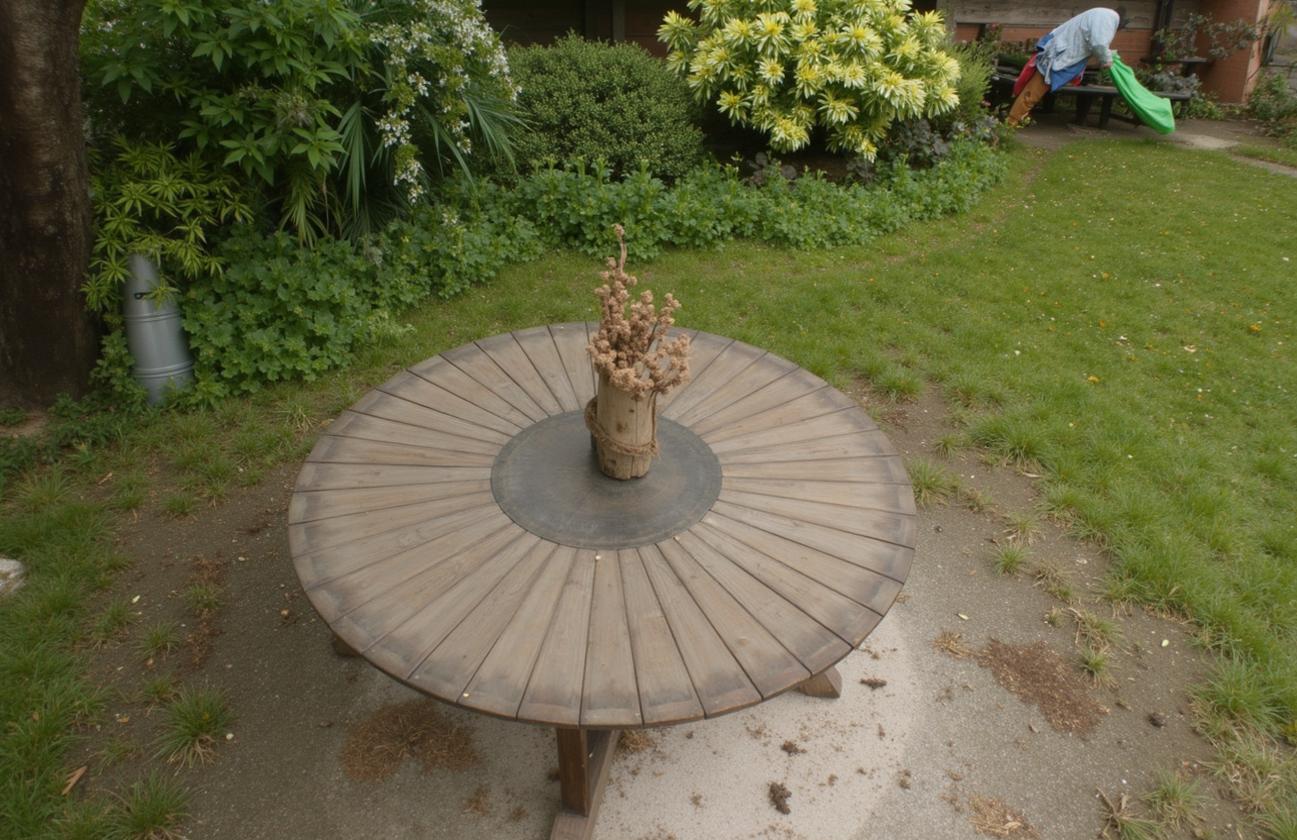}  &
     \includegraphics[width=\mywidth]{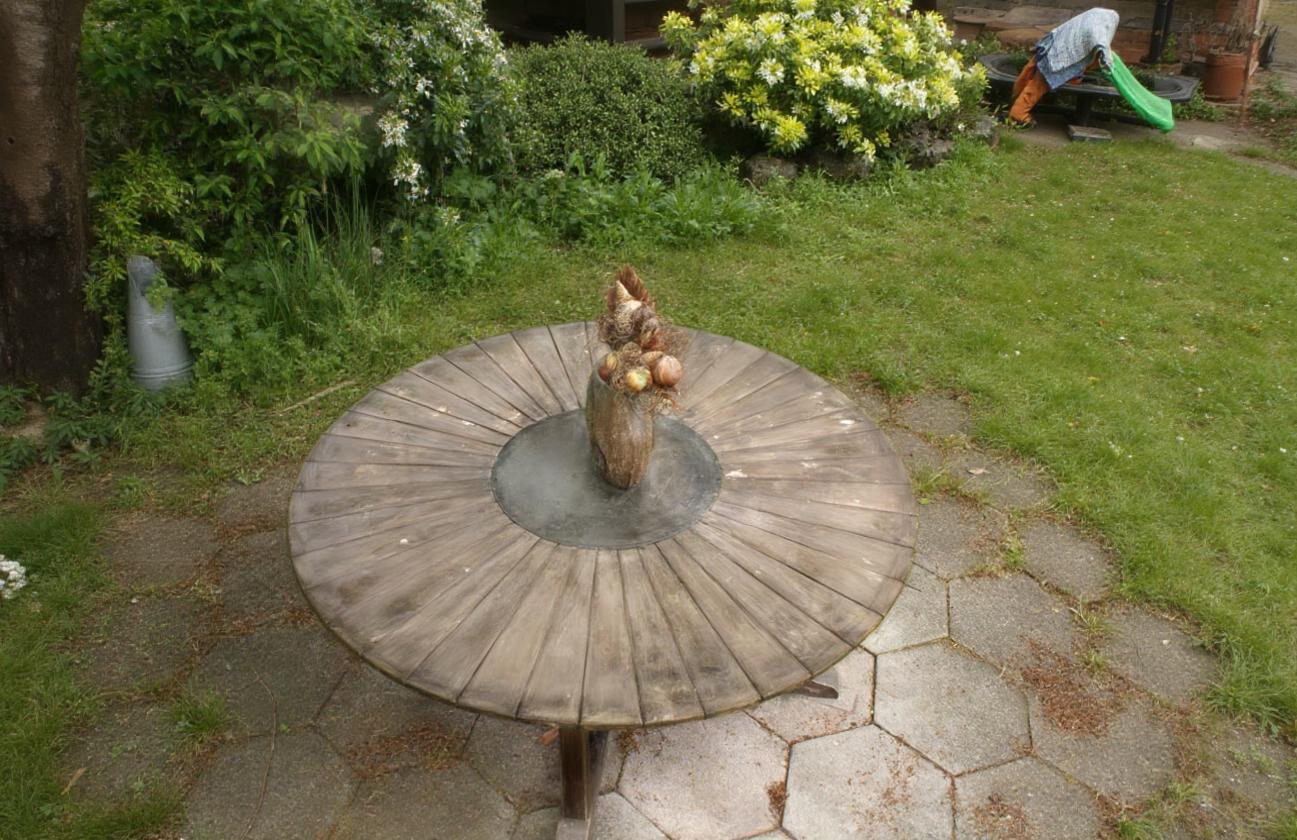}  \\

     \includegraphics[width=\mywidth]{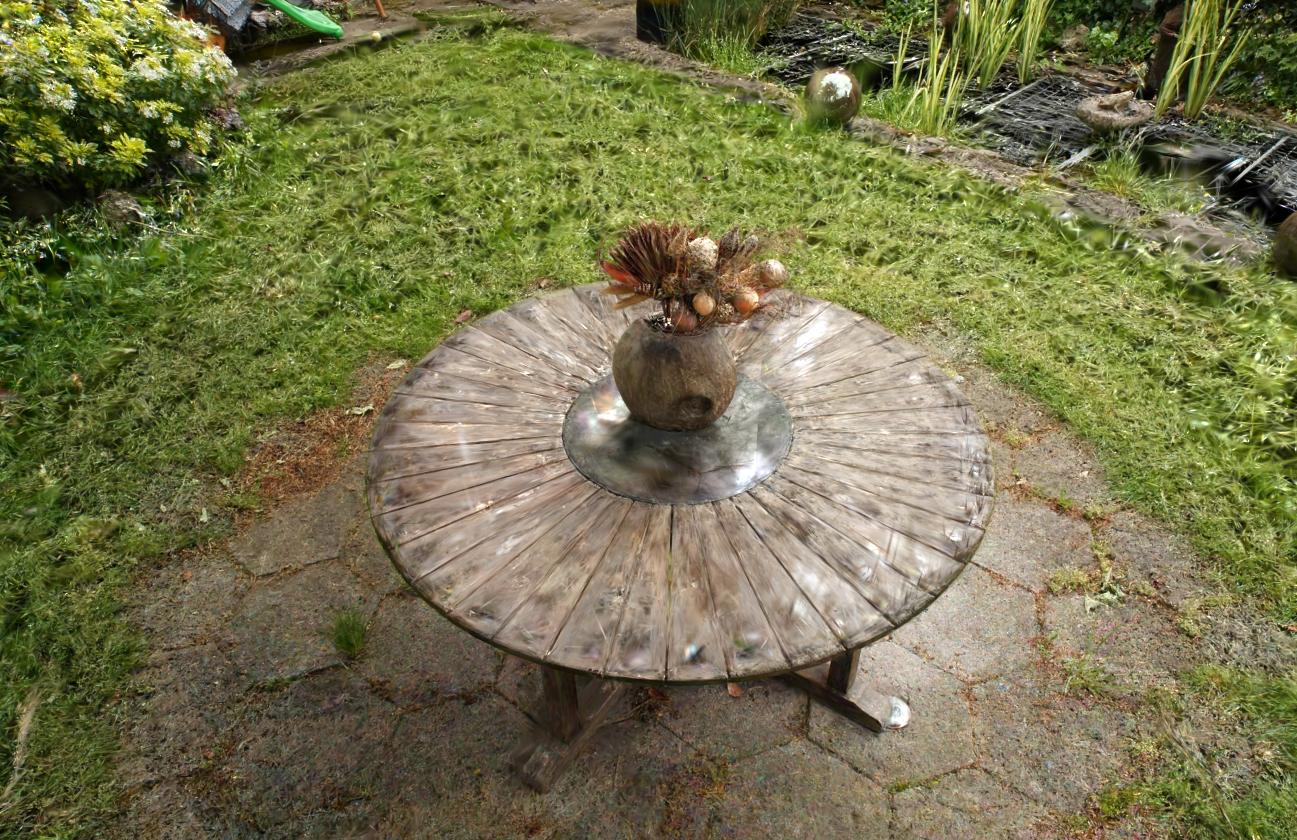}  &
     \includegraphics[width=\mywidth]{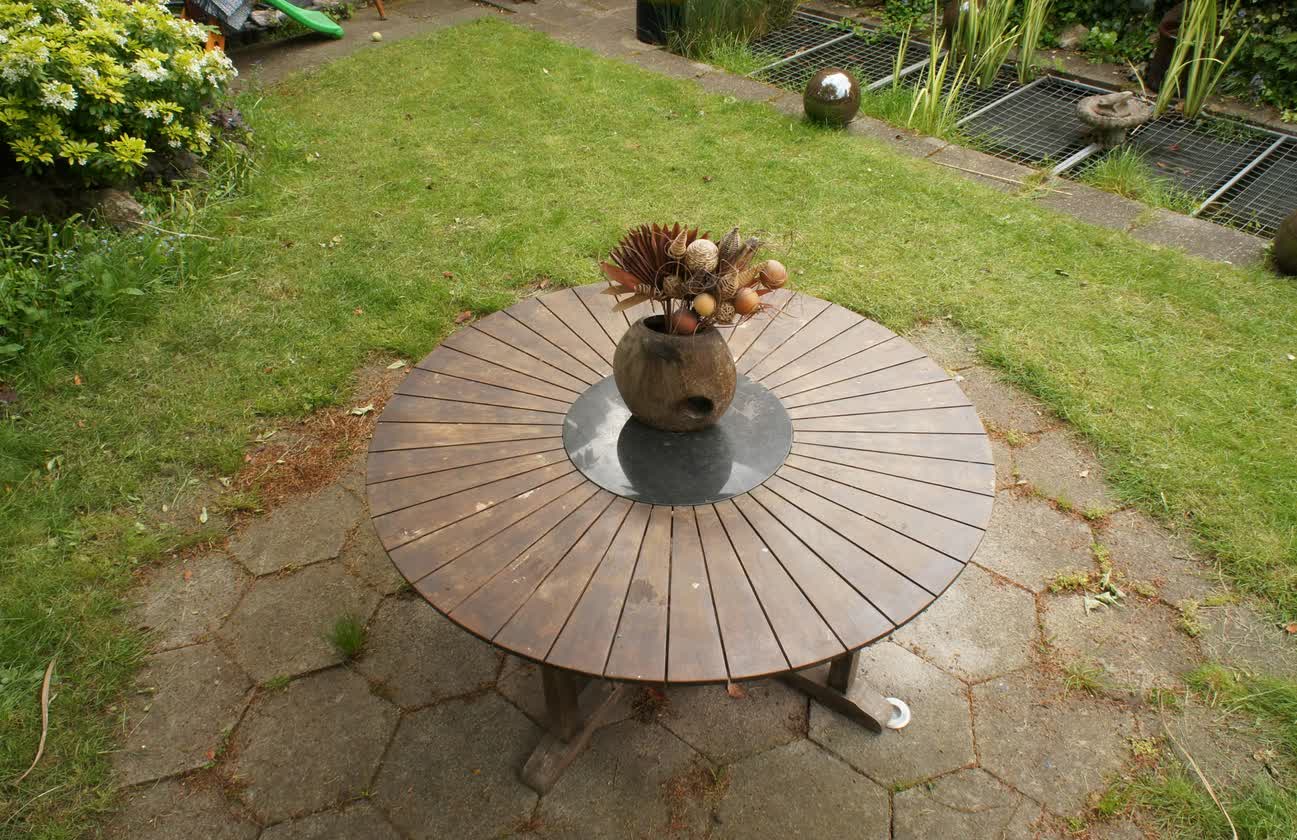}  &
     \includegraphics[width=\mywidth]{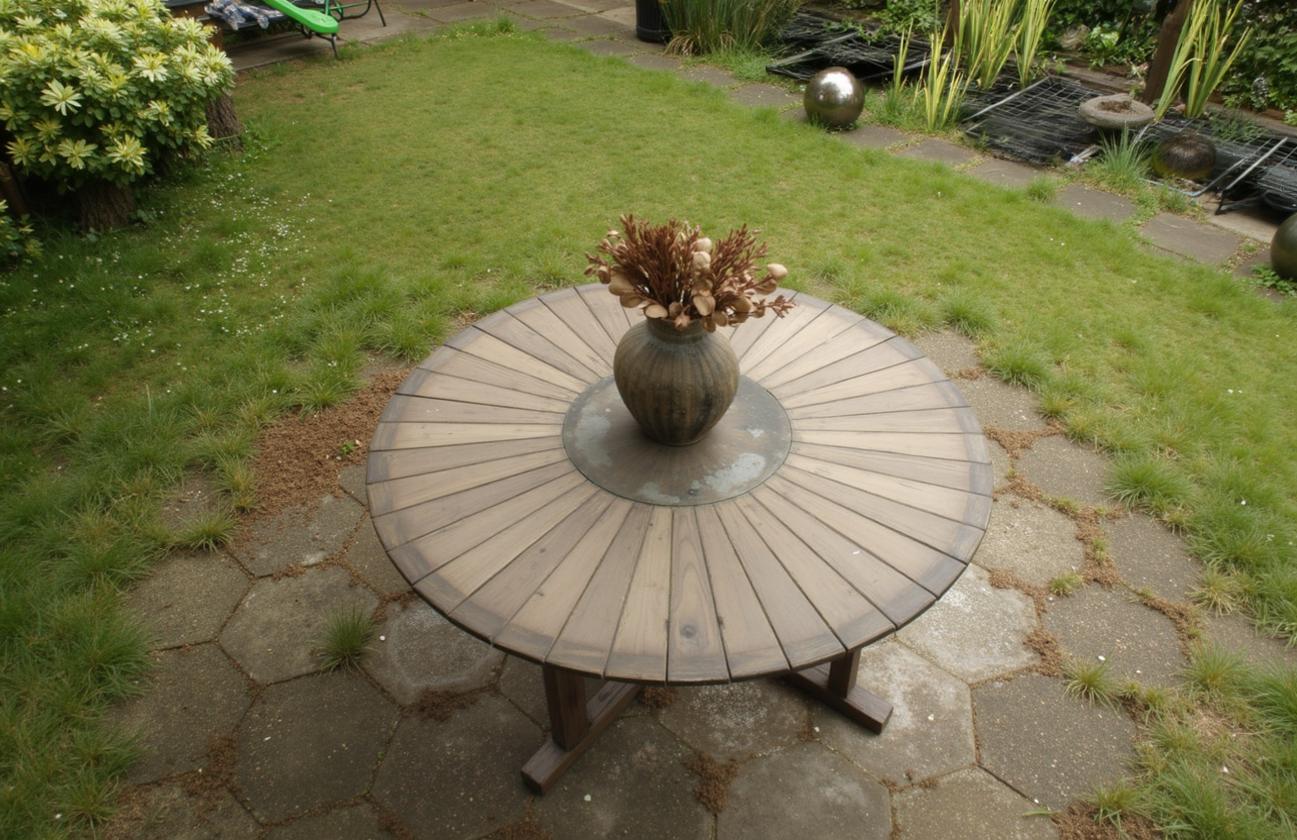}  &
     \includegraphics[width=\mywidth]{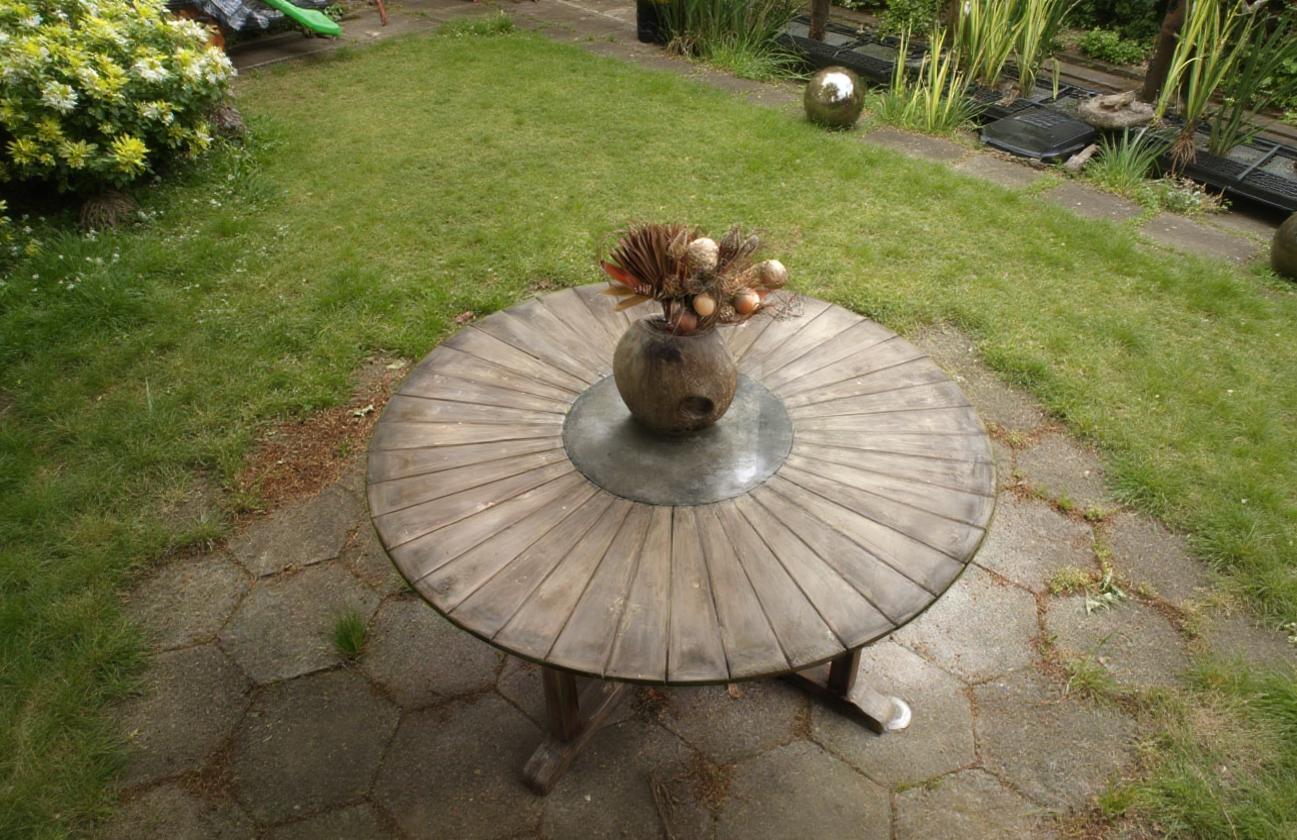}  \\

     3DGS & Ground Truth & Uncertainty & Certainty

     \end{tabular}
     \vspace{-0.2cm}
     \caption{\textbf{Generated Results Comparison between Uncertainty and Certainty as Guidance.}}
     \vspace{-0.6cm}
\label{fig: uncertain}
\end{figure*}

\definecolor{leafgreen}{rgb}{0.05, 0.54, 0.25}
\definecolor{orange}{rgb}{1.0, 0.6, 0.0}
\def \tick{\color{leafgreen}{\ding{52}}}
\def \cross{\color{red}{\ding{56}}}

\definecolor{tabfirst}{rgb}{1, 0.7, 0.7} %
\definecolor{tabsecond}{rgb}{1, 0.85, 0.7} %
\definecolor{tabthird}{rgb}{1, 1, 0.7} %

\begin{table*}[]
\centering
\small
\setlength{\tabcolsep}{2pt}
\begin{tabular}{@{\extracolsep{8pt}}llcccccc@{}} 
\toprule

& & 3DGS & FreeFix + Flux & FreeFix + SDXL & ViewExtrapolator \cite{liu2024novel} & NVS-Solver \cite{you2024nvs} & Difix3D+ \cite{wu2025difix3d+} \\

\midrule
\multirow{3}{*}{Bicycle} 
& PSNR$\uparrow$ & 20.71 & \cellcolor{tabfirst}22.61 & \cellcolor{tabsecond}22.48 & 20.0 & 14.58 & \cellcolor{tabthird}21.39 \\
& SSIM$\uparrow$ & 0.497 & \cellcolor{tabfirst}0.589 & \cellcolor{tabsecond}0.588 & 0.482 & 0.266 & \cellcolor{tabthird}0.519 \\
& LPIPS$\downarrow$ & 0.327 & \cellcolor{tabfirst}0.267 & \cellcolor{tabsecond}0.269 & 0.419 & 0.626 & \cellcolor{tabthird}0.293 \\

\midrule
\multirow{3}{*}{Bonsai} 
& PSNR$\uparrow$ & 23.68 & \cellcolor{tabfirst}24.5 & \cellcolor{tabthird}24.07 & 22.01 & 10.27 & \cellcolor{tabsecond}24.19 \\
& SSIM$\uparrow$ & 0.828 & \cellcolor{tabsecond}0.837 & \cellcolor{tabthird}0.829 & 0.725 & 0.221 & \cellcolor{tabfirst}0.841 \\
& LPIPS$\downarrow$ & 0.147 & \cellcolor{tabsecond}0.132 & \cellcolor{tabthird}0.14 & 0.205 & 0.632 & \cellcolor{tabfirst}0.128 \\

\midrule
\multirow{3}{*}{Counter} 
& PSNR$\uparrow$ & 22.2 & \cellcolor{tabfirst}23.29 & \cellcolor{tabsecond}23.06 & 22.01 & 10.56 & \cellcolor{tabthird}23.03 \\
& SSIM$\uparrow$ & 0.788 & \cellcolor{tabfirst}0.806 & \cellcolor{tabthird}0.803 & 0.762 & 0.281 & \cellcolor{tabfirst}0.806 \\
& LPIPS$\downarrow$ & 0.157 & \cellcolor{tabsecond}0.149 & \cellcolor{tabthird}0.152 & 0.199 & 0.65 & \cellcolor{tabfirst}0.137 \\

\midrule
\multirow{3}{*}{Garden} 
& PSNR$\uparrow$ & 18.38 & \cellcolor{tabfirst}19.72 & \cellcolor{tabsecond}19.42 & 17.86 & 12.41 & \cellcolor{tabthird}19.09 \\
& SSIM$\uparrow$ & 0.415 & \cellcolor{tabfirst}0.52 & \cellcolor{tabsecond}0.517 & 0.409 & 0.234 & \cellcolor{tabthird}0.449 \\
& LPIPS$\downarrow$ & 0.357 & \cellcolor{tabfirst}0.288 & \cellcolor{tabsecond}0.294 & 0.505 & 0.626 & \cellcolor{tabthird}0.305 \\

\midrule
\multirow{3}{*}{Kitchen} 
& PSNR$\uparrow$ & 22.58 & \cellcolor{tabfirst}23.97 & \cellcolor{tabthird}22.9 & 19.65 & 12.46 & \cellcolor{tabsecond}23.02 \\
& SSIM$\uparrow$ & 0.759 & \cellcolor{tabfirst}0.776 & \cellcolor{tabthird}0.765 & 0.586 & 0.296 & \cellcolor{tabsecond}0.773 \\
& LPIPS$\downarrow$ & 0.199 & \cellcolor{tabfirst}0.168 & \cellcolor{tabthird}0.18 & 0.396 & 0.618 & \cellcolor{tabsecond}0.172 \\

\midrule
\multirow{3}{*}{Room} 
& PSNR$\uparrow$ & 26.3 & \cellcolor{tabfirst}26.9 & \cellcolor{tabsecond}26.79 & 25.06 & 10.42 & \cellcolor{tabthird}26.7 \\
& SSIM$\uparrow$ & 0.87 & \cellcolor{tabfirst}0.884 & \cellcolor{tabsecond}0.88 & 0.813 & 0.345 & \cellcolor{tabthird}0.877 \\
& LPIPS$\downarrow$ & \cellcolor{tabthird}0.099 & \cellcolor{tabsecond}0.098 & 0.106 & 0.171 & 0.67 & \cellcolor{tabfirst}0.093 \\

\midrule
\multirow{3}{*}{Stump} 
& PSNR$\uparrow$ & 18.97 & \cellcolor{tabfirst}20.14 & \cellcolor{tabsecond}20.06 & 19.31 & 16.45 & \cellcolor{tabthird}19.6 \\
& SSIM$\uparrow$ & 0.343 & \cellcolor{tabfirst}0.415 & \cellcolor{tabsecond}0.414 & 0.356 & 0.222 & \cellcolor{tabthird}0.359 \\
& LPIPS$\downarrow$ & 0.386 & \cellcolor{tabsecond}0.351 & \cellcolor{tabthird}0.355 & 0.431 & 0.597 & \cellcolor{tabfirst}0.339 \\

\bottomrule
\end{tabular}
\vspace{-0.2cm}
\caption{\textbf{Quantitative Comparison with Baselines for each scene in Mip-NeRF 360.} }

\vspace{-0.3cm}
\label{tab: mip_per_scene}
\end{table*}

\definecolor{leafgreen}{rgb}{0.05, 0.54, 0.25}
\definecolor{orange}{rgb}{1.0, 0.6, 0.0}
\def \tick{\color{leafgreen}{\ding{52}}}
\def \cross{\color{red}{\ding{56}}}

\definecolor{tabfirst}{rgb}{1, 0.7, 0.7} %
\definecolor{tabsecond}{rgb}{1, 0.85, 0.7} %
\definecolor{tabthird}{rgb}{1, 1, 0.7} %

\begin{table*}[]
\centering
\small
\setlength{\tabcolsep}{2pt}
\begin{tabular}{@{\extracolsep{2pt}}lccccccc@{}} 
\toprule

& 3DGS & FreeFix + Flux & FreeFix + SDXL & ViewExtrapolator \cite{liu2024novel} & NVS-Solver \cite{you2024nvs} & Difix3D+ \cite{wu2025difix3d+} & StreetCrafter \cite{yan2025streetcrafter} \\

\midrule

Seq102751-Trans & 0.181 & \cellcolor{tabfirst}0.169 & \cellcolor{tabthird}0.176 & 0.242 & 0.282 & \cellcolor{tabsecond}0.173 & 0.225 \\

Seq134763-Rot & 0.133 & \cellcolor{tabthird}0.125 & 0.133 & 0.155 & 0.314 & \cellcolor{tabsecond}0.114 & \cellcolor{tabfirst}0.112 \\

Seq134763-Trans & 0.156 & \cellcolor{tabthird}0.144 & \cellcolor{tabfirst}0.134 & 0.184 & 0.213 & \cellcolor{tabsecond}0.142 & 0.178 \\

Seq143481-Rot & \cellcolor{tabthird}0.113 & \cellcolor{tabsecond}0.112 & \cellcolor{tabfirst}0.103 & 0.124 & 0.323 & 0.124 & 0.122 \\

Seq148697-Rot & 0.1 & \cellcolor{tabfirst}0.089 & \cellcolor{tabthird}0.094 & 0.175 & 0.281 & \cellcolor{tabfirst}0.089 & 0.124 \\

Seq177619-Rot & 0.214 & \cellcolor{tabthird}0.204 & 0.21 & \cellcolor{tabfirst}0.182 & 0.31 & \cellcolor{tabsecond}0.2 & 0.262 \\

Seq177619-Trans & \cellcolor{tabthird}0.187 & \cellcolor{tabsecond}0.182 & 0.197 & 0.192 & 0.296 & \cellcolor{tabfirst}0.163 & 0.192 \\

\bottomrule
\end{tabular}
\vspace{-0.2cm}
\caption{\textbf{Quantitative Comparison with Baselines for each scene in Waymo.} The metric in this table is KID $\downarrow$. }

\vspace{-0.3cm}
\label{tab: waymo_per_scene}
\end{table*}

\subsection{Uncertainty as Guidance}
\vspace{-0.2cm}
In this paper, we apply certainty as guidance during denoising. In this subsection, we provide a comparison between using the uncertainty mask from \cite{jiang2024fisherrf} as guidance and our certainty mask as guidance. Specifically, for rendered uncertain masks $\mc{M}^{\bar{c}}$, we use $1 - \mc{M}^{\bar{c}}$ as guidance to experiment on Garden in Mip-NeRF 360. As shown in \cref{fig: uncertain} and \cref{tab: uncertain}, the images generated using uncertainty masks as guidance exhibit significant inconsistency, resulting in less satisfying performance.

\begin{table}
\centering
\small 
\begin{tabular}{lcccc}
\toprule 
& PSNR$\uparrow$ & SSIM$\uparrow$  & LPIPS$\downarrow$ \\
\cline{2-4}
Uncertainty Mask & 19.30 & 0.515 & 0.310 \\
Certainty Mask & \textbf{19.72} & \textbf{0.520} & \textbf{0.287} \\ 
\bottomrule
\end{tabular}
\vspace{-0.2cm}
\caption{\textbf{Quantitative Comparison between Uncertainty and Certainty as Guidance}. }
\label{tab: uncertain}
\vspace{-0.4cm}
\end{table}

\subsection{Ablation on Affine Transform}
\begin{figure}
     \centering
     \small 
     \setlength{\tabcolsep}{0pt}
     \def\mywidth{4.2cm}
     \begin{tabular}{P{\mywidth}P{\mywidth}}
     
     \includegraphics[width=\mywidth]{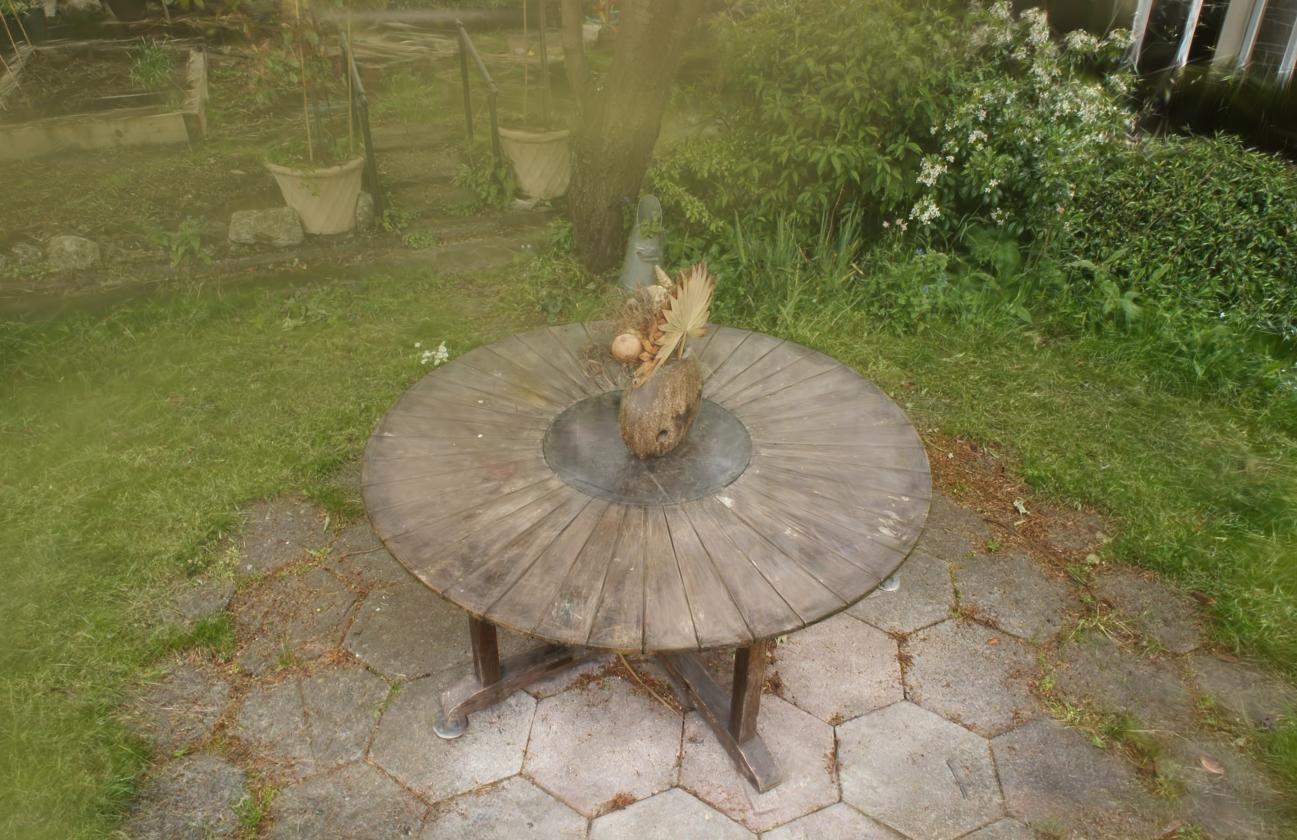}  &
     \includegraphics[width=\mywidth]{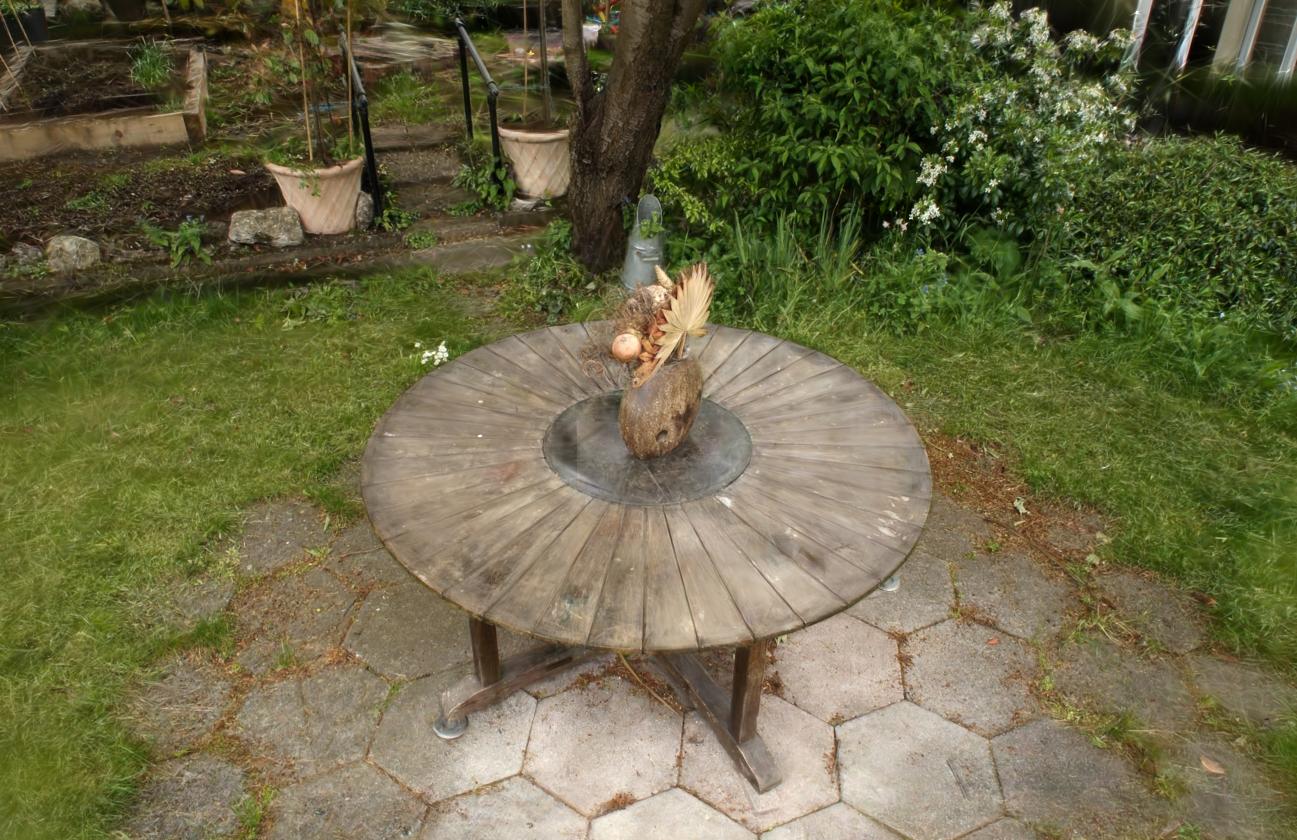}  \\
     
     \includegraphics[width=\mywidth]{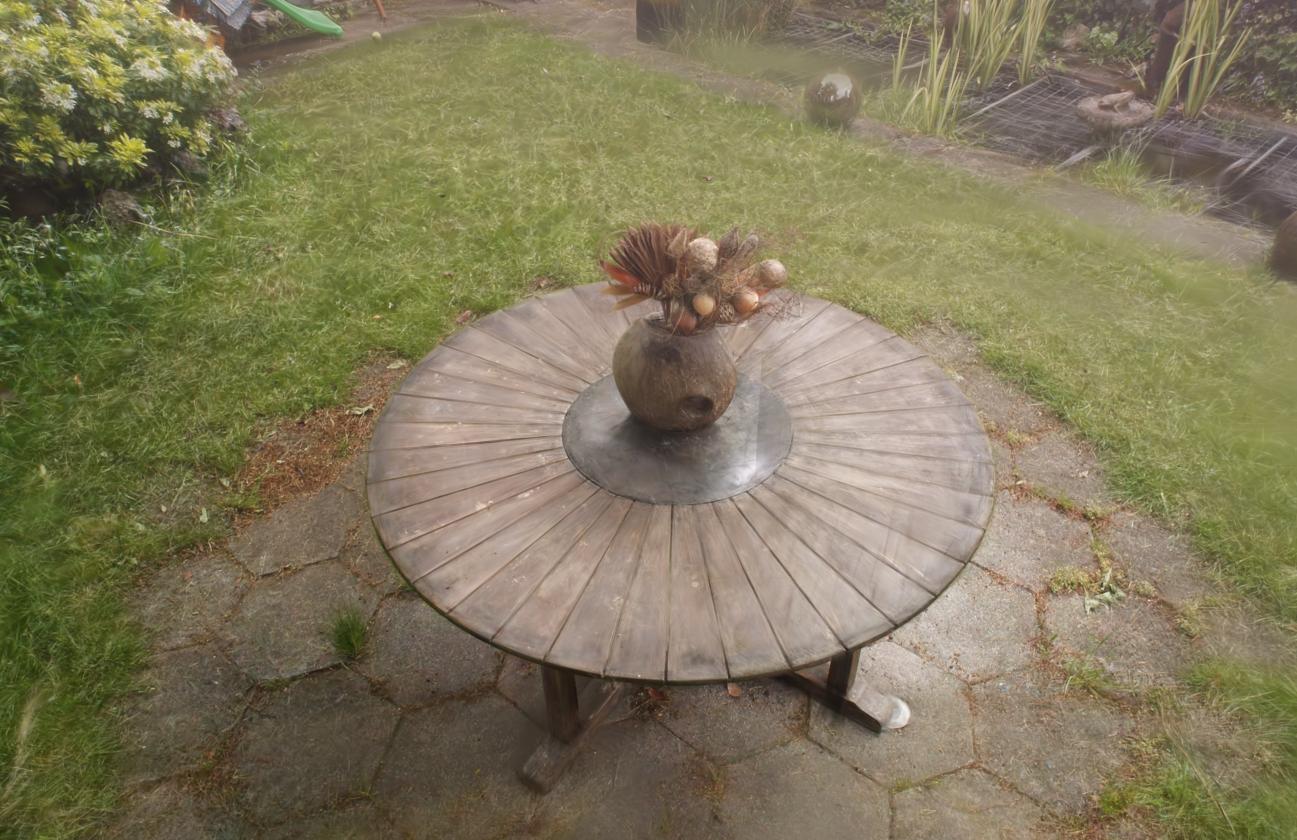} &
     \includegraphics[width=\mywidth]{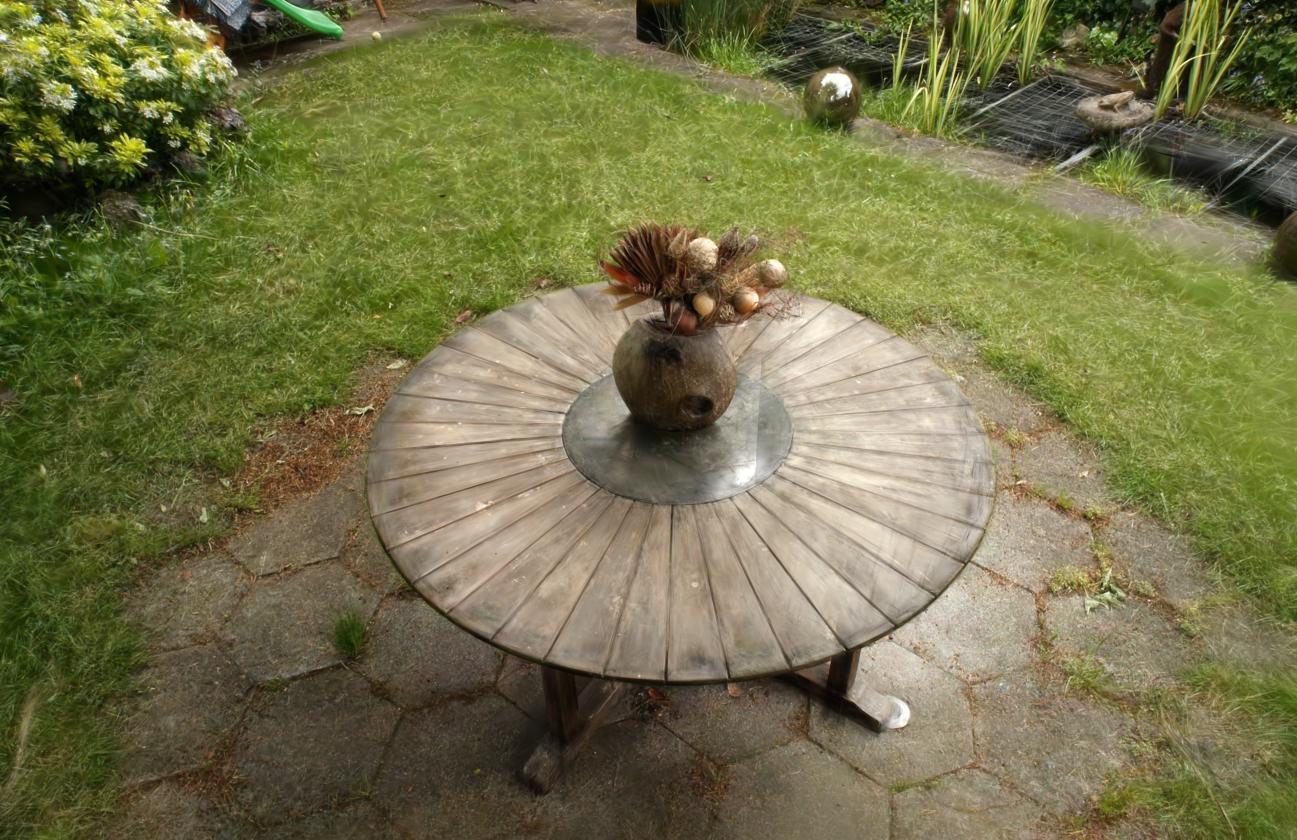} \\

     w/o Affine & w/ Affine

     \end{tabular}
     \vspace{-0.2cm}
     \caption{\textbf{Comparison on Affine Transform Ablation Study.} The absence of the affine transform can lead to significant floaters in the testing views.
     }
     \vspace{-0.6cm}
\label{fig: affine_abl}
\end{figure}

\vspace{-0.2cm}
We apply an affine transform during 3D refinement to prevent 3DGS from learning slightly different color styles generated by diffusion models. In this subsection, we present an ablation study for this component on Garden in Mip-NeRF 360. As shown in \cref{tab: affine_abl}, although removing the affine transform slightly improves PSNR, it results in a decrease in SSIM and LPIPS. Furthermore, as illustrated in \cref{fig: affine_abl}, removing the affine transform results in large floaters in testing views, which can significantly lower human sensory preference.

\begin{figure}
     \centering
     \small 
     \setlength{\tabcolsep}{0pt}
     \def\mywidth{4.2cm}
     \begin{tabular}{P{\mywidth}P{\mywidth}}
     
     \includegraphics[width=\mywidth]{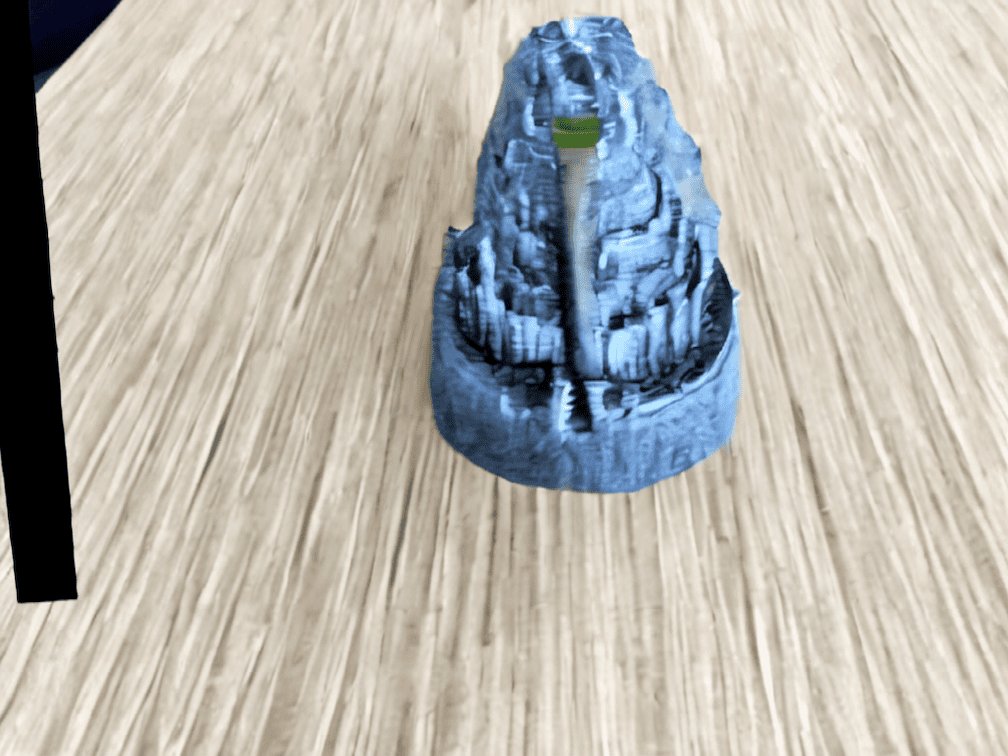}  &
     \includegraphics[width=\mywidth]{figures/exp/llff_mip_comp/fortress/freefix_flux_0017.jpg}  \\
     
     \includegraphics[width=\mywidth]{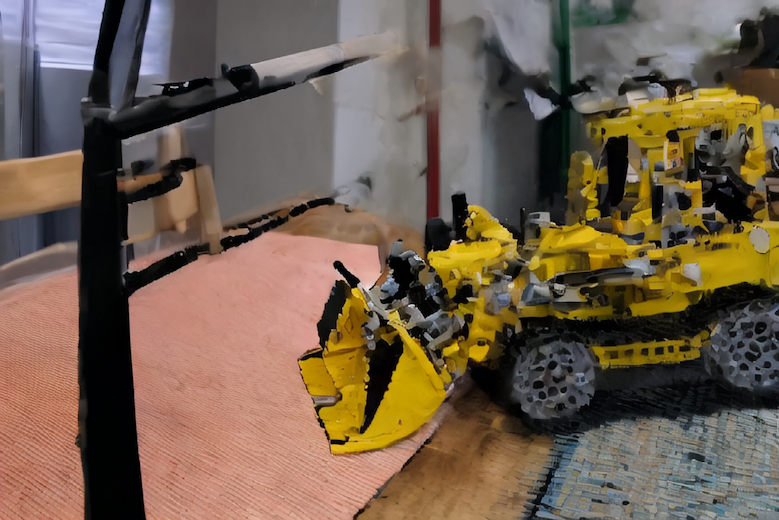} &
     \includegraphics[width=\mywidth]{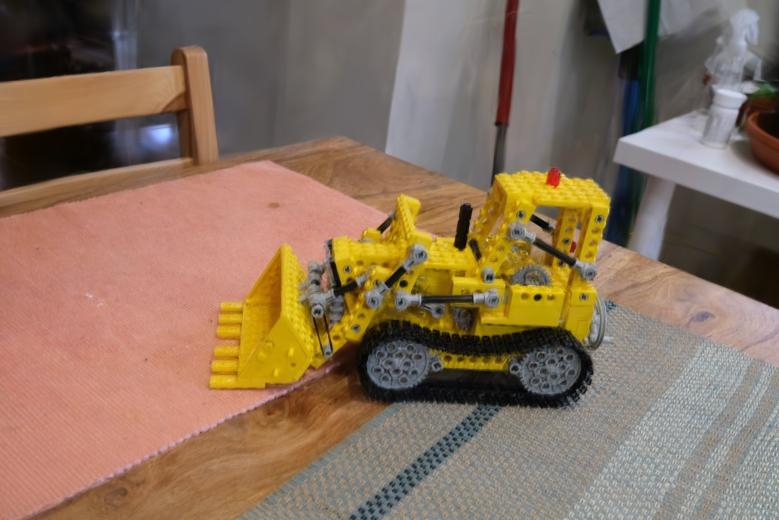} \\

     NVS-Solver \cite{you2024nvs} & FreeFix

     \end{tabular}
     \vspace{-0.2cm}
     \caption{\textbf{Comparisons on FreeFix and NVS-Solver.} The less satisfying results may lead by inaccurate depth and warp results. 
     }
     \vspace{-0.6cm}
\label{fig: nvs}
\end{figure}

\begin{table}
\centering
\small 
\begin{tabular}{lcccc}
\toprule 
& PSNR$\uparrow$ & SSIM$\uparrow$  & LPIPS$\downarrow$ \\
\cline{2-4}
\paperName w/o Affine & \textbf{20.03} & 0.517 & 0.317 \\
\paperName & 19.72 & \textbf{0.520} & \textbf{0.287} \\ 
\bottomrule
\end{tabular}
\vspace{-0.2cm}
\caption{\textbf{Ablation Study on Affine Transform}. Although the affine transform results in a slight decrease in PSNR, this component helps to avoid significant floaters, thereby enhancing SSIM, LPIPS, and overall subjective quality.}
\label{tab: affine_abl}
\vspace{-0.4cm}
\end{table}

\end{document}